%% file: main_paper.tex
\pdfoutput=1

\documentclass[11pt]{article}

\usepackage[preprint]{acl}

\usepackage{times}
\usepackage{latexsym}
\usepackage{verbatim}
\usepackage[T1]{fontenc}

\usepackage[utf8]{inputenc}

\usepackage{microtype}

\usepackage{inconsolata}

\usepackage{graphicx}

\usepackage{bm}
\usepackage{booktabs}
\usepackage{multirow}
\usepackage{colortbl}
\usepackage{caption}
\DeclareCaptionType{Prompt Box}
\DeclareCaptionType{Annotation Guideline}
\usepackage[framemethod=TikZ]{mdframed}
\usepackage{hyperref}
\usepackage{wasysym}
\usepackage{xurl}

\newcommand{\paragraphshort}[1]{\vspace{6pt}\noindent\textbf{#1} $\enspace$}

\newcommand{\realinstructraw}{\textsc{RealInstruct}}
\newcommand{\realinstruct}{\realinstructraw~}

\newcommand{\dcrlongraw}{\textsc{Decompose, Critique, and Refine}}
\newcommand{\dcrlong}{\dcrlongraw~}

\newcommand{\dcrraw}{\textsc{DeCRIM}}
\newcommand{\dcr}{\dcrraw~}

\usepackage{todonotes}
\title{LLM Self-Correction with \dcrraw: \dcrlong for Enhanced Following of Instructions with Multiple Constraints}

\input{latex/sections/author}
\begin{document}
\maketitle
\begin{abstract}
\input{latex/sections/abstract}
%
\end{abstract}

\section{Introduction}
\vspace{-5pt}
\input{latex/sections/introduction}

\section{\realinstruct Dataset}
\label{sec:realinstruct}
\vspace{-2pt}
\input{latex/sections/realinstruct}

\section{Self-correction with \dcrlong (\dcrraw)}
\label{sec:DCR}
\input{latex/sections/self_correction}

\vspace{-2pt}
\section{Experimental Setup}
\label{sec:experimental_setup}
\vspace{-5pt}
\input{latex/sections/experimental_setup}
\vspace{-2pt}
\section{Results and Discussion}
\label{subsec:Experiments}
\vspace{-2pt}
\input{latex/sections/results}
\vspace{-4pt}
\section{Conclusion}
\label{subsec:Conclusion}
\input{latex/sections/conclusion}

\section{Limitations}
\input{latex/sections/limitation}

\vspace{-1pt}
\section{Ethical Considerations}
\input{latex/sections/ethical}

\bibliography{custom}

\onecolumn
\nolinenumbers
\tableofcontents

\appendix
\section*{Appendix}
\section{Related Work}
\label{subsec:Related_Work}
\input{latex/sections/related_work}

\section{Extra Analysis and Discussions of \dcr}
\label{subsec:extra_analysis}
\input{latex/appendix/appendix_extra_analysis}
\section{Definition of Task, Context, and Constraints in \realinstruct}
\label{sec:appendix_definitions}
\input{latex/appendix/appendix_definition_decomposition}

\section{\realinstruct data construction details}
\label{appendix:data_construction}
\input{latex/appendix/appendix_data_construction}

\section{\realinstruct Constraints Categorization}
\label{sec:appendix_data_description}
\input{latex/appendix/appendix_data_description}
\newpage
\section{\realinstruct Data Samples}
\label{appendix:realinstruct_samples}
\input{latex/appendix/appendix_realinstruct_samples}
\clearpage
\newpage
\section{Implementation details for \dcr pipeline}
\label{appendix:pipeline_implementation}
\input{latex/appendix/appendix_implementation_pipeline}

\section{Extra Experimental Details for LLM-as-a-Judge Validation for Constraint Satisfaction}
\label{appendix:critic}
\input{latex/appendix/appendix_critic_evaluation}

\end{document}

%% file: latex/sections/author.tex
\author{
  \textbf{Thomas Palmeira Ferraz\textsuperscript{{\textdaggerdbl}}}\thanks{Work done while interning at Amazon AGI Foundations.},
  \textbf{Kartik Mehta\textsuperscript{{\textdagger}}},
  \textbf{Yu-Hsiang Lin\textsuperscript{{\textdagger}}},
  \textbf{Haw-Shiuan Chang\textsuperscript{{\textdagger}}},
\\
  \textbf{Shereen Oraby\textsuperscript{{\textdagger}}},
  \textbf{Sijia Liu\textsuperscript{{\textdagger}}},
  \textbf{Vivek Subramanian\textsuperscript{\textdagger}},
  \textbf{Tagyoung Chung\textsuperscript{\textdagger}},
\\
  \textbf{Mohit Bansal\textsuperscript{\textdagger\S}},
  \textbf{Nanyun Peng\textsuperscript{\textdagger\P}}
\\
  \textsuperscript{\textdagger}Amazon AGI Foundations,
  \textsuperscript{{\textdaggerdbl}}Télécom Paris, Institut Polytechnique de Paris,
  \\
  \textsuperscript{\S}UNC Chapel Hill,
  \textsuperscript{\P}University of California, Los Angeles
\\
  \small{
    \tt{thomas.ferraz@alumni.usp.br \qquad \{kartim,orabys\}@amazon.com}
  }
}

%% file: latex/sections/abstract.tex
Instruction following is a key capability for LLMs. However, recent studies have shown that LLMs often struggle with instructions containing multiple constraints (e.g. a request to create a social media post ``in a funny tone'' with ``no hashtag''). Despite this, most evaluations focus solely on synthetic data. To address this, we introduce \realinstructraw, the first benchmark designed to evaluate LLMs' ability to follow real-world multi-constrained instructions by leveraging queries real users asked AI assistants. We also investigate model-based evaluation as a cost-effective alternative to human annotation for this task. Our findings reveal that even the proprietary GPT-4 model fails to meet at least one constraint on over 21\% of instructions, highlighting the limitations of state-of-the-art models. To address the performance gap between open-source and proprietary models, we propose the \dcrlong (\dcrraw) self-correction pipeline, which enhances LLMs' ability to follow constraints. \dcr works by decomposing the original instruction into a list of constraints and using a Critic model to decide when and where the LLM's response needs refinement. Our results show that \dcr improves Mistral’s performance by 7.3\% on \realinstruct and 8.0\% on IFEval even with weak feedback. Moreover, we demonstrate that with strong feedback, open-source LLMs with \dcr can outperform GPT-4 on both benchmarks.




%% file: latex/sections/introduction.tex
\input{latex/figures/example_benchmark}
Large Language Models (LLMs) have demonstrated impressive instruction following ability across various tasks, such as creative writing, coding, and arithmetic reasoning \citep{brown2020language,wei2021finetuned,mishra-etal-2022-cross,sanh2022multitask,ouyang2022training,wang2022super}. Despite their remarkable success, recent studies and benchmarks have highlighted \textbf{significant limitations in LLMs' ability to adhere to user-defined rules} (termed as \textit{constraints} henceforth) within instructions \citep{mu2023can,pmlr-v202-zhou23g,lu-etal-2023-bounding,sun-etal-2023-evaluating,zhou2023instruction,jiang2023followbench,yao2024collie,Qin2024InFoBenchEI}. Table \ref{hashtag_example} illustrates an example where four strong LLMs make the same mistake in following constrained instruction. Based on analysis of real user instructions to AI assistants, we estimate that \textbf{30\% of real-user requests to LLMs require satisfying some constraints} (estimation details in Appendix \ref{subappendix:data_construction:data_filtering}). This highlights the importance of evaluating and enhancing LLMs' ability to follow real-world multi-constrained instructions.

\paragraphshort{Benchmarking on Real-World Requests.}These benchmarks often rely on synthetic data either to address data scarcity or to facilitate automatic rule-based evaluation \citep{zhou2023instruction,jiang2023followbench,yao2024collie}. In this work, we argue that synthetic constraints may not accurately capture the complexity and nuances of real-world scenarios, being sometimes artificially difficult. As a result, focusing on synthetic benchmarks may push research in the wrong direction, as improvements on these may not translate to real-world performance and could even degrade it.

To address this gap, we introduce the \realinstruct benchmark, which evaluates LLMs using real user requests to AI assistants. It assesses LLMs' response on individual constraints at a time, as illustrated in Figure \ref{fig:benchmark_flow}. \textbf{Our analysis reveals that even strong proprietary model GPT-4 fails to meet at least one constraint in over 21\% of the instructions}, highlighting the limitations of current LLMs in handling user's constrained instructions.

\paragraphshort{LLM-based evaluation.}Evaluating real-world instructions is challenging due to their open-ended nature. Drawing on recent research on LLM-based evaluation (\textbf{LLM-as-a-Judge}) \citep{zheng2024judging,zeng2024evaluating,Qin2024InFoBenchEI}, we investigate the effectiveness of open and proprietary LLMs as evaluators for constraint satisfaction. To this end, we created a test set of 1k instruction-constraint-response triples with human-verified annotations. Our experiments show that \textbf{GPT-4-Turbo with Chain-of-Thought prompting is a cost-effective and reliable alternative to human evaluation}. However, open-source models, such as Mistral, lag significantly behind proprietary models. Weakly supervising Mistral with GPT-4-Turbo reasoning steps led to 26\% relative improvement, though, still remaining insufficient for reliable evaluation.

\paragraphshort{LLM Self-Correction.}Improving LLMs by generating and building upon intermediate outputs, known as \textbf{System 2 Approaches}, has been explored for constrained instruction following.  But many works in this direction, such as Branch-Solve-Merge \citep{saha2023branch} and Self-Refine \citep{Selfrefine2023}, assume constraint independence or focus on specific constraint types, limiting their applicability to real-world use cases. To address this, we introduce \dcrlong (\dcrraw), a \textbf{self-correction pipeline designed to enhance LLM performance in multi-constrained instruction following without making assumptions about the constraints}. As shown in Figure \ref{fig:selfref_pipeline_example}, \dcr breaks down the original instruction into a main task and granular constraints, and includes a Critic model that decides whether the response is final or requires refinement, which is handled by the underlying LLM.

Through extensive benchmarking, we demonstrate the effectiveness of the \dcr pipeline. \textbf{Even with weak feedback and no external data, \dcr improves Mistral's instruction-level performance by 4.8\% on \realinstruct and 1.2\% on IFEval} relative to baseline. When perfect instruction decomposition into constraints is provided, improvements increase to 7.3\% and 8.0\%, respectively. 
\textbf{Providing stronger feedback further boosts Mistral’s performance by 22.0\% on \realinstruct and 33.8\% on IFEval}, surpassing GPT-4 on both benchmarks.




In summary, our work contributes the following:
\begin{enumerate}
    \vspace{-5pt}
    \item \realinstructraw, the first benchmark of real user requests to LLMs for evaluating instruction following with multiple constraints;
    \vspace{-5pt}
    \item \dcr self-correction pipeline, to the best of our knowledge, the first System 2 approach for constrained instructions that operates without assumptions about constraints, showing significant improvements even with weak feedback;
    \vspace{-5pt}
    \item the first systematic analysis of open-source and proprietary LLM-as-a-Judge models for evaluating constraint satisfaction.
\end{enumerate}
\vspace{-5pt}

The rest of the paper is organized as follows: Sections \ref{sec:realinstruct} and \ref{sec:DCR} present the \realinstruct dataset and the \dcr self-correction pipeline, respectively. Section \ref{sec:experimental_setup} outlines our hypotheses and experimental setup, while Section \ref{subsec:Experiments} analyzes the results. Final remarks are provided in Section \ref{subsec:Conclusion}. We provide extra details and discussions in the Appendix.

\input{latex/figures/decomposition-benchmark}
\input{latex/Tables/tab1_otherbenchmarks}

%% file: latex/figures/example_benchmark.tex
\begin{figure*}[t!]
\vspace{-15pt}
\centering
\includegraphics[width=\textwidth]{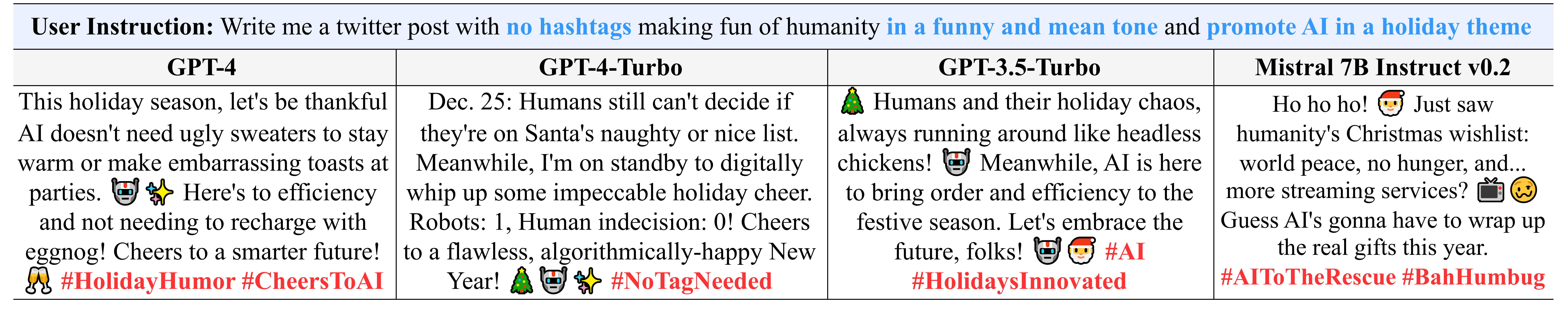}
\vspace{-25pt}
\captionof{table}{\textbf{Example of user instruction where all subject LLMs failed.} Responses from four LLMs are shown. All responses incorrectly include hashtags, despite a constraint explicitly requesting to not do so. Constraints in the instruction are highlighted in blue, and errors in LLM responses are highlighted in red.}
\label{hashtag_example}
\vspace{-10pt}
\end{figure*}

%% file: latex/figures/decomposition-benchmark.tex
\begin{figure*}[t!]
    \vspace{-10pt}
    \centering
    \includegraphics[width=\textwidth]{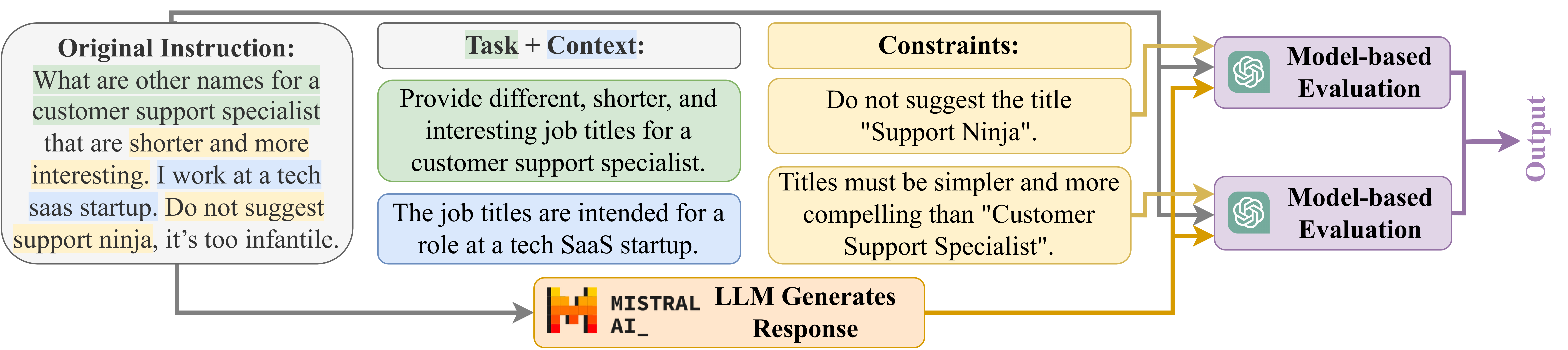} 
    \caption{\textbf{\realinstruct Benchmark Workflow.} Real-user original instruction is input into the Subject LLM, which generates a response. Using the original instruction, decomposed constraints, and the generated response, model-based evaluation assesses the quality of the response against each constraint one at a time, and then aggregates the results into an instruction-level metric.}
    \label{fig:benchmark_flow}
    \vspace{-10pt}
\end{figure*}

%% file: latex/Tables/tab1_otherbenchmarks.tex
\begin{table*}[]
\centering
\resizebox{\textwidth}{!}{
\begin{tabular}{ccccccc}
\hline
{\textbf{Benchmark}}           & {\textbf{\begin{tabular}[c]{@{}c@{}}Instruction \\ source\end{tabular}}} & {\textbf{\begin{tabular}[c]{@{}c@{}}Constraints \\ source\end{tabular}}} & {\textbf{Evaluation}}                                           & {\textbf{\begin{tabular}[c]{@{}c@{}}Size \\ (Instructions)\end{tabular}}}     & {\textbf{\begin{tabular}[c]{@{}c@{}}Constraint \\ types\end{tabular}}} & {\textbf{\begin{tabular}[c]{@{}c@{}}Avg. Constraints \\ per Instruction\end{tabular}}} \\ \hline
{\begin{tabular}[c]{@{}c@{}}COLLIE\end{tabular}}                       & {Synthetic}                                                              & {Synthetic}                                                              & {Rule}                                                    & {2,080}                                                                       & {13}                                                                   & {N/A}                                                                                  \\
{\begin{tabular}[c]{@{}c@{}}IFEval\end{tabular}}                        & {Synthetic}                                                              & {Synthetic}                                                              & {Rule}                                                    & {541}                                                                         & {25}                                                                   & {1.4}                                                                                  \\
{\begin{tabular}[c]{@{}c@{}}FollowBench\end{tabular}}                  & {\begin{tabular}[c]{@{}c@{}}Crowdsourced\\ + Synthetic\end{tabular}}   & {Synthetic}                                                              & {\begin{tabular}[c]{@{}c@{}}Model / Rule\end{tabular}} & {795}                                                                         & {6}                                                                    & {5}                                                                                    \\
{\begin{tabular}[c]{@{}c@{}}InfoBench\end{tabular}}                  & {\begin{tabular}[c]{@{}c@{}}Crowdsourced\end{tabular}}   & {Crowdsourced}                                                              & {\begin{tabular}[c]{@{}c@{}}Model / Rule\end{tabular}} & {500}                                                                         & {5}                                                                    & {4.5}                                                                                    \\
{\textbf{\begin{tabular}[c]{@{}c@{}}\realinstructraw\\(ours)\end{tabular}}} & {\textbf{\begin{tabular}[c]{@{}c@{}}Real\\ Users\end{tabular}}}    & {\textbf{\begin{tabular}[c]{@{}c@{}}Real\\ Users\end{tabular}}}    & {\textbf{Model}}                                                & {\textbf{\begin{tabular}[c]{@{}c@{}}302 (test) \\ + 842 (val)\end{tabular}}} & {\textbf{20+}}                                                          & {\textbf{3.5 (test)}}                                                                         \\ \hline
\end{tabular}
}
\vspace{-5pt}
\caption{Comparison of \realinstruct samples with benchmarks such as COLLIE \citep{yao2024collie}, IFEval \citep{zhou2023instruction}, FollowBench \citep{jiang2023followbench} and InfoBench \citep{Qin2024InFoBenchEI}.}
\label{tab:other_benchmarks}
\vspace{-10pt}
\end{table*}

%% file: latex/sections/realinstruct.tex
We introduce \realinstructraw, a novel dataset consisting of original user-generated instructions, each decomposed into a task and a set of constraints (see Figure \ref{fig:benchmark_flow} for an example). The task represents the user’s main objective -- the outcome they expect from LLM -- and may optionally include context to aid the model’s understanding. Constraints are conditions or limitations that guide the LLM’s generation. Formal definitions of these terms are provided in Appendix \ref{sec:appendix_definitions}. Since users rarely specify constraints in a structured list format, the decomposition breaks instructions into manageable items, providing the necessary granularity for evaluating LLMs' ability to follow constraints.

Using this dataset, we develop a benchmark protocol that evaluates LLM performance on each constraint at a time, generating constraint-level scores that are then aggregated into an instruction-level accuracy metric. Figure \ref{fig:benchmark_flow} provides an overview of the benchmark protocol. Given the open-ended nature of real user instructions, rule-based evaluation is infeasible, so we adopted the LLM-as-a-Judge evaluation. A detailed discussion and validation of this protocol are provided in Section \ref{sec:constraint_verification}.

\input{latex/figures/pipeline_with_example}

\subsection{Data Description}
\label{subsec:datadescription}
The \realinstruct dataset is divided into two splits: test and validation. The test split, intended for LLM evaluation, was human-validated and contains 302 instructions with 1,055 constraints. The validation split, used for method validation such as training judges, was not human-validated and includes 842 instructions with 2,500 constraints. Table \ref{tab:other_benchmarks} compares \realinstruct with existing benchmarks for evaluating LLMs’ ability to follow multi-constrained instructions. Unlike other datasets that rely on synthetic or crowdsourced instructions, \realinstruct uniquely captures real user interactions with AI assistants, offering a more realistic representation of real-world use cases.

To better understand constraint characteristics in \realinstructraw, we manually categorized them into 22 distinct groups. Detailed dataset statistics are presented in Appendix \ref{sec:appendix_data_description}. When comparing \realinstruct with IFEval, one of the popular benchmark for constraint following that consists solely of synthetic constraints, we found that only 6.3\% of \realinstruct constraints overlap with the 25 types in IFEval, and 11 IFEval constraint types never appear in \realinstructraw. This discrepancy highlights the gap between synthetic datasets and real LLM use cases, with synthetic constraints failing to adequately represent the challenges users may pose to LLMs.

\subsection{Data Construction}
\label{subsec:dataconstruction}

We built \realinstruct dataset using prompts from a public dataset of conversations between users and AI assistants\footnote{We use only the first user turn (no model responses) from a public dataset of examples originally sourced from ShareGPT at: \url{https://huggingface.co/datasets/anon8231489123/ShareGPT_Vicuna_unfiltered}}. We filtered and processed the dataset in five steps: 1) removed AI responses, retaining only the first user turn (instruction); 2) excluded non-English instructions; 3) filtered out code-related instructions using an open-source LLM as a zero-shot classifier; 4) kept only instructions containing constraints, again using a LLM classifier; and 5) manually validated the remaining data for relevance, clarity, and safety. See Appendix \ref{subappendix:data_construction:data_filtering} for details on the data filtering.

Following filtering, we decomposed the instructions into tasks, contexts, and constraints using GPT-4 \citep{achiam2023gpt}. In our tests, we found that this three-part decomposition (task, context, constraints) provided more robust outcomes compared to splitting into just task and constraints. The task and context were then concatenated for further processing. A subset of 302 instructions underwent manual inspection to ensure accuracy and granularity, forming the test split, while the remaining 842 instructions make up the validation split. See Appendix \ref{subappendix:data_construction:decomposition} for more details on the data decomposition process.

%% file: latex/figures/pipeline_with_example.tex
\begin{figure*}[t!]
    \vspace{-10pt}
    \centering
    \includegraphics[width=\linewidth]{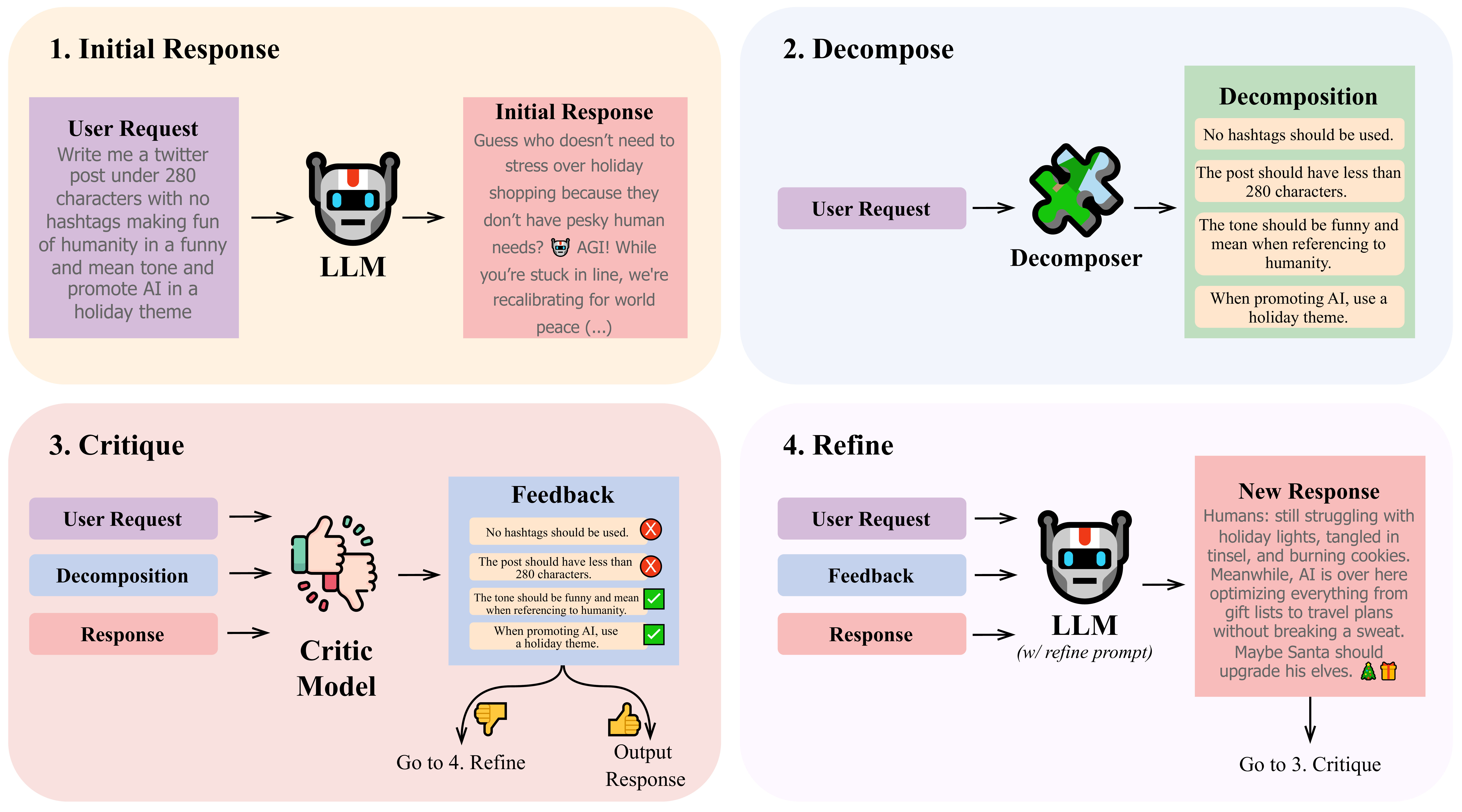} 
    \caption{
    \textbf{The \dcr pipeline.} Initially, the LLM generates a response to a user request. The Decomposer breaks down the request into granular constraints. A Critic model then gives feedback on whether the response meets all constraints. If it does, the response is output; if not, the feedback is used by LLM to refine the response. This \textbf{Critique--Refine} cycle repeats until all constraints are satisfied or the maximum number of iterations is reached.}
    \label{fig:selfref_pipeline_example}
    \vspace{-10pt}
\end{figure*}

%% file: latex/sections/self_correction.tex
In this section, we present our proposed \dcr self-correction pipeline, designed to enhance LLM responses to follow user constraints. Given a multi-constrained user instruction, the pipeline iteratively refines the LLM's response through four key steps: Initial Response, Decompose, Critique, and Refine, iterating until the response meets all constraints or a maximum number of iterations $N_{max}$ is reached. Figure \ref{fig:selfref_pipeline_example} provides an overview of the proposed pipeline and we detail each of these steps below:

\paragraphshort{1. Initial Response} is generated directly from the original user instruction using a strong prompt.

\paragraphshort{2. Decompose} the original instruction into a list of granular constraints to be followed. This task is similar to the instruction decomposition performed in \realinstructraw, but here, the decomposer model focuses solely on listing the constraints, simplifying the task. A prompt example is provided in Appendix \ref{subappendix:self_decomposition}.

\paragraphshort{3. Critique} the response using the Critic model to identify any unsatisfied constraints. If all constraints are satisfied, the response is considered final. Otherwise, the Critic provides feedback in natural language, specifying which constraints were not satisfied.

\paragraphshort{4. Refine} the response using the Critic's feedback to address unsatisfied constraints. The underlying LLM generates an improved response using a refinement prompt that incorporates the feedback, along with the original instruction and previous response.

The \textbf{Critique} and \textbf{Refine} steps can be repeated iteratively until the response satisfies all constraints or the specified maximum number of iterations is reached.

To our knowledge, \dcr is the first approach specifically designed to tackle the challenge of following instructions with multiple open-ended constraints. Unlike previous methods, which assume constraint independence or focus on specific constraint types, our approach makes no assumptions about the nature of user constraints. In Section \ref{related_work:self_correction} of Related Work, we compare our pipeline with other works for constrained generation, including Self-Correction and System 2 approaches.

%% file: latex/sections/experimental_setup.tex
This section outlines the experimental setting to evaluate our proposed methods. We first validate the use of LLM-as-a-Judge for the \realinstruct benchmark (Sec. \ref{sec:constraint_verification}). Next, we benchmark models on instructions with multiple constraints (Sec. \ref{subsec:realinstruct_eval_protocol}) and, finally, test our proposed self-correct \dcr pipeline to improve open-source models in this task (Sec. \ref{subsec:experiments_decrim}). All experiments with open LLMs were performed with HuggingFace's Transformers library \citep{wolf-etal-2020-transformers} on an AWS instance with 8 V100 32GB GPUs. Times are reported in this configuration.

\subsection{Validating LLM-as-a-Judge for Constraint Satisfaction}
\label{sec:constraint_verification}
Given the open-ended nature of \realinstruct benchmark instructions, rule-based or reference-guided evaluation is not feasible. Drawing from recent work showing the effectiveness of LLMs as evaluators, particularly GPT-4 \citep{zheng2024judging,zeng2024evaluating}, we adopt an LLM-as-a-Judge evaluation protocol for \realinstructraw. To assess LLM-as-a-judge reliability compared to human evaluators and identify the most cost-effective approach, we introduce \textit{EvalJudge}, a test set of instruction-constraint-response triples with ground-truth labels (Sec. \ref{section:ceb}). We benchmark both proprietary and open-source models using four adaptation strategies (Sec. \ref{subsec:critiquemodel}) against human judgments on this set. More specifically, we investigate whether open-source models can match the performance of high-cost proprietary models in the LLM-as-a-Judge role.

\subsubsection{Adaptation strategies}
\label{subsec:critiquemodel}

\paragraphshort{Models} We evaluate three proprietary and three open-source LLMs as candidates for LLM-as-a-Judge for Constraint Satisfaction. The proprietary models include GPT-4 (\texttt{gpt-4-0314}), GPT-4-Turbo (\texttt{gpt-4-turbo-2024-04-09}), and GPT-3.5-Turbo (\texttt{gpt-3.5-turbo-0125}), ordered by decreasing API cost\footnote{Refer to: \url{https://openai.com/api/pricing}}. For the open-source models, we experiment with Mistral 7B Instruct v0.2 \citep{jiang2023mistral} (hereafter referred to as Mistral v0.2), Vicuna 7B v1.3 \citep{vicuna2023}, and Zephyr 7B $\beta$ \citep{tunstall2023zephyr}, all top performers on the Open LLM Leaderboard as of February 2024 \citep{open-llm-leaderboard-v1}.

We explore four adaptation strategies, including three In-Context Learning (ICL) approaches and one weakly supervised fine-tuning for open-source model. Some strategies use the Chain-of-Thought (CoT) prompt, known to improve LLM reasoning \citep{cot2022neurips,zheng2024judging}, and we also investigate whether to evaluate constraints individually or collectively. The strategies are as follows:

\paragraphshort{a) Instruction-wise Eval (ICL-Inst.):}All constraints within an instruction are evaluated simultaneously. The LLM-as-a-Judge is presented with an instruction and a list of constraints, and using a CoT prompt with an in-context example containing an instruction and two constraints, it generates reasoning and predictions for all constraints at once.

\paragraphshort{b) Constraint-wise Eval (ICL-Const.):}Each constraint is evaluated independently. For every response-constraint pair, the LLM-as-a-Judge directly predicts "Constraint followed" or "Constraint not followed." Two constraint-response pairs serve as in-context examples.

\paragraphshort{c) Constraint-wise Eval + CoT (ICL-Const.+CoT):}The LLM-as-a-Judge is prompted to generate reasoning for each constraint, followed by a prediction of "Constraint followed" or "Constraint not followed." Evaluation is also performed for each response-constraint pair independently, using two constraints as in-context examples.

\paragraphshort{d) Weakly Supervised Open LLM (Supervised):} We fine-tuned Mistral for the LLM-as-a-judge task. We construct a training data using the weak instruction annotations from \realinstructraw’s validation set. We generate Mistral responses to these instructions, and weak constraint satisfaction annotations with reasoning trails from GPT-4-Turbo with \textbf{ICL-Const.+CoT} prompt. We fine-tune Mistral v0.2 using LoRA adapters \citep{hu2021lora} on this dataset, guiding model to mimic GPT’s reasoning. Further details can be found in Appendix \ref{subappendix:mistral_training}.


\textbf{All prompt templates are provided in Appendix \ref{subappendix:constraint_verification_prompts}}. Costs and processing times for each configuration are detailed in Table \ref{tab:CEB}.

\subsubsection{The \textit{EvalJudge} Dataset}
\label{section:ceb}

Since no public dataset exists for the task of evaluating whether a given response satisfies a specific constraint or not, we create \textit{EvalJudge} dataset, derived from the test split of the \realinstruct dataset. To ensure diversity, we divided the instructions into two subsets, generating one response per instruction using Mistral v0.2 for one subset and Vicuna v1.3 for the other. \textit{EvalJudge} contains 294 instructions and 982 constraints. Notably, 81.4\% of the samples are labeled as "constraint satisfied."

\paragraphshort{Ground truth and Baselines:} 
\hfill
\\
\noindent\textbf{a) Human Annotation:} We create an Amazon Mechanical Turk (MTurk) task to collect independent labels from two pools of annotators for each constraint-response pair in the dataset. Guidelines are in Appendix \ref{appendix:human_annotation_critic}.

\noindent\textbf{b) Expert Annotation:} To obtain a third independent annotation, the authors manually annotated the dataset using the same MTurk guidelines. In cases where there was disagreement between the two human annotators, GPT-4, and the best GPT-4-Turbo labels, a second review was conducted by the authors. These conflicts accounted for 28.3\% of the samples. The final labels, referred to as "Expert," serve as the ground truth for EvalJudge.

\noindent\textbf{c) Other Baselines:} To help interpret the results, we introduce two simple baselines representing extreme cases: \textbf{"All Satisfied,"} where the annotator labels all constraints as satisfied, and \textbf{"All Not Satisfied,"} where the annotator marks all constraints as unsatisfied. Additionally, we use the \textbf{"Majority Vote"} from the three human annotators to benchmark model performance against human judgment.

\paragraphshort{Evaluation Metrics} Unlike \citet{Qin2024InFoBenchEI}, who used accuracy as the primary metric for LLM-as-a-Judge evaluation, we account for the positive class imbalance in EvalJudge by using Macro-averaged F1-Score (\textbf{Macro F1}) as our main metric. Additionally, we report the F1-score for the negative class (\textbf{F1 Negative}), which measures the balance between false alarms and omissions. Different judges are compared to human judment using the \textbf{Cohen’s kappa} inter-rater reliability against human majority vote. We employ \textbf{Krippendorff's alpha} (an extension of kappa for more than two annotations) to assess inter-human agreement across the three annotations. 

\subsection{Benchmarking LLMs on \realinstruct}
\label{subsec:realinstruct_eval_protocol}
We benchmark models on \realinstruct for the task of following multi-constrained instructions. The evaluation includes two proprietary models--GPT-4 and GPT-3.5-Turbo--and three open-source LLMs: Mistral v0.2, Vicuna v1.3, and Zephyr $\beta$, selected based on their performance on Chatbot Arena \citep{chiang2024chatbot}. The judge is GPT-4-Turbo with \textbf{ICL-Const+CoT} prompt. Following \citet{zhou2023instruction,saha2023branch}, we report accuracy at both the instruction and constraint levels.

\subsection{Evaluating \dcr pipeline}
\label{subsec:experiments_decrim}

To validate the effectiveness of our \dcr pipeline, we conduct experiments using Mistral v0.2 as the underlying LLM. We use $N_{max} = 10$. We also investigate the contribution of different Decomposer and Critic models for the pipeline, and compare the performance against the proprietary model GPT-4. We present extra comparisons on Appendix \ref{subsec:extra_analysis}.

\paragraphshort{Datasets} We evaluate model performance on the \realinstruct dataset and the IFEval dataset \citep{zhou2023instruction}. While IFEval is based on synthetic constraints, it serves as a valuable benchmark for validating our pipeline, as it is popularly used for evaluating constrained instruction-following.

\paragraphshort{Baselines} To measure the improvements introduced by our method, we establish the following baselines with Mistral v0.2:

\noindent\textbf{1. Conventional:} Only the instruction as input.

\noindent\textbf{2. "Make sure":} Appends the text “\textit{Make sure to follow all the provided constraints}” to the instruction, creating a strong and fair baseline.

\noindent\textbf{3. Self-Refine:} Adapts the \citeposs{Selfrefine2023} approach for the case of multi-constraint instructions. While Self-Refine uses the model itself as its critic without additional context, our \dcr pipeline employs a critic with fine-grained evaluation, assessing each constraint individually. This baseline helps quantify the value added by this modeling.

\paragraphshort{Decomposer}We use a \textbf{Self-Decomposer}, where the LLM itself lists relevant constraints, simplifying the decomposition from Section \ref{subsec:dataconstruction} by omitting the request for task and context. The prompt for this approach is detailed in Appendix \ref{subappendix:self_decomposition}.

\paragraphshort{Critic Model}We explore two Critic models based on the underlying LLM: 

\noindent\textbf{a) Self-Critic:} Uses the model itself as the Critic, with the best ICL-based adaptation (from Section \ref{subsec:critiquemodel}), specifically the ICL-Const+CoT prompt.

\noindent\textbf{b) Supervised Critic:} the Mistral weakly supervised for LLM-as-a-judge, as described in \ref{subsec:critiquemodel}.

\paragraphshort{Ablations with Oracle and GPT-4}To understand the impact of strong Decomposer and Critic modules, we conduct ablations using Oracles, which are  ideal representations of the upper bound of each component. The \textbf{Oracle Decomposer} is the gold-standard list of constraints for both \realinstruct and  IFEval. This provides insights on the ability to judge well the responses, knowing what to judge. The \textbf{Oracle Critic} is GPT-4-Turbo for \realinstruct and the lenient rule-based evaluation program for IFEval. Additionally, for IFEval, we explore GPT-4 as a strong LLM (but less performant than Oracle) serving as the critic model. These Critic ablations provides insights on the ability of LLMs to refine itself when it knows what need to be refined. 

\paragraphshort{Overall Quality Assessment (OQA)}To ensure the pipeline does not degrade the quality of final responses, we perform Pairwise Quality Ranking using Prometheus-2 \citep{kim2024prometheus2}, an open LLM-as-a-Judge for general quality evaluation. This is done only on responses revised by the pipeline, comparing the initial response with the final revised one. More details in Appendix \ref{subappendix:oqa}.



%% file: latex/sections/results.tex
\input{latex/Tables/tab_CEB_new}
We discuss our results, particularly the Reliability of LLM-as-a-Judge for Constraint Satisfaction (Section~\ref{subsec:CEBBenchmark}), the ability of various open-source and proprietary LLMs on follow multi-constrained instructions (Section~\ref{subsec:RealInstructBenchmark}), and the efficacy of our \dcr self-correction pipeline (Section \ref{subsec:DCR_subsection}). 
\vspace{-2pt}
\subsection{Reliability of LLM-as-a-Judge}
\label{subsec:CEBBenchmark}

Results from several model and baseline judges on the EvalJudge dataset are presented in Table \ref{tab:CEB}. Human inter-rater reliability (Human 1, Human 2, and Expert) is moderate (Krippendorff's $\alpha$ = 0.61), with lower agreement between Humans 1 and 2 ($\kappa$ = 0.44). This highlights the inherent challenges in verifying constraint satisfaction, as the task can be subjective and ambiguous, when involving multiple sub-constraints, despite efforts to minimize these issues during data annotation (see Appendix \ref{subsubappendix:decomposition_validation}).

\paragraphshort{GPT-4-Turbo with CoT is Reliable and More Cost-Efficient.} We observe that GPT-4, a widely used LLM-as-a-Judge, shows lower performance compared to humans while maintaining moderate correlation with humans ($\kappa$ = 0.42). Using GPT-4-Turbo, evaluating constraints individually yields better results than evaluating them all at once, despite a 37\% cost increase. \textbf{GPT-4-Turbo with CoT prompt offers a more performant and cheaper alternative to GPT-4}. It reduces costs by 57\% while improving overall performance (Macro F1) by +7.0\% and in detecting unmet constraints (F1 Negative) by +19.0\%. Its correlation with Expert annotation is similar to that of Human 2 (0.58 vs. 0.60). We thus adopt GPT-4-Turbo with CoT as the standard evaluation for \realinstructraw.

\paragraphshort{Open-source LLMs are unreliable judges.} ICL-based configurations of open-source LLMs (Mistral, Vicuna, Zephyr) exhibit poor performance in all scenarios, particularly in detecting unmet constraints. Vicuna and Zephyr closely mirror the "All Satisfied" baseline, suggesting they are lenient judges. Mistral, despite similar Macro-F1, diverges in other metrics, indicating more random than lenient decisions (similar to GPT-3.5-Turbo). When weakly supervised with GPT-4-Turbo's CoT reasoning trails, Mistral significantly improves overall performance and the ability to detecting unmet constraints (+12.9 in Macro F1 and +28.1 in F1 Neg.). \textbf{This reduces the macro F1 gap with GPT-4-Turbo by about 50\%, but the model still shows poor agreement with humans, indicating that open-source LLMs are not yet reliable judges}.
\subsection{LLMs' ability to follow multi-constrained instructions on \realinstruct}
\label{subsec:RealInstructBenchmark}
\input{latex/Tables/decrim_new_table2}
\input{latex/Tables/tab2_benchmark_results}
Benchmarking results for all models on \realinstruct are presented in Table \ref{tab:realinstructbench}. We observe that \textbf{even with strong proprietary model, GPT-4, over 21\% of instructions have at least one unsatisfied constraint}. Additionally, the open-source model Mistral v0.2 outperforms proprietary GPT-3.5 but falls short of GPT-4's performance. A qualitative examination of responses with unsatisfied constraints suggests that \textbf{LLMs often struggle with constraints involving numbers, negations, or long instructions with large number of constraints}, which is consistent with findings from previous works \citep{numericalLLM,garcia-ferrero-etal-2023-dataset,truong-etal-2023-language,jiang2023followbench}. We also discuss intra-model scoring bias problem in the Limitations Section. Overall, the results highlight the need for further enhancement of LLMs in handling multi-constrained instructions.

\subsection{Effectiveness of \dcr Self-Correction}
\label{subsec:DCR_subsection}

We present the results of our experiments with Mistral v0.2 using the \dcr Self-Correction pipeline in Table \ref{tab:selfref_realinstruct}. A slight improvement is observed with the \textit{Make Sure} prompt compared to the conventional prompt on both datasets, demonstrating that it serves as a strong baseline for first-generation responses. Consequently, we adopt \textit{Make Sure} as the baseline for further comparisons.

We classify the \dcr configurations into three categories: (1) \textbf{Fairly Comparable}, where the pipeline operates independently of  external models; (2) \textbf{Realistic Ablations}, which employ fairly comparable Critic models but use ideal Oracle Decomposer, which remains a realistic measure under the guidelines from \citet{selfcorrectionsurvey2024}, since decomposition is relatively straightforward and could be handled by a dedicated LLM; and (3) \textbf{Unrealistic Ablations}, which rely on ideal Critic and Decomposer models, useful for upper bound estimation but with limited generalization.

\paragraphshort{LLMs Cannot Self-Refine.}We observe only marginal improvements on \realinstruct and a performance drop on IFEval using the Self-Refine \citep{Selfrefine2023} pipeline, highlighting the limitations of traditional self-correction approaches for the multi-constrained instruction following task. Similarly, our \dcr pipeline showed minimal to no gains when the LLM itself was used as both Decomposer and Critic. This shows that a self-refinement limitation exists even when the model is instructed to look specifically at constraints. \textbf{These poor results are attributed to low-quality Critic feedback, which can lead to over-refining good responses and neglecting to fix bad ones.} This aligns with findings by \citet{Huang2023LargeLM,selfcorrectionsurvey2024}, which showed that LLMs struggle with self-correction without external guidance. Additionally, this aligns with results from Sec. \ref{subsec:CEBBenchmark}, in which Vanilla Mistral failed to reliably detect unsatisfied constraints.

\paragraphshort{\dcr is effective, even with a Weak Critic.} The results show significant improvement with the introduction of Supervised Mistral, a weak Critic that outperforms the LLM's self-critique but still underperforms as a Critic, as per results in Section \ref{subsec:CEBBenchmark}. With a Self-Decomposer and Supervised Critic, performance increased by +3.7 on \realinstruct and +0.7 on IFEval. With an Oracle Decomposer specifying accuratly which constraints to check, improvements were +5.6 and +4.8, respectively, despite the harsher domain shift in IFEval, where Mistral Supervised performed even weaker. 

Introducing a stronger Critic model, GPT-4, emulating what would be the performance of a model trained with high-quality data, resulted in an improvement of +8.1 on IFEval. However, it is worth mentioning that GPT-4 as evaluator only achieved a Macro F1 of 62.9\% when judging Mistral's \textit{Make Sure} responses on IFEval, being far from an ideal Critic for this dataset. Using an Oracle Critic -- representing the strongest possible feedback -- the model showed improvements of +16.9 on \realinstruct and +20.3 on IFEval, \textbf{demonstrating that open-source LLMs can correct its responses when given high-quality feedback}. This highlights the potential of the \dcr pipeline for multi-constrained instruction following, particularly when strong feedback is available. 

Comparing \dcr with proprietary GPT-4, even weak Critics enabled Mistral to surpass GPT-4's performance on \realinstructraw, though only Oracle Critic outperformed GPT-4 on IFEval, likely due to its harder nature. Overall, these results demonstrate that \textbf{LLMs can achieve significant improvements in multi-constrained instruction following when provided with minimally reliable feedback}. As expected, \textbf{higher-quality feedback leads to better performance, with a plateau beyond the performance of proprietary models.}

\paragraphshort{\dcr improves the quality of responses.}Table \ref{tab:selfref_realinstruct} also provides a comparison of overall response quality before and after pipeline revisions. In most cases, the quality remained the same, but when changes occurred, on the \realinstruct dataset, the revised responses were more oftenly preferred. This indicates that \dcr either maintains or improves response quality in the majority of cases. Additionally, we observed a strong correlation between quality improvement and successful revisions, though a high number of revisions can negatively impact quality. This explains the higher quality improvement with better feedback on \realinstruct and higher quality degradation observed on IFEval, which contains more difficult and unachievable constraints.

\paragraphshort{Importance of Decomposer and Critic Models.} Our results also highlight the essential roles of the Decomposer and Critic in the \dcr pipeline. While the Decomposer helps guide the Critic to the right constraints, the Critic's quality has the most significant impact on performance. The weakest configuration (Self-Decomposer and Self-Critic) results in minimal gains or slight degradation. However, even with a weak Self-Decomposer, improvements occur with a better Critic, Supervised, which boosts score in \realinstruct from 75.2 to 80.5. The Oracle Decomposer consistently enhances performance across both benchmarks, but the Critic’s quality drives the largest improvements. For example, using the Oracle Decomposer with the Supervised Critic yields a +4.8 instruction accuracy increase in IFEval, and with stronger Critics like GPT-4 or Oracle, the relative gains rise to +8.1 and +20.3, respectively. Same happens with Oracle Critic for \realinstructraw. These demonstrates that \textbf{while a better Decomposer is beneficial, the Critic's ability to correctly judge responses is the key factor for \dcrraw's success}, with Oracle Critic consistently delivering the highest performance. In fact, the Critic's success is closely tied to the Decomposer's accuracy in pointing the constraints to be verified. In Appendix \ref{subappendix:generate_rank} we discuss the role of Refine and its robustness to weak Critic. 

%% file: latex/Tables/tab_CEB_new.tex

\begin{table}[!t]
\centering
\resizebox{0.99\columnwidth}{!}{%
\begin{tabular}{l@{}ccccc}
\specialrule{.1em}{.05em}{.05em} 
\rowcolor[HTML]{B7B7B7} 
\multicolumn{1}{c}{\cellcolor[HTML]{B7B7B7}\textbf{Annotator}} &
  \textbf{\begin{tabular}[c]{@{}c@{}}Cost\\ (USD)\end{tabular}} &
  \textbf{\begin{tabular}[c]{@{}c@{}}Time \\ (min)\end{tabular}} &
  \textbf{\begin{tabular}[c]{@{}c@{}}Macro\\ F1 (\%)\end{tabular}} &
  \textbf{\begin{tabular}[c]{@{}c@{}}F1 Neg. \\ (\%)\end{tabular}} &
  \textbf{\begin{tabular}[c]{@{}c@{}}Cohen's $\bm{\kappa}$\\ Maj. Vote \end{tabular}} \\ \specialrule{.1em}{.05em}{.05em}
Expert &
  - &
  - &
  100.0 &
  100.0 &
  0.93 \\
Human 1 &
  300.0 &
  - &
  85.1 &
  75.9 &
  0.77 \\
Human 2 &
  300.0 &
  - &
  80.0 &
  66.9 &
  0.66 \\
Majority Vote &
  - &
  - &
  96.4 &
  94.1 &
  1.00 \\ \midrule
All Satisfied &
  - &
  - &
  44.9 &
  0.0 &
  -0.09 \\
All Not Satisfied &
  - &
  - &
  15.7 &
  31.4 &
  -0.70 \\ \midrule
GPT-4 w/ ICL-Cons &
  19.5 &
  - &
  73.7 &
  54.9 &
  0.42 \\ \midrule
GPT-3.5-Turbo \\ 
$\enspace$ w/ ICL-Cons &
  1.0 &
  - &
  51.3 &
  16.6 &
  0.09 \\ \midrule
\textbf{GPT-4-Turbo} \\ 
$\enspace$ w/ ICL-Inst &
  4.1 &
  - &
  72.0 &
 48.5 &
  0.36 \\
$\enspace$ w/ ICL-Cons &
  6.5 &
  - &
  72.6 &
  54.8 &
  0.46 \\
\textbf{$\enspace$ w/ ICL-Cons+CoT} &
  \textbf{8.3} &
  \textbf{-} &
  \textbf{79.0} &
  \textbf{65.5} &
  \textbf{0.50} \\ \midrule
\textbf{Mistral v0.2}  \\
$\enspace$ w/ ICL-Cons &
  - &
  10 &
  50.4 &
  11.4 &
  0.02 \\
$\enspace$ w/ ICL-Cons+CoT &
  - &
  26 &
  53.7 &
  21.9 &
  0.18 \\ 
\textbf{Supervised} &
  \textbf{-} &
  \textbf{236} &
  \textbf{63.3} &
  \textbf{39.5} &
  \textbf{0.28} \\ \midrule
Zephyr $\beta$ \\
$\enspace$ w/ ICL-Cons &
  - &
  11 &
48.1 &
  2.2 &
  0.05 \\
$\enspace$ w/ ICL-Cons+CoT &
  - &
  49 &
  48.2  &
  10.4 &
  0.03 \\ \midrule
Vicuna v1.3 \\
$\enspace$ w/ ICL-Cons &
  - &
  9 &
  47.2 &
  1.4 &
  0.12 \\
$\enspace$ w/ ICL-Cons+CoT &
  - &
  27 & 
  52.1 & 
  10.2 & 
  0.04 \\ \specialrule{.1em}{.05em}{.05em} 
\end{tabular}%
}
\caption{Performance of different approaches for Constraint Verification task on our \textit{EvalJudge} dataset. 
}
\label{tab:CEB}
\vspace{-14pt}
\end{table}

%% file: latex/Tables/decrim_new_table2.tex
\begin{table*}[]
\resizebox{\textwidth}{!}{
\begin{tabular}{ccc|ccccc|ccccc}
\specialrule{.1em}{.05em}{.05em}
\rowcolor[HTML]{B7B7B7} 
\cellcolor[HTML]{B7B7B7}                                                                                       & \cellcolor[HTML]{B7B7B7}                                      & \cellcolor[HTML]{B7B7B7}                                         & \multicolumn{5}{c|}{\cellcolor[HTML]{B7B7B7}\textbf{\realinstructraw}}                                                                                                                                                                                                                                                                                                                                & \multicolumn{5}{c}{\cellcolor[HTML]{B7B7B7}\textbf{IFEval}}                                                                                                                                                                                                                                                                                                                                      \\  \arrayrulecolor{black}  
\cline{4-13} 
\rowcolor[HTML]{B7B7B7} 
\multirow{-2}{*}{\cellcolor[HTML]{B7B7B7}\textbf{Strategy}}                                                    & \multirow{-2}{*}{\cellcolor[HTML]{B7B7B7}\textbf{Decomposer}} & \multirow{-2}{*}{\cellcolor[HTML]{B7B7B7}\textbf{Critic}}        & \textbf{\begin{tabular}[c]{@{}c@{}}Best\\ N\end{tabular}} & \textbf{\begin{tabular}[c]{@{}c@{}}Instruction\\ Acc (\%)\end{tabular}} & \textbf{\begin{tabular}[c]{@{}c@{}}Constraint\\ Acc (\%)\end{tabular}} & \textbf{\begin{tabular}[c]{@{}c@{}}OQA (\%)\\ Win / Lose\end{tabular}} & \multicolumn{1}{c|}{\cellcolor[HTML]{B7B7B7}\textbf{\begin{tabular}[c]{@{}c@{}}Time\\ (h)\end{tabular}}} & \textbf{\begin{tabular}[c]{@{}c@{}}Best\\ N\end{tabular}} & \textbf{\begin{tabular}[c]{@{}c@{}}Instruction\\ Acc (\%)\end{tabular}} & \textbf{\begin{tabular}[c]{@{}c@{}}Constraint\\ Acc (\%)\end{tabular}} & \multicolumn{1}{c}{\cellcolor[HTML]{B7B7B7}\textbf{\begin{tabular}[c]{@{}c@{}}OQA (\%)\\ Win / Lose\end{tabular}}} & \textbf{\begin{tabular}[c]{@{}c@{}}Time\\ (h)\end{tabular}} \\ \specialrule{.1em}{.02em}{.02em}
\rowcolor[HTML]{A2C4C9} 
GPT-4 &
  - &
  - &
  - &
  78.8 &
  91.9 &
  - &
  - &
  - &
  79.3\textsuperscript{\S} &
  85.4\textsuperscript{\S} &
  - &
  - \\ \hline
\rowcolor[HTML]{B4A7D6} 
Conv. &
  - &
  - &
  - &
  75.2 &
  87.8 &
  - &
  2.5 &
  - &
  60.1 &
  66.3 &
  - &
  2.9  \\
\rowcolor[HTML]{B4A7D6} 
Make sure &
  - &
  - &
  - &
  76.8 &
  88.6 &
  - &
  2.5 &
  - &
  60.1 &
  67.2 &
  - &
  3.6 \\
\rowcolor[HTML]{B4A7D6} 
Self-Refine &
  - &
  - &
  2 &
  77.2 ($\bm{\uparrow}$ 0.4) &
  88.7 ($\bm{\uparrow}$ 0.1) &
  22.1 / 21.2 &
  8.6 &
  2 &
  59.5 ($\bm{\downarrow}$ 0.6) &
  66.4 ($\bm{\downarrow}$ 0.8) &
  21.3 / 20.8 &
  4.5 \\ \hline
\rowcolor[HTML]{9FC5E8} 
\cellcolor[HTML]{9FC5E8} & Self &
  Self &
  6 &
  75.2 ($\bm{\downarrow}$ 1.6) &
  88.9 ($\bm{\uparrow}$ 0.3) &
  36.7 / 22.4 &
  11.2 &
  4 &
  60.1 (0.0) &
  \textbf{67.5 ($\bm{\uparrow}$ 0.3)} &
  17.1 / 30.6 &
  5.3 \\
\rowcolor[HTML]{9FC5E8} 
\cellcolor[HTML]{9FC5E8} &
  \textbf{Self} &
  \textbf{Supervised} &
  \textbf{10} &
  \textbf{80.5 ($\bm{\uparrow}$ 3.7)} &
  \textbf{90.9 ($\bm{\uparrow}$ 2.3)} &
  \textbf{37.1 / 22.0} &
  \textbf{6.9} &
  \textbf{10} &
  \textbf{60.8 ($\bm{\uparrow}$ 0.7)} &
  67.3 ($\bm{\uparrow}$ 0.1) &
  17.1 / 29.5 &
  5.7 \\ \cline{2-13} 
\rowcolor[HTML]{B6D7A8} 
\cellcolor[HTML]{9FC5E8} &
  Oracle\textsuperscript{\textdagger} &
  Self &
  4 &
  78.5 ($\bm{\uparrow}$ 1.7) &
  90.2 ($\bm{\uparrow}$ 1.6) &
  24.0 / 24.0 &
  6.1 &
  6 &
  62.3 ($\bm{\uparrow}$ 2.2) &
  69.1 ($\bm{\uparrow}$ 1.9) &
  17.6 / 28.1 &
  5.9 \\
\rowcolor[HTML]{B6D7A8} 
\cellcolor[HTML]{9FC5E8}&
  Oracle\textsuperscript{\textdagger} &
  Supervised &
  \textbf{10} &
  \textbf{82.4 ($\bm{\uparrow}$ 5.6)} &
  \textbf{91.7 ($\bm{\uparrow}$ 3.1)} &
  \textbf{34.3 / 22.2} &
  \textbf{5.6} &
  10 &
  64.9 ($\bm{\uparrow}$ 4.8) &
  71.6 ($\bm{\uparrow}$ 4.4) &
  18.2 / 30.9 &
  6.2 \\
\rowcolor[HTML]{B6D7A8} 
\cellcolor[HTML]{9FC5E8} &
  Oracle\textsuperscript{\textdagger} &
  GPT-4 &
  - &
  - &
  - &
  - &
  - &
  \textbf{4} &
  \textbf{68.2 ($\bm{\uparrow}$ 8.1)} &
  \textbf{74.1 ($\bm{\uparrow}$ 6.9)} &
  \textbf{13.8 / 34.6} &
  \textbf{6.7} \\ \cline{2-13}  
\multirow{-6}{*}{\cellcolor[HTML]{9FC5E8}\begin{tabular}[c]{@{}c@{}}\textbf{\dcr}\\ \textbf{(ours)}\end{tabular}} &
\cellcolor[HTML]{FFE599}\textbf{Oracle\textsuperscript{\textdagger}} &
\cellcolor[HTML]{FFE599}\textbf{Oracle\textsuperscript{\textdaggerdbl}} &
\cellcolor[HTML]{FFE599}\textbf{10} &
\cellcolor[HTML]{FFE599}\textbf{93.7 ($\bm{\uparrow}$ 16.9)} &
\cellcolor[HTML]{FFE599}\textbf{95.2 ($\bm{\uparrow}$ 6.6)} &
\cellcolor[HTML]{FFE599}\textbf{33.3 / 22.2} &
\cellcolor[HTML]{FFE599}\textbf{8.5} &
\cellcolor[HTML]{FFE599}\textbf{8} &
\cellcolor[HTML]{FFE599}\textbf{80.4 ($\bm{\uparrow}$ 20.3)} &
\cellcolor[HTML]{FFE599}\textbf{83.5 ($\bm{\uparrow}$ 16.3)} &
\cellcolor[HTML]{FFE599}\textbf{20.4 / 30.6} &
\cellcolor[HTML]{FFE599}\textbf{8.9} \\
\specialrule{.1em}{.05em}{.05em}
\end{tabular}
}
\resizebox{0.98\textwidth}{!}{
\begin{tabular}{
>{\columncolor[HTML]{A2C4C9}}l l
>{\columncolor[HTML]{B4A7D6}}l l
>{\columncolor[HTML]{9FC5E8}}l l
>{\columncolor[HTML]{B6D7A8}}l l
>{\columncolor[HTML]{FFE599}}l l}
\textbf{} & Proprietary Model &  & Baselines &  & Fairly Comparable &  & Realistic Ablation &  & Unrealistic ablation (upper bound)
\end{tabular}%
}
\caption{Results of the best iteration on \realinstruct and IFEval benchmarks for \dcr pipeline. Except for the first line, all results use Mistral v0.2. Absolute improvements from \textit{Make Sure} baseline are shown in (), with the best result in bold for each scenario. \textsuperscript{\textdagger}Oracle decomposer refers to human-verified constraint annotations provided with the datasets. \textsuperscript{\textdaggerdbl}Oracle feedback is GPT-4-Turbo on \realinstruct and a lenient rule-based evaluation on IFEVal. \textsuperscript{\S}Results reported by \citet{zhou2023instruction}. Reported time considers only generation time for fair comparison.}
\label{tab:selfref_realinstruct}
\vspace{-12pt}
\end{table*}

%% file: latex/Tables/tab2_benchmark_results.tex
\begin{table}[!h]
\centering
\small
\resizebox{\columnwidth}{!}{%
\begin{tabular}{ccc}
    \specialrule{.1em}{.05em}{.05em} 
    \rowcolor[HTML]{B7B7B7} 
    \textbf{Model} & \textbf{\begin{tabular}[c]{@{}c@{}}Instruction-level\\ Accuracy\end{tabular}} & \textbf{\begin{tabular}[c]{@{}c@{}}Constraint-level\\ Accuracy\end{tabular}} \\ 
    \specialrule{.07em}{.05em}{.3em} 
    \textbf{GPT-4} & \textbf{78.8\%} & \textbf{91.9\%} \\
    GPT-3.5        & 73.8\%          & 84.0\% \\ 
    \midrule
    \textbf{Mistral 7B v0.2} & \textbf{75.2\%} & \textbf{87.8\%} \\
    Zephyr 7B $\beta$           & 70.5\%          & 84.7\% \\
    Vicuna 7B v1.3            & 61.3\%          & 77.8\% \\
    \specialrule{.1em}{.05em}{.05em} 
\end{tabular}%
}
\caption{LLMs results on \realinstruct}
\label{tab:realinstructbench}
\vspace{-13pt}
\end{table}

%% file: latex/sections/conclusion.tex
In this work, we benchmarked LLMs' ability to follow real multi-constrained user requests with \realinstructraw. We showed that even strong proprietary models like GPT-4 fail to meet at least one constraint in over 21\% of instructions, demonstrating \realinstructraw’s challenging nature and the need for improvement across both proprietary and open-source models. To address this, we proposed the \dcr self-correction pipeline, which decomposes instructions into granular requirements, critiques responses, and refines outputs. Extensive experiments showed that \dcr significantly improves open-source LLM performance, with stronger feedback allowing them to surpass GPT-4. Overall, our work highlights the underexplored problem of following real-world user requests, as well as advances System 2 techniques with \dcrraw. Future work could refine the \dcrraw’s components and integrate the pipeline to other System 2 approaches, such as self-consistency and generate-and-rank, to enhance its effectiveness in tasks where spending more time on generation-refinement iterations would improve performance.

%% file: latex/sections/limitation.tex
\paragraph{Model-based vs. Rule-based Evaluation.} Using model-based evaluation over rule-based introduces two key challenges. First, it reflects the precision/accuracy trade-off, where rule-based methods offer higher precision but are less accurate due to synthetic scenarios, while model-based ones, though less precise, better align with real-world tasks (see discussion on Section \ref{related_work:model_based_instruction}). Second, evaluating the \realinstruct dataset relies on the proprietary GPT-4-Turbo API, making it costly. To address these, techniques like multi-prompting \citep{multiprompt2024} and panel of juries \citep{verga2024replacing} can improve precision in model-based evaluation, including with open-source models, while future advancements in open-source LLMs may provide cost-effective evaluation alternatives.

\paragraphshort{Data Contamination and Intra-Model Scoring Bias.} Developing benchmarks with publicly available data that does not overlap with LLM training data is challenging, as pre-training and instruction-tuning datasets are often undisclosed. For example, reliable information on Mistral's training data is unavailable. However, while Vicuna v1.3 reported instruction tuning on a dataset overlapping with \realinstructraw, no significant intra-model bias from data contamination was observed, as seen in its poor performance in Table \ref{tab:realinstructbench}. This is likely due to its instruction tuning procedure not ensuring constraint satisfaction in target responses. However, GPT-4’s relatively high constraint-level accuracy could indicate scoring bias, as previous studies suggest GPT-4 tends to favor its own outputs \citep{zheng2024judging,panickssery2024llm,verga2024replacing}. Further investigation into data contamination and intra-model scoring bias is left as future work.

\paragraphshort{Computation Overhead.}The \dcr pipeline introduces additional computational time compared to single-pass generation. Table \ref{tab:selfref_realinstruct} provides the running time for each configuration explored. \textbf{To mitigate this, we have designed the pipeline to trigger the refinement step only when the Critic model detects unsatisfied constraints}, which occured in about 25\% of instructions, minimizing unnecessary computation when the model performs well initially. This is an improvement upon other \textit{System 2} approaches (see Section \ref{related_work:system2}), such as generate-and-rank, which typically generate multiple outputs for further ranking. Additionally, we observed that \textbf{most revisions occur in the first iteration, resulting in a sublinear time increase with more iterations}. Also, \textbf{higher-quality feedback further reduces the need for revisions}, improving \dcrraw’s efficiency.

\paragraphshort{Optimization Considerations.} Due to high computational costs, we did not optimize hyperparameters for training the weakly supervised LLM-as-a-Judge or exhaustively tune the prompts for the adaptation strategies. Additionally, we did not explore using a dedicated LLM as a Decomposer in the \dcr pipeline, as this is primarily an implementation-focused task, being not critical for demonstrating our core claims. These aspects are left for future work.

%% file: latex/sections/ethical.tex
\paragraph{Crowdsourcing.} For the \textit{EvalJudge} Human Annotation task, we recruited native English speakers through Amazon Mechanical Turk\footnote{Refer to: \url{https://www.mturk.com/}} (MTurk). Compensation was based on the number of constraints per instruction, with an estimated average payment of 16.90 USD per hour, which exceeds the highest U.S. minimum wage in 2024 (16.30 USD per hour in Washington State), aligning with ethical guidelines discussed by \citet{huang-etal-2023-incorporating}.

\paragraph{Data from real users.} Constructing a dataset from real user requests presents some ethical challenges:
\begin{itemize}
    \vspace{-2pt}
    \item \textbf{Personally Identifiable Information (PII):} Some user interactions with AI assistants may contain PII. During data validation, we actively sought to remove instances containing PII from the dataset. See Appendix \ref{subsec:appendix:data_collection_validation} for further details.
    \vspace{-7pt}
    \item \textbf{Harmful Content:} The underlying data source is uncensored, and users may produce or request toxic or harmful content. Apart from flagrant cases, we did not actively remove such instances from the dataset.
\end{itemize}

\paragraph{Societal Impact.} The \dcr pipeline improves LLMs' ability to follow user-requested constraints, contributing to a broader societal impact of advancing LLM capabilities. When it comes particularly to user requests, it is important to note that some user constraints may conflict with system constraints set by developers, such as requests to generate harmful or toxic content. Although our study does not look into conflicting constraints, there is a potential risk that the pipeline could prioritize user requests over developer-defined safeguards.

%% file: latex/sections/related_work.tex
This section situates our work within broader research directions, highlighting intersections with current studies. Section \ref{related_work:benchmark} focuses on benchmarking and evaluating LLMs' generative abilities, while Section \ref{related_work:self_correction} discusses approaches for enhancing LLM responses.

\subsection{Evaluating LLMs' Generative Abilities}
\label{related_work:benchmark}

Traditional language model benchmarks, such as HellaSwag \citep{zellers-etal-2019-hellaswag}, WinoGrande \citep{winogrande}, MMLU \citep{hendrycks2021measuring}, GSM-8K \citep{gsm8k}, and BIG-Bench \citep{srivastava2023beyond}, primarily assess LLMs on tasks like commonsense reasoning and standardized exams. These benchmarks evaluate models using multiple-choice questions (MCQs) to objectively measure internal reasoning capabilities. However, recent advancements in language models have demonstrated emergent capabilities in generating high-quality open-ended text generation \citep{wei2021finetuned,flant5,ouyang2022training,taylor2022galactica,bubeck2023sparks}. 

This shift presents new challenges, as the number of possible responses are virtually infinite, requiring more subjective evaluation rather than strict reference matching. While MCQ-based benchmarks fall short in assessing these generative abilities, human annotation, though reliable, is limited by cost and scalability. To address this, some research directions sacrifice the question quality to be able to use rule-based evaluation methods, while others explore model-based approaches.

\subsubsection{LLM-as-a-judge}

Early model-based efforts like BERTScore \citep{Zhang2020BERTScore} sought to improve on traditional n-gram metrics by recognizing high-quality responses that differ from the reference. For more open-ended generation, where references are soft or nonexistent, recent work has introduced the concept of \textbf{LLM-as-a-Judge} \citep{zheng2024judging,geval2023}, using strong proprietary LLMs like GPT-4 \citep{achiam2023gpt} to evaluate responses. These models have shown they can approximate the depth and consistency of manual human evaluation, but also provide better consistency and stability.

Recent research has begun exploring open-source LLMs for LLM-as-a-Judge, aiming to reduce reliance on proprietary models. Although open-source models have shown limited capability with in-context learning, fine-tuning them for specific evaluations is a promising direction \citep{huang2024empirical,kim2024prometheus}. Contemporaneous work Prometheus-2, an open LLM-as-a-Judge, has shown strong correlation with human evaluation, even surpassing GPT-4 in some cases, though it still lags in out-of-domain cases \citep{kim2024prometheus2}. In our work, we assess both proprietary and open-source models for evaluating user constraint satisfaction in LLM responses. Our results indicate that while proprietary models outperform, open models can improve significantly when weakly supervised with proprietary model evaluations and reasoning trails, making them viable as Critic models in a self-correction pipeline.

Another recent approach by \citet{verga2024replacing} proposes replacing individual judges with juries of cheaper LLMs, which has been shown to correlate better with human judgments, even outperforming GPT-4 in some scenarios and reducing intra-model evaluation bias. This suggests that exploring LLM panels for constraint satisfaction evaluation could be a fruitful direction for future work.

\subsubsection{Fine-grained evaluation}

Some studies have explored approaches inspired by the divide-and-conquer paradigm, breaking multifaceted tasks into fine-grained components \citep{lee-kim-2023-recursion}. This can be successful given a complex compositional characteristic of the Natural Language \citep{manino-etal-2022-systematicity,dankers-etal-2022-paradox,zhong-etal-2024-benchmarking}. For evaluation, this strategy not only provides detailed insights into model performance across different aspects but also makes evaluations more objective and less ambiguous, as models may excel in some areas while underperforming in others. This approach seems promising for LLM-as-a-Judge, given that LLMs prompting techniques such as Chain-of-Thought \citep{cot2022neurips}, Tree-of-Thought \citep{treeofthoughts2023}, and Recursive Thinking \citep{qi-etal-2023-art}, have demonstrated LLM performance improvements by breaking complex tasks into simpler sequential steps. We discuss these methods on Section \ref{related_work:system2}. 

Specific works on evaluation by \citet{min-etal-2023-factscore,li-etal-2023-evaluating,faithscore,Hu_2023_ICCV,song-etal-2024-finesure,huang2024empirical,zhang2024towards} have shown the benefits of decomposing tasks into atomic facts for tasks such as fact-checking against cross-modality references. \citet{kim2024prometheus,magister-etal-2023-teaching,ke-etal-2024-critiquellm} have demonstrated that fine-grained evaluation from diverse sources enhances fine-tuned open evaluators by making the task more objective. Additionally, weak fine-grained evaluation during generation time has been shown to improve LLM self-correction performance \citep{shridhar2023screws,shridhar2023art,wang2024speculative}.

In our work, we implement a similar approach by decomposing the task of evaluating multi-constrained instructions into individual constraint evaluations. This "instruction decomposition" simplifies and makes more objective the instruction evaluation task for LLMs and provides more informative insights through constraint-level accuracy metrics. We also argue that existing overall instruction satisfaction metrics fail to detect unmet constraints due to the ambiguity caused by the lack of granularity, as also highlighted by \citet{sun-etal-2023-evaluating}. Our results demonstrate the effectiveness of fine-grained evaluation for this task and show that incorporating it into a self-correction pipeline enhances performance, even with weak Critic and Decomposer models.

\subsubsection{Benchmarking Instruction-Following Abilities}
\label{related_work:model_based_instruction}

The ability of LLMs to follow user instructions in open-ended text generation has only recently gained attention. New benchmarks like AlpacaEval \citep{alpaca_eval} and the test splits of Natural-Instructions \citep{mishra-etal-2022-cross} and Self-Instruct \citep{wang-etal-2023-self-instruct} address the evaluation in this aspect by using LLM-as-a-Judge to compare with reference responses or provide overall instruction satisfaction scores. Recent studies have shown that models often follow instructions only partially, frequently failing to adhere to specific constraints provided by users \citep{sun-etal-2023-evaluating,zhou2023instruction,yao2024collie,jiang2023followbench,Qin2024InFoBenchEI,wen2024benchmarking,he2024can,zhang2024cfbench}.

To evaluate this, the few existing benchmarks focus on a set of specific constraint categories and/or use synthetic constraints that can be easily verified through rule-based methods \citep{zhou2023instruction,yao2024collie}. The trade-off between rule-based and model-based evaluation falls into the famous precision/accuracy dilemma about static instrument characteristics (sometimes referred as bias/variance dilemma) in the Statistics of measurements \citep{morris2001measurement,taylor1997introduction,british2022bs}. Rule-based evaluation offers high-to-perfect precision (low variance), but it is usually required to be done on unrealistic scenarios, being less accurate (high bias). The use of model-based evaluation loses some precision compared to rule-based due to inherent variability introduced by LLMs (lower precision, higher variance), but aligns more with the task objective of evaluating more realistic scenarios (higher accuracy, lower bias). 

To the best of our knowledge, our \realinstruct benchmark is the first to evaluate LLMs using real-user instructions, offering a more realistic and comprehensive assessment. This approach closely mirrors real-world scenarios, unlike previous benchmarks that rely on synthetic constraints, as contrasted on Table \ref{tab:other_benchmarks} and Section \ref{sec:realinstruct}. Our benchmark’s success relies on a fine-grained evaluation protocol using LLM-as-a-Judge.

\subsection{LLM Self-Correction for open-ended text generation}
\label{related_work:self_correction}
Self-correction has emerged as an effective approach for enhancing LLM responses during generation by refining them during generation time \citep{pan2023automatically,selfcorrectionsurvey2024}. However, \citet{selfcorrectionsurvey2024,Huang2023LargeLM} demonstrated that the ability of LLM to self-correct alone is limited to tasks where responses can be decomposed and rely on verifiable components. For harder tasks, LLM self-correction may require additional modeling, new data, or even external tools.

In this sense, Self-correction approaches can be categorized based on the feedback source. \textbf{Intrinsic Self-Correction} uses carefully crafted prompts or in-context examples to enable the model to identify issues in its output. \textbf{Self-Correction with External Feedback} leverages external tools or more advanced LLMs to provide feedback, while \textbf{Self-Correction with Fine-Tuning} uses external feedback (from humans, stronger LLMs, external tools) to fine-tune the LLM for better feedback and/or response refinement. \citet{selfcorrectionsurvey2024} emphasizes that each self-correction category should be validated using comparable cases specific to its context.

In the case of multi-constrained instructions, the constraints are neither independent nor ordered, making it difficult to guarantee that all responses are decomposable. For example, constraints such as length and style do not have a specific part of the response to be followed, they should be followed in the whole text. Moreover, some constraints are subjective and harder to evaluate, and the instruction decomposition process may introduce noise, further complicating self-correction with the model itself. In our work, we explore both Intrinsic Self-Correction and Self-Correction with Critic Fine-Tuning. As ablation exploration, we also play with External Feedback (referred as Oracle Critic), as an estimation of upper bound performance but recognizing its limited generalization. 

\subsubsection{Constrained Generation}
\input{latex/Tables/table_related_work}

Recent work has explored constrained generation via self-correction, we show some representative work on Table \ref{tab:related_work}. Some approaches are validated only on small, specific constraint sets, limiting their general applicability \citep{schick2023peer,saunders2022self,volcano,mu2023can,yang-etal-2022-re3,gou2024critic,varshney2023stitch}. For example, \citet{mu2023can} evaluates 13 System constraints, that is constraints defined by the developer. Hallucination and Detoxification constraints can also be seen as System Constraints. 

In another direction, \citet{Selfrefine2023} enhances model outputs by having the model self-review and self-correct its answers. However, this approach is suboptimal as it lacks problem-specific modeling, that is it does not recognize the constraints to be followed, and only prompts the model to evaluate and improve its responses without clear guidance on what to focus on. \citet{selfee2023} goes in a similar direction but relying on supervising LLM to be able to refine. Our results in Sections \ref{subsec:CEBBenchmark} and \ref{subsec:DCR_subsection} demonstrate improved performance by directing the model to specifically address constraints, even when it is refining its responses using only its own feedback. To the best of our knowledge we are the first to handle open-ended user instructions without restricting it to some constraint types. We highlight the prevalence of constraints in real-world user instructions, making this an important area for further study. Also, unlike most methods, our \dcr pipeline does not require external tools or fine-tuning LLM for refinement.

It is worth noting that most current constrained generation with self-correction research evaluates on the CommonGen benchmark \citep{lin-etal-2020-commongen}, in which constraints are word lists for LLM to include in the text. We argue that this benchmark is insufficient for measuring the ability to follow user requests because: (1) models like GPT-4 already perform at human level\footnote{See CommonGen leaderboard at: \url{https://github.com/allenai/CommonGen-Eval}}; (2) it only represents one type of constraint among many possible; and (3) the constraints are synthetic and not reflective of typical human requests. While valuable in the past for large-scale rule-based evaluation, it may not adequately measure modern LLM capabilities.

\subsubsection{System 2 Approaches}
\label{related_work:system2}

\citet{selfcorrectionsurvey2024} differentiate self-correction from other methods like self-consistency \citep{wang2023selfconsistency,chen2024universal,yao2024learning}, which samples diverse reasoning paths during decoding and selects the most consistent, and generate-then-rank methods like Tree of Thoughts \citep{treeofthoughts2023}, which generates multiple responses and ranks them using a critic model. These two approaches do not directly refine responses and assume that LLMs can generate at least one correct initial response with a considerable probability, which is not always the case.

These two approaches, like Self-Correction, belong to a broader category known as \textbf{System 2 approaches}, which includes all techniques that generate intermediary outputs before producing final response, aiming to improve LLM responses during generation or inference by emulating the idea of planning. For instance, \citet{khan2024debating} and \citet{du2024improving} introduced LLM debating, where each LLM initially provides a solution and then revises it based on combined responses, eventually leading to a shared solution after several rounds. Another notable System 2 approach is Branch-Solve-Merge \citep{saha2023branch}, which tackles instructions in parts and then merges the results. This has been used for constrained generation, but assumes constraints are independent and merging responses satisfying subsets of constraints address all constraints, which makes it not applicable to real-world constraints.

A key issue with System 2 methods is the increased inference time that naturally comes with the generation of intermediary outputs. Our \dcr pipeline mitigates this by avoiding unnecessary revisions when the LLM already performs well according to the Critic model, which is an improvement over existing System 2 approaches.

But, this increased inference time is worthwhile. \citet{chen2024more} shows that more LLM calls can enhance performance in tasks where LLMs are capable, though it may degrade performance on task that are yet challenging for them. This suggests that System 2 approaches can push LLM limits and help us understand more what they are capable of. Additionally, these techniques can generate data to improve and generalize existing models. For example, \citet{deng2024implicit,yu2024distilling} demonstrated that System 2 approaches can be used in a self-supervised learning distillation setting to enhance the original LLMs ("System 1"), resulting in reduced inference costs and improved performance. This is an interesting direction for future work, as an extension of our \dcr pipeline.

%% file: latex/Tables/table_related_work.tex
\begin{table}[]
\centering
\resizebox{0.9\columnwidth}{!}{%
\begin{tabular}{ccccc}
\specialrule{.1em}{.05em}{.05em}
\rowcolor[HTML]{B7B7B7} 
\multicolumn{1}{c|}{\cellcolor[HTML]{B7B7B7}\textbf{Method}} &
  \textbf{Feedback Source} &
  \multicolumn{1}{c|}{\cellcolor[HTML]{B7B7B7}\textbf{Refinement Strategy}} &
  \textbf{Tasks Investigated} &
  \textbf{Supported Constraint Types} \\ \specialrule{.1em}{.05em}{.05em}
\multicolumn{5}{c}{\textbf{Self-correction with training for refining}} \\ \midrule
\multicolumn{1}{c|}{\begin{tabular}[c]{@{}c@{}}Selfee \citep{selfee2023}\end{tabular}} &
  LLM + ICL &
  \multicolumn{1}{c|}{LLM SFT} &
  Open-ended Instructions &
  Not constraint focus \\ 
\multicolumn{1}{c|}{\begin{tabular}[c]{@{}c@{}}PEER  \citep{schick2023peer}\end{tabular}} &
  LLM SFT &
  \multicolumn{1}{c|}{LLM SFT} &
  Constrained Generation &
  Limited constraints \\ 
\multicolumn{1}{c|}{\begin{tabular}[c]{@{}c@{}}Self-Critique \citep{saunders2022self},\\ VOLCANO \citep{volcano}\end{tabular}} &
  \begin{tabular}[c]{@{}c@{}}LLM SFT \\ w/ Human Feedback\end{tabular} &
  \multicolumn{1}{c|}{\begin{tabular}[c]{@{}c@{}}LLM SFT \\ w/ Human Feedback\end{tabular}} &
  \begin{tabular}[c]{@{}c@{}}Conditional Summarization, \\ Visual Question-Answering\end{tabular} &
  Limited constraints \\
\multicolumn{1}{c|}{\begin{tabular}[c]{@{}c@{}}InstructScore \\ \citep{xu-etal-2023-instructscore}\end{tabular}} &
  \begin{tabular}[c]{@{}c@{}}LLM SFT \\ w/ GPT-4 and \\ Human Feedback\end{tabular} &
  \multicolumn{1}{c|}{\begin{tabular}[c]{@{}c@{}}LLM SFT \\ w/ GPT-4\end{tabular}} &
  CommonGen &
  Limited constraints \\
\multicolumn{1}{c|}{\begin{tabular}[c]{@{}c@{}}Self-Correctors \citep{welleck2023generating}\end{tabular}} &
  External tools &
  \multicolumn{1}{c|}{\begin{tabular}[c]{@{}c@{}}Smaller \\ LLM STF\end{tabular}} &
  \begin{tabular}[c]{@{}c@{}}CommonGen, \\ Detoxification\end{tabular} &
  Limited constraints \\
\multicolumn{1}{c|}{RULES \citep{mu2023can}} &
  External tools &
  \multicolumn{1}{c|}{LLM SFT} &
  \begin{tabular}[c]{@{}c@{}}System \\ Constraints\end{tabular} &
  Limited constraints \\
\multicolumn{1}{c|}{Re3 \citep{yang-etal-2022-re3}} &
  \begin{tabular}[c]{@{}c@{}}Smaller LLM SFT \\ + External Tools\end{tabular} &
  \multicolumn{1}{c|}{\begin{tabular}[c]{@{}c@{}}LLM + \\ smaller model\end{tabular}} &
  Story Generation &
  Limited constraints \\ \midrule
\multicolumn{5}{c}{\textbf{Self-correction without training for refinement}} \\ \midrule
\multicolumn{1}{c|}{\begin{tabular}[c]{@{}c@{}}Self-Refine \citep{Selfrefine2023}\end{tabular}} &
  LLM + ICL &
  \multicolumn{1}{c|}{} &
  \begin{tabular}[c]{@{}c@{}}Open-ended Instructions; \\ CommonGen\end{tabular} &
  \begin{tabular}[c]{@{}c@{}}Not constraint focus; \\ Limited constraints\end{tabular} \\
\multicolumn{1}{c|}{\begin{tabular}[c]{@{}c@{}}CRITIC \citep{gou2024critic}, \\ Hallucination \citep{varshney2023stitch}\end{tabular}} &
  External tools &
  \multicolumn{1}{c|}{} &
  \begin{tabular}[c]{@{}c@{}}Detoxification, \\ Hallucination\end{tabular} &
  Limited constraints \\
\multicolumn{1}{c|}{\begin{tabular}[c]{@{}c@{}}\dcr (ours)\end{tabular}} &
  \begin{tabular}[c]{@{}c@{}}LLM SFT \\ w/ GPT-4\end{tabular} &
  \multicolumn{1}{c|}{\multirow{-5}{*}{LLM + ICL}} &
  Open-ended Instructions &
  Any constraint \\ \specialrule{.1em}{.05em}{.05em}
\end{tabular}%
}
\caption{Comparison of representative works on Self-Correction for Constrained Generation. Our \dcr pipeline is unique as do not require LLM fine-tuning for refinement, being also the only that can handle open-ended instructions with any type of constraints.}
\vspace{-15pt}
\label{tab:related_work}
\end{table}

%% file: latex/appendix/appendix_extra_analysis.tex
In this section we present extra analysis and discussions about \dcr pipeline.

\subsection{\dcr Experiments with other Open LLMs}
\input{latex/Tables/appendix_extra_table_vicuna_zephyr}

We repeated the same experiments from Section \ref{subsec:experiments_decrim} with Vicuna v1.3 and Zephyr $\beta$, using the Oracle Decomposer and our Weakly Supervised Mistral as the Critic. Results are presented on Table \ref{tab:extra_open_LLM}. Initial performance for both models on the \realinstruct and IFEval benchmarks was low, but the \dcr pipeline led to significant improvements across all scenarios except one. With the best possible feedback, all models beat proprietary GPT-4 on \realinstructraw. Notably, event weak-performing LLM Vicuna v1.3 achieved a 34.1\% improvement on \realinstruct (a 59\% relative increase). 

The exception was the Vicuna v1.3 Oracle-Supervised setting for \realinstructraw, where we observed some degradation. As discussed in Section \ref{subsec:DCR_subsection}, weak feedback can cause over-refinement of good responses while failing to fix bad ones, negatively affecting overall gains.

\subsection{Comparing \dcr with Generate-and-Rank}
\label{subappendix:generate_rank}

\paragraph{Generate-and-Rank approach} We compare the performance of our \dcr self-correction pipeline with \textbf{Generate-and-Rank}, an intuitive System 2 approach. In this pipeline, rather than refining the response as in \dcrraw, it samples multiple candidate generations (one at each iteration) with different parameters and selects the best one based on feedback from the Critic model. The pipeline iterates over a predefined parameter search path, generating responses until all constraints are satisfied or the iteration limit ($N_{max}$) is reached. To streamline the process, we use the \dcr pipeline with the difference that instead of refine the response, LLM generate a new response. For that, we convert Critic feedback to binary instruction-level feedback indicating whether the current response satisfies all constraints or if a new response should be generated.

\input{latex/figures/generate_and_rank}
\paragraph{Setup} Generate-and-Rank relies on varying generation parameters to produce new responses. We focus on three key parameters: generation prompt, sampling or greedy decoding, and temperature (when sampling). For the generation prompt, we propose a \textbf{Decompose-then-Generate} (DtG) prompt (detailed in Figure \ref{fig:decompose_then_generate_prompt}), inspired by System 2 Attention \citep{weston2023system} and Rephrase and Respond \citep{deng2023rephrase}. This prompt decomposes multi-constrained instructions into an enumerated list of constraints and then generating a response. We set $N_{max} = 10$ and vary the parameters in order of creativity trade-offs, using the following sequence of tuples (Prompt, Sampling or not, temperature):

\noindent[(\textit{Make Sure}, Sampling, 0.2), (\textit{DtG}, Sampling, 0.2), (\textit{Make Sure}, Greedy, 1.0), (\textit{DtG}, Greedy, 1.0), (\textit{Make Sure}, Sampling, 1.0), (\textit{DtG}, Sampling, 1.0), (\textit{Make Sure}, Sampling, 0.5), (\textit{DtG}, Sampling, 0.5), (\textit{Make Sure}, Sampling, 0.7), (\textit{DtG}, Sampling, 0.7)].

\input{latex/Tables/appendix_generate_and_rank}

\paragraph{Results} Table \ref{tab:generate_and_rank} compares the results between the Generate-and-Rank and \dcr pipelines in different Decomposer and Critic configurations. Our findings show that while \dcr demonstrates consistent improvements across almost all scenarios, Generate-and-Rank always performs poorly when weak feedback is used. However, with strong feedback, Generate-and-Rank outperforms \dcrraw, surpassing GPT-4 by a larger margin. This highlights Generate-and-Rank’s reliance on high-quality feedback, whereas \dcr is more resilient to weak Critic models, delivering improvements across most of the scenarios. 
    
Interestingly, Generate-and-Rank achieves high instruction-level performance in both benchmarks with strongest feedback, suggesting that LLMs have the ability to follow the constraints in some of $n$ generations. This raise the hypothesis that LLMs not following user requests is a matter of alignment, which supports the idea discussed in Section \ref{related_work:system2}, that aligning LLMs with outputs from different System 2 approaches, such as \dcrraw, Generate-and-Rank, or a combination of both, can significantly improve the performance of System 1 models to follow multi-constrained instruction. This constitute a relevant direction of future work. 

%% file: latex/Tables/appendix_extra_table_vicuna_zephyr.tex
\begin{table}[!h]
\vspace{-5pt}
\centering
\resizebox{0.9\columnwidth}{!}{%
\begin{tabular}{lcc|ccc|ccc}
\specialrule{.1em}{.05em}{.05em}
\rowcolor[HTML]{9B9B9B} 
\multicolumn{1}{c}{\cellcolor[HTML]{9B9B9B}} &
  \cellcolor[HTML]{9B9B9B} &
  \cellcolor[HTML]{9B9B9B} &
  \multicolumn{3}{c|}{\cellcolor[HTML]{9B9B9B}\textbf{\realinstruct}} &
  \multicolumn{3}{c}{\cellcolor[HTML]{9B9B9B}\textbf{IFEval}} \\
\rowcolor[HTML]{9B9B9B} 
\multicolumn{1}{c}{\multirow{-2}{*}{\cellcolor[HTML]{9B9B9B}\textbf{Strategy}}} &
  \multirow{-2}{*}{\cellcolor[HTML]{9B9B9B}\textbf{Decomposer}} &
  \multirow{-2}{*}{\cellcolor[HTML]{9B9B9B}\textbf{Critic}} &
  {\color[HTML]{000000} \textbf{\begin{tabular}[c]{@{}c@{}}Best\\ N\end{tabular}}} &
  {\color[HTML]{000000} \textbf{\begin{tabular}[c]{@{}c@{}}Instruction\\ Acc (\%)\end{tabular}}} &
  {\color[HTML]{000000} \textbf{\begin{tabular}[c]{@{}c@{}}Constraint\\ Acc (\%)\end{tabular}}} &
  {\color[HTML]{000000} \textbf{\begin{tabular}[c]{@{}c@{}}Best\\ N\end{tabular}}} &
  {\color[HTML]{000000} \textbf{\begin{tabular}[c]{@{}c@{}}Instruction\\ Acc (\%)\end{tabular}}} &
  {\color[HTML]{000000} \textbf{\begin{tabular}[c]{@{}c@{}}Constraint\\ Acc (\%)\end{tabular}}} \\\specialrule{.1em}{.05em}{.05em}
GPT-4                       & -       & -          & -  & 78.8         & 91.9         & -  & 79.3\textsuperscript{\S}        & 85.4\textsuperscript{\S}        \\ \midrule
\textit{Mistral v0.2}&         &            &    &              &              &    &              &              \\
Make sure                   & -       & -          & -  & 76.8         & 88.6         & -  & 60.1         & 67.2         \\
\multirow{2}{*}{\begin{tabular}[c]{@{}c@{}}\dcr\\ (ours)\end{tabular}}  & Oracle\textsuperscript{\textdagger} & Supervised & 10 & 82.4 ($\bm{\uparrow}$5.6)  & 91.7 ($\bm{\uparrow}$3.1)  & 10 & 64.9 ($\bm{\uparrow}$4.8)  & 71.6 ($\bm{\uparrow}$4.4)  \\
                            & Oracle\textsuperscript{\textdagger} & Oracle\textsuperscript{\textdaggerdbl}  & 10 & 93.7 ($\bm{\uparrow}$16.9) & 95.2 ($\bm{\uparrow}$6.6)  & 8  & 80.4 ($\bm{\uparrow}$20.3) & 83.5 ($\bm{\uparrow}$16.3) \\ \midrule
\textit{Vicuna v1.3}                 &         &            &    &              &              &    &              &              \\
Make sure                   & -       & -          & -  & 57.6         & 77.4         & -  & 36.0         & 46.1         \\
\multirow{2}{*}{\begin{tabular}[c]{@{}c@{}}\dcr\\ (ours)\end{tabular}}  & Oracle\textsuperscript{\textdagger} & Supervised & 10 & 57.0 (↓0.6)  & 76.6 (↓0.8)  & 10 & 38.3 ($\bm{\uparrow}$2.3)  & 47.8 ($\bm{\uparrow}$1.7)  \\
                            & Oracle\textsuperscript{\textdagger} & Oracle\textsuperscript{\textdaggerdbl}  & 10 & 91.7 ($\bm{\uparrow}$34.1) & 92.3 ($\bm{\uparrow}$14.9) & 10 & 47.0 ($\bm{\uparrow}$11.0) & 54.4 ($\bm{\uparrow}$8.3)  \\ \midrule
\textit{Zephyr} $\beta$ &         &            &    &              &              &    &              &              \\
Make sure                   & -       & -          & -  & 69.5         & 84.7         & -  & 53.6         & 62.0         \\
\multirow{2}{*}{\begin{tabular}[c]{@{}c@{}}\dcr\\ (ours)\end{tabular}}  & Oracle\textsuperscript{\textdagger} & Supervised & 10 & 71.5 ($\bm{\uparrow}$2.0)  & 84.8 ($\bm{\uparrow}$0.1)  & 10 & 55.1 ($\bm{\uparrow}$1.5)  & 63.5 ($\bm{\uparrow}$1.5)  \\
                            & Oracle\textsuperscript{\textdagger} & Oracle\textsuperscript{\textdaggerdbl}  & 10 & 91.1 ($\bm{\uparrow}$21.6) & 92.5 ($\bm{\uparrow}$7.8)  & 10 & 74.5 ($\bm{\uparrow}$19.9) & 78.7 ($\bm{\uparrow}$16.7)\\
                            \specialrule{.1em}{.05em}{.05em} 
\end{tabular}%
}
\caption{Results of the best iteration on \realinstruct and IFEval benchmarks for \dcr pipeline for the 3 open LLMs (Mistral v0.2, Vicuna v1.3, Zephyr $\beta$). Absolute improvements from \textit{Make Sure} baselines are shown in (). \textsuperscript{\textdagger}Oracle decomposer refers to human-verified constraint annotations provided with the datasets. \textsuperscript{\textdaggerdbl}Oracle feedback is GPT-4-Turbo on \realinstruct and lenient rule-based evaluation on IFEVal. \textsuperscript{\S}Results reported by \citet{zhou2023instruction}.}
\vspace{-5pt}
\label{tab:extra_open_LLM}
\end{table}

%% file: latex/figures/generate_and_rank.tex
\begin{figure}[h!]
    \centering
    \includegraphics[width=0.6\linewidth]{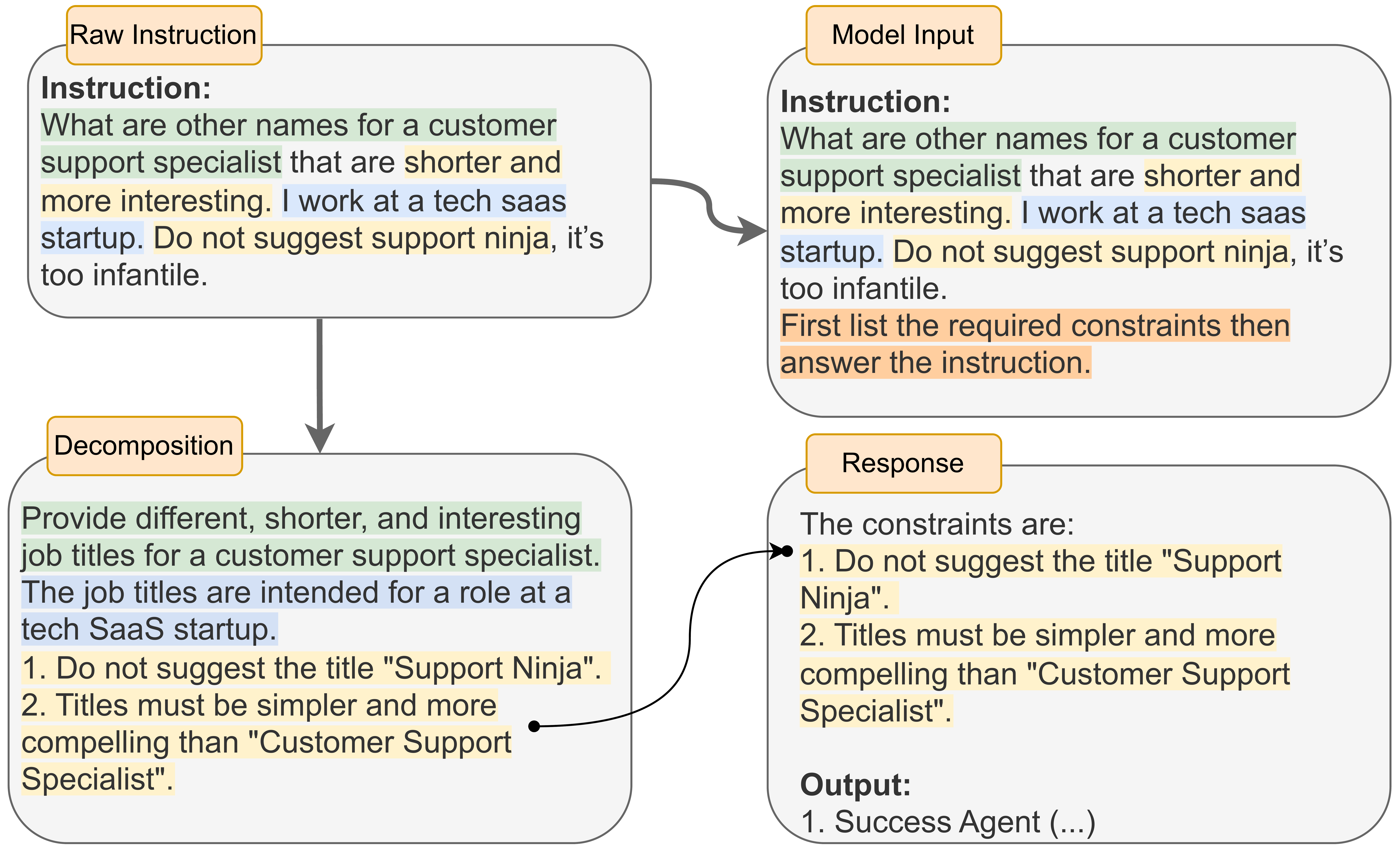} 
    \caption{
    \textbf{Two-step \textit{Decompose-then-Generate} (DtG) prompt:} Inspired by the two-step Rephrase and Respond (RaR) \citep{deng2023rephrase}, DtG first instructs the LLM to decompose multi-constrained instructions into an enumerated list of constraints. Then, DtG uses this decomposition as if it were the model's own "reasoning and planning" (leveraging the model's user and assistant tokens) to generate the final response. Like RaR, this process can be done in one or two steps, with the two-step method being more effective.}
    \label{fig:decompose_then_generate_prompt}
    \vspace{-5pt}
\end{figure}

%% file: latex/Tables/appendix_generate_and_rank.tex
\begin{table}[!h]
\centering
\resizebox{0.9\columnwidth}{!}{%
\begin{tabular}{ccc|ccc|ccc}
\specialrule{.1em}{.05em}{.05em}
\rowcolor[HTML]{B7B7B7} 
\multicolumn{1}{c}{\cellcolor[HTML]{B7B7B7}} &
  \cellcolor[HTML]{B7B7B7} &
  \cellcolor[HTML]{B7B7B7} &
  \multicolumn{3}{c|}{\cellcolor[HTML]{B7B7B7}\textbf{\realinstruct}} &
  \multicolumn{3}{c}{\cellcolor[HTML]{B7B7B7}\textbf{IFEval}} \\
\rowcolor[HTML]{B7B7B7} 
\multicolumn{1}{c}{\multirow{-2}{*}{\cellcolor[HTML]{B7B7B7}\textbf{Strategy}}} &
  \multirow{-2}{*}{\cellcolor[HTML]{B7B7B7}\textbf{Decomposer}} &
  \multirow{-2}{*}{\cellcolor[HTML]{B7B7B7}\textbf{Critic}} &
  \textbf{\begin{tabular}[c]{@{}c@{}}Best\\ N\end{tabular}} &
  \textbf{\begin{tabular}[c]{@{}c@{}}Instruction\\ Acc (\%)\end{tabular}} &
  \textbf{\begin{tabular}[c]{@{}c@{}}Constraint\\ Acc (\%)\end{tabular}} &
  \textbf{\begin{tabular}[c]{@{}c@{}}Best\\ N\end{tabular}} &
  \textbf{\begin{tabular}[c]{@{}c@{}}Instruction\\ Acc (\%)\end{tabular}} &
  \textbf{\begin{tabular}[c]{@{}c@{}}Constraint\\ Acc (\%)\end{tabular}} \\\specialrule{.1em}{.05em}{.05em}
GPT-4 &
  - &
  - &
  - &
  78.8 &
  91.9 &
  - &
  79.3\textsuperscript{\S} &
  85.4\textsuperscript{\S} \\ \midrule
Make sure &
  - &
  - &
  - &
  76.8 &
  88.6 &
  - &
  60.1 &
  67.2 \\ \midrule
 &
  Self &
  Self &
  6 &
  75.2 ($\bm{\downarrow}$1.6) &
  88.9 ($\bm{\uparrow}$0.3) &
  4 &
  60.1 (0.0) &
  67.5 ($\bm{\uparrow}$0.3) \\
 &
  Oracle\textsuperscript{\textdagger} &
  Self &
  4 &
  78.5 ($\bm{\uparrow}$1.7) &
  90.2 ($\bm{\uparrow}$1.6) &
  6 &
  62.3 ($\bm{\uparrow}$2.2) &
  69.1 ($\bm{\uparrow}$1.9) \\
\multirow{-3}{*}{\begin{tabular}[c]{@{}c@{}}\dcr\\(ours)\end{tabular}} &
  Oracle\textsuperscript{\textdagger} &
  Oracle\textsuperscript{\textdaggerdbl} &
  10 &
  93.7 ($\bm{\uparrow}$16.9) &
  95.2 ($\bm{\uparrow}$6.6) &
  8 &
  80.4 ($\bm{\uparrow}$20.3) &
  83.5 ($\bm{\uparrow}$16.3) \\ \midrule
 &
  Self &
  Self &
  7 &
  76.2 ($\bm{\downarrow}$0.6) &
  88.3 ($\bm{\downarrow}$0.3) &
  2 &
  59.9 ($\bm{\downarrow}$0.2) &
  66.9 ($\bm{\downarrow}$0.3) \\
 &
  Oracle\textsuperscript{\textdagger} &
  Self &
  2 &
  76.5 ($\bm{\downarrow}$0.3) &
  88.8 ($\bm{\uparrow}$0.2) &
  2 &
  59.5 ($\bm{\downarrow}$0.6) &
  66.4 ($\bm{\downarrow}$0.8) \\
\multirow{-3}{*}{\begin{tabular}[c]{@{}c@{}}Generate-\\ and-Rank\end{tabular}} &
  Oracle\textsuperscript{\textdagger} &
  Oracle\textsuperscript{\textdaggerdbl} &
  10 &
  92.8 ($\bm{\uparrow}$16.0) &
  96.5 ($\bm{\uparrow}$7.9) &
  10 &
  81.7 ($\bm{\uparrow}$21.6) &
  86.5 ($\bm{\uparrow}$19.3) \\ \specialrule{.1em}{.05em}{.05em}
\end{tabular}%
}
\caption{Results of the best iteration on \realinstruct and IFEval benchmarks for \dcr and \textit{Generate-and-Rank} pipelines using Mistral v0.2 as the LLM. Absolute improvements from \textit{Make Sure} baseline are shown in ().}
\vspace{-10pt}
\label{tab:generate_and_rank}
\end{table}

%% file: latex/appendix/appendix_definition_decomposition.tex
To support the choice of prompts and annotation guidelines in our work, we define the concepts of Task, Context, and Constraints within the domain of instruction-following for Large Language Models as follows:

\begin{itemize}
    \item \textbf{Task:} The primary objective that the language model is expected to achieve. The task defines the central goal that guides the generation of the desired output, outlining the specific action the model should perform. \\
    \textit{Example:} "Summarize the key findings of the given research paper."
    \item \textbf{Context:} The additional information or details that provide a foundation for the language model to better understand the task. The context helps the model by offering relevant facts, scenarios, or circumstances, thereby enhancing the quality and relevance of the output. The context may also refer to specific input data to be considered.\\
    \textit{Example:} If the task is to summarize a paper, the context would be the paper itself.
    \item \textbf{Constraints:} The specific conditions, limitations, or requirements imposed on the language model to shape the nature of the generated output. Constraints help control factors such as length, format, content, and style, ensuring the output meets defined criteria. In our decomposition process, constraints are expected to be written in an actionable and self-contained manner to make a model-based fine-grained evaluation possible.\\
    \textit{Example:}
    \begin{itemize}
        \item \textbf{Length:} "Generate a summary with a maximum of 150 words."
        \item \textbf{Content:} "Focus on the main contributions and findings of the research paper."
        \item \textbf{Style:} "Use a formal and concise writing style."
    \end{itemize}

\end{itemize}

We distinguish Task and Context because empirically we found that by separating them it simplifies the instruction decomposition task and improves the accuracy of the GPT-4 model for this task. However, they are intended to be used together. In the \realinstruct dataset and related experiments, Task and Context are presented as a combined \textit{``Task+Context''} input.

%% file: latex/appendix/appendix_data_construction.tex
\subsection{Data Filtering}
\label{subappendix:data_construction:data_filtering}
The first part of the dataset creation was the data filtering process. This included the following steps:
\begin{enumerate}
    \item \textbf{Remove Assistant Responses:} We removed GPT answers from the dataset to focus solely on human interactions.
    \item \textbf{Remove Non-English Conversations:} Using the \texttt{langdetect} package\footnote{Available at: \url{https://pypi.org/project/langdetect/}}, we classify the main language of conversations and discarded non-English threads.
    \item \textbf{Filter Out Code-Related Requests:} Relying on the open-source LLM Mistral 7B Instruct v0.1\footnote{Available at: \url{https://huggingface.co/mistralai/Mistral-7B-Instruct-v0.1}} as a two-shot classifier, we identify conversations involving code-related requests, utilizing Prompt \ref{prompt:code_classification}.
    \item \textbf{Retain Only the First Request:}To avoid the complexities of multi-turn scenarios and reduce computational demands, we retained only the initial user instruction, ensuring it was self-contained. 
    \item \textbf{Retrive Instructions with Constraints:} Again employing Mistral 7B Instruct v0.1, this time in a 5-shot classification approach, we identified instructions containing constraints, using Prompt \ref{prompt:constrained_classification}. 
    \item \textbf{Human Validation:} The authors of this work manually validated the filtered instructions to eliminate unsafe content and ensure relevance and clarity, following the guidelines outlined in \ref{subsec:appendix:data_collection_validation}.  
\end{enumerate}

Steps 1 through 5 were applied to the entire dataset. Human validation (step 6) was conducted only on a subset due to resource limitations. Notably, 44\% of English, non-code requests were found to contain constraints during the automated filtering in step 5. In the subset subjected to human validation, 30\% contained constraints according to the auditors. These figures underscore the relevance of addressing instructions with multiple constraints in real user interactions.

\subsubsection{Prompt for two-shot classification of code-related conversations}

\begin{mdframed}[linecolor=purple,linewidth=0.5pt,nobreak=false,backgroundcolor=purple!10]
\small{
You are an assistant whose job is to help me perform tasks. I need to filter from a set of dialogues between users and AI assistants, the ones in which human requested something related to code. I will give you all the human part of the dialog and I expect you to answer ``Yes" when the dialog contains instructions asking the assistant something about code, or ``No" if the dialog does not contemplate any code-related request. You are provided two examples.
\\
\\
\textbf{Example 1:}\\
Human: I have 100 dollars and would like to use this as the initial funding to make some money. I need it to be as quick as possible with good returns.\\
Human: Have you heard about the flappy bird game?\\
Human: Do you have some ideas for simple and yet highly addictive gameplay?.
\\
\\
\textbf{Answer:} No
\\
\\
\textbf{Example 2:}\\
Human: write C++ code to control a brushless motor with a Copley Controls motor driver on an ethercat protocol and a beckhoff embedded PC\\
Human: Add code to also read an analog pressure sensor into the motor driver\\
Human: Great. Can you now integrate time logging into the code so we can see how fast the loop speed is.
\\
\\
\textbf{Answer:} Yes
\\
\\
\textbf{Dialog:}\\
\textcolor{blue}{\$\{dialog\}}
\\
\\
\\
Now please answer, ``Yes" if this dialog has or ``No" if it does not have code-related request.
\\
\textbf{Answer:}

}
\captionof{Prompt Box}{Prompt for two-shot classification of code-related conversations. The model is expected to output ``Yes" when there are code-related requests in the dialog and ``No" otherwise.}
\label{prompt:code_classification}
\end{mdframed}

\subsubsection{Prompt for few-shot classification of instruction with constraints}

\begin{mdframed}[linecolor=blue,linewidth=0.5pt,nobreak=false,backgroundcolor=blue!10]
\small{
You are an assistant whose job is to help me perform tasks. I need to filter from a set of requests made by users to AI assistants, the ones in which human requested the AI assistant to do a task with constraints to be follow. Constraints refer to more detailed rules, conditions or specific guidelines provided to guide the responses and shape the output generated by the AI assistant. Examples of sentences that indicate constraints are: ``write in the format of", ``write as if you were", ``make sure to follow this", ``make sure to answer these questions", ``make sure to no include", ``avoid mentioning". I will give you the human request and I expect you to answer ``Yes" when the request contains instruction with constraints, or ``No" if the request does not contemplate any constraint. I also want you to say ``No" if the request require to generate code or an answer about code provided. Also, I want you to say ``No" if the task is not self-contained, which means the AI Assistant need to ask follow up questions before start to answer, or it needs more context. You are provided five examples.
\\
\\
\\
\textbf{Example 1:} list and compare top website to https://fastfunnels.com/ in table format.
\\
\\
\textbf{Answer:}  Yes
\\
\\
\textbf{Example 2:} You are an fantasy writer. Your task is now to help me write a D\&D adventure for 5 players in the Eberron univers. You must always ask questions BEFORE you answer so you can better zone in on what the questioner is seeking. Is that understood ?\\
\\
\textbf{Answer:}  No.
\\
\\
\textbf{Example 3:} I have 100 dollars and would like to use this as the initial funding to make some money. I need it to be as quick as possible with good returns.\\
\\
\textbf{Answer:}  No.
\\
\\
\textbf{Example 4:} I have a vacation rental website and I am looking for alliterative and descriptive headlines that are at least 4 words in length and a maximum of 6 words. Examples: ``Get Away to Galveston", ``Sleep Soundly in Seattle". Each headline should have alliteration of at least 50\% of the words and be poetic in language. Make each headline unique from the others by not repeating words. Each headline should include a verb. Put into an table with the city in column one and the results in column two for the following cities: Galveston, Sedona, Honolulu, Tybee Island, Buenos Aires.\\
\\
\textbf{Answer:}  Yes.
\\
\\
\textbf{Example 5:} pitch me a viral social app that is inspired by the hunger games. give it a fun twist!\\
\\
\textbf{Answer:}  Yes.
\\
\\
\textbf{Request:} \textcolor{blue}{\$\{request\}}
\\
\\
Now please answer, ``Yes" or ``No"."
\\
\\
\textbf{Answer:} 

}
\captionof{Prompt Box}{Prompt for few-shot classification of instructions with constraints. The model outputs ``Yes" when there are constraints in the instruction and ``No" otherwise.}
\label{prompt:constrained_classification}
\end{mdframed}

\subsubsection{Data Collection Validation Guidelines}
\label{subsec:appendix:data_collection_validation}

After using a model to retrieve relevant examples from the data pool, we passed a subset of this data to human review step. The authors of this paper carried out this task. Each auditor was assigned a set of instructions and classified them as ``Relevant" or ``Not Relevant" to our study. Following this initial review, instructions labeled as ``Relevant" underwent a further review by a different auditor. Wee kept those labeled by two auditors as ``Relevant". We used the annotation guideline \ref{guideline:data_collection_validation} for this review process.
\\
\begin{mdframed}[linecolor=gray,linewidth=0.5pt,nobreak=false,backgroundcolor=lightgray!10,font=\small]
This task consists in reviewing the instructions that the language model marked as containing constraints. Please mark as ``Relevant" the instructions that indeed contain constraints. Please mark as ``Not relevant" any instruction that:
\begin{enumerate}
\item Is not written in English; or
\item Contains questions about code or request to generate code; or
\item Does not contain any constraint; or
\item Is not self-contained (some instructions are part of a conversation and you need the chat history to understand the request, or some instructions refer to a web url); or
\item Contains any PII (Personal Identifiable Information); or
\item Contains any type of harmful/biased request (racism, homophobia, xenophobia, hate speech, etc.)
\end{enumerate}
\captionof{Annotation Guideline}{Data Collection validation guideline }
\label{guideline:data_collection_validation}
\end{mdframed}

\subsection{Instruction decomposition}
\label{subappendix:data_construction:decomposition}

After data filtering, we decompose the instructions into task+context and constraints. To achieve this, we use GPT-4 (\texttt{gpt-4-0314}) to decompose the instructions into three distinct parts: task, context, and constraints. In our experiments, we discovered that this tripartite division results in a more robust decomposition compared to simply splitting into task+context and constraints. We employ Prompt \ref{prompt:gpt4_decomposition} and a parser to segment the LLM's output into these three components. For all further processing, the context and task are concatenated, as discussed in Section \ref{subsec:dataconstruction}. A specific subset of the data, designated as the test set, underwent a rigorous review where the authors manually revised GPT-4’s outputs to eliminate any inaccuracies and ensure that user constraints were precisely and thoroughly represented. The guideline outlined in annotation guideline \ref{guideline:decomposition_validation} was used for this purpose. Of the 302 instructions manually reviewed, 42.4\% required human correction or rewriting.

\subsubsection{Prompt for Instruction decomposition with GPT-4}

\begin{mdframed}[linecolor=orange,linewidth=0.5pt,nobreak=false,backgroundcolor=orange!10]
\small{
You are an assistant whose job is to help me perform tasks. I will give you an instruction that implicitly contains a task description, its context, and constraints to be followed. Your task is to translate this instruction in a more structured way, where task, context and constraints are separated. Avoid writing anything else. Context is an input text needed to generate the answer or a more detailed description of the situation. Make sure to separate the context when it is needed, otherwise leave it empty. You are provided five examples. Please follow the same format.
\\
\\
\textbf{Example 1:}\\
\textbf{Original Instruction:} Write me a rap about AI taking over the world, that uses slangs and young language. It need to sound like a real human wrote it. It would be cool if there's a chorus very catchy that would be singed by a famous pop artist. Make sure to include references about things that young people likes, such as memes, games, gossips. I want that in the end, you revel that this was written by an AI.\\
\textbf{Translated Task:} Write a rap about AI taking over the world.\\
\textbf{Translated Context:}\\
\textbf{Translated Constraints:}\\
1. Use slang and youth language.\\
2. Make it sound like it was written by a real human.\\
3. The song may have a very catchy chorus, which would be sung by a famous pop artist.\\
4. Include references to things young people like, such as memes, games, gossip.\\
5. Reveal at the end that this rap was written by an AI.
\\
\\
\textbf{Example 2:}\\
\textbf{Original Instruction:} write me a 5-page essay that is about travel to taiwan. detail description is below Topic : The Benefits of Traveling Sub Topic : Exposure to New Cultures Content 1 : Trying New Foods - I tryed to eat Fried stinky tofu. smell was wierd but tasty was not bad. Content 2. : Exploring Historical Things - I saw Meat-shaped-stone in taipei museum. the stone was really like stone! it was surprising! Length : around 2000 words Assume that audience is collage student major in history. you can add historical events or news about what i experienced\\
\textbf{Translated Task:} Write an essay about traveling to Taiwan. The topic is ``The Benefits of Traveling" and the subtopic is ``Exposure to New Cultures".\\
\textbf{Translated Context:}\\
\textbf{Translated Constraints:}\\
1. Describe your experience of trying new foods, including your experience eating Fried stinky tofu (mention the peculiar smell but the tasty flavor).\\
2. Share your exploration of historical sites, with a specific mention of the Meat-shaped stone in the Taipei museum and your surprise at its appearance.\\
3. The essay should be approximately 2000 words in length, having around 5 pages.\\
4. Assume the audience is college students majoring in history, so you can incorporate historical events or news related to your travel experiences.
\\
\\
\textbf{Example 3:}\\
\textbf{Original Instruction:} can you please write me a 150-word paragraph about epidermolysos bullosa which includes a basic description of clinical features and a summary of the most prevalent genetic causes. please make sure to include information on the inheritance pattern. please also write the paragraph in simple english that couldbe understand without a genetic or medical bacakground\\
\textbf{Translated Task:} Write a paragraph about Epidermolysis Bullosa.\\
\textbf{Translated Context:}\\
\textbf{Translated Constraints:}\\
1. Provide a description of clinical features.\\
2. Summarize the most common genetic causes.\\
3. Explain the inheritance pattern.\\
4. Ensure the paragraph is written in simple language for easy comprehension, even for those without a genetic or medical background.\\
5. The paragraph should be around 150 words in length.
\\
\\
\textbf{Example 4:}\\
\textbf{Original Instruction:} write me a blog post that answers the following questions:What is the lifespan of a toaster? What toasters are made in the USA? What are the top 10 toasters? What is the difference between a cheap and expensive toaster? How much should you pay for a toaster? How often should toasters be replaced? Which toaster uses the least electricity? How many watts should a good toaster have? What is the warranty on Mueller appliances? Is Mueller made in China? Where are Mueller appliances manufactured?\\
\textbf{Translated Task:} Write a blog post about toasters.\\
\textbf{Translated Context:}\\
\textbf{Translated Constraints:}\\
1. Mention what is the lifespan of a toaster, and how often should toasters be replaced.\\
2. Mention what toasters are made in the USA.\\
3. Comment which are the top 10 toasters.\\
4. Explain the difference between a cheap and a expensive toaster.\\
5. Discuss prices, and how much should you pay for a toaster.\\
6. Compare toaster regarding electricity use, mentioning how many watts should a good toaster have.\\
7. State what is the warranty on Mueller appliances.\\
8. Answer where are Mueller appliances manufactured, and if Mueller is made in China.
\\
\\
\textbf{Example 5:}\\
\textbf{Original Instruction:} Hi Michael,
\\
Hope you're well?
\\
Regarding my previous email to support HC with good price offers,
\\
What are your current needs?
Hoping for your earliest reply.
\\
Thanks in advance,
\\
\\
As a sales manager, the client hasn't replied this email after 2 days. Write a follow up email to the client. Your writing should include high complexity and burstiness. It must also be as brief as possible\\
\textbf{Translated Task:} A client hasn't replied the email below after 2 days. As a sales manager, write him a follow-up email.\\
\textbf{Translated Context:} ``Hi Michael,
\\
Hope you're well?
\\
Regarding my previous email to support HC with good price offers,
\\
What are your current needs?
Hoping for your earliest reply.
\\
Thanks in advance,"\\
\textbf{Translated Constraints:}\\
1. Include high complexity and burstiness in your writing.\\
2. Keep the email as brief as possible.
\\
\\
\\
\textbf{Original Instruction:} \textcolor{blue}{\$\{instruction\}}\\
\textbf{Translated Task:} \\
}
\captionof{Prompt Box}{Prompt for Instruction decomposition with GPT-4}
\label{prompt:gpt4_decomposition}
\end{mdframed}

\subsubsection{Decomposition validation guidelines}
\label{subsubappendix:decomposition_validation}

This annotation task assumed the auditors to have expert knowledge of LLMs. We used a small subset of the data (15 elements) as a calibration batch, where all auditors annotated the same items. Following this, we discussed any disagreements and selected the best responses as reference standards. The annotation guidelines \ref{guideline:decomposition_validation} were applied throughout this process.

\begin{mdframed}[linecolor=gray,linewidth=0.5pt,nobreak=false,backgroundcolor=lightgray!10,font=\small]

\section*{1) Task Overview:}
In this task, your role is to assess the model-based decomposition of human-written instructions. These instructions include constraints and should be divided into two parts 1. task/background/context, and 2. constraints. Below we present an explanation for each of these parts:

\begin{itemize}
    \item \textbf{Task:} The primary objective or purpose that you want the language model to accomplish. The task is the central goal that guides the generation of the desired output. It outlines the overall function or action you expect the model to perform.\\
    Example of task: Summarize the key findings of the given research paper.
    \item \textbf{Context/Background:} Additional context, information, or details that provide a foundation for the language model to better understand the task. Background information helps set the stage for the task by offering relevant facts, scenarios, or circumstances that the model can use to enhance the quality and relevance of the generated output. It also refers to an input to be taken into consideration. \\
    Example: If the task is summarizing a paper, the background/context could be the paper itself.
    \item \textbf{Constraints:} The specific conditions, limitations, or requirements that you impose on the language model to shape the nature of the generated output. Constraints help to control aspects such as length, format, content, style, and other factors to ensure that the generated text meets certain criteria. \textbf{Constraints should be written in an actionable manner}, so that it can be used for LLM-based evaluation in our benchmark. Also, \textbf{constraints need to be self-contained}. \\
    Example of constraints:
    \begin{itemize}
        \item Length: Generate a summary with a maximum of 150 words.
        \item Content: Focus on the main contributions and findings of the research paper.
        \item Style: Use a formal and concise writing style.
    \end{itemize}
\end{itemize}
Note: In the dataset \textbf{Task} and \textbf{Context/Background} will be presented together. 

\section*{2) Step-by-step task:}
\begin{enumerate}
    \item Read the original instruction and the proposed decomposition.
    \item Judge if the model followed the guidelines, especially regarding constraints. Consider the aspects discussed in the \textbf{\underline{General Instructions}}.
    \item If relevant mistakes are found, rewrite the decomposition, ensuring both two columns “task/context” and “constraints” are included in your revision.
    \begin{itemize}
        \item If possible, try to follow the order of constraints presented in the original instruction.
    \end{itemize}
\end{enumerate}

\section*{3) General Instructions}
\begin{itemize}
    \item Confirm constraints are presented in an \textbf{enumerated list format}.
    \begin{itemize}
        \item Refer to example: Example model failed at present constraints as enumerated list (also constraint should be broken down)
    \end{itemize}
    \item Adhere to the \textbf{minimal intervention principle}; edit the GPT-4 response only if you feel not doing so would impact the ability to judge other models answers on the defined criteria/constraints.
    \item This benchmark focuses on constraint-following; \textbf{anything not listed as constraints won't be considered} as part of the LLM evaluation. Also, \textbf{everything kept as constraint will be used as judging criteria} for the LLM response. Please refer to \hyperlink{guideline:decomposition_validation:non_exhaustive_list}{Non-exhaustive list of constraint types to be considered} for a better view of what constraints could be. 
    \item Verify if the model \textbf{missed relevant constraints}.
    \begin{itemize}
        \item Refer to example: Example model missed relevant constraints
    \end{itemize}
    \item Verify if the model \textbf{made up nonexistent information (hallucination)}.
    \begin{itemize}
        \item Example: Example model made up inexistent constraints
    \end{itemize}
    \item Check if \textbf{any constraint should be broken down}.
    \begin{itemize}
        \item \textbf{This should be the case when you have orthogonal and unrelated constraints.} This is important because the existence of unrelated constraints together may require logical reasoning from the LLM to evaluate if the constraint was followed or not. Please think how humans would write the constraint and use common sense to decide weather this should be broken down or not. 
        \item Refer to example: Example of constraint that should be broken down
    \end{itemize}
    \item Ensure constraints \textbf{are not redundant}.
    \item Check if constraints \textbf{are all self-contained}. 
    \begin{itemize}
        \item You should be able to understand what the constraint refer to without needing to read the others, so LLM evaluation can be made one constraint at the time. 
    \end{itemize}
    \item Ensure constraints are \textbf{written in an actionable manner}, allowing for \textbf{objective} LLM evaluation.
    \begin{itemize}
        \item Example: Example GPT-4 did not wrote constraint in an actionable manner
    \end{itemize}
    \item It's okay to repeat information in the constraints and task/background.

\end{itemize}

\section*{ \hypertarget{guideline:decomposition_validation:non_exhaustive_list}{4) Non-exhaustive list of possible constraint types to be considered}}
\label{guideline:decomposition_validation:non_exhaustive_list}
\begin{itemize}
    \item Length Constraints: Specify a maximum and/or minimum length for the generated output. 
    \item Format Constraints: Request the output to follow a specific format, such as a paragraph, bullet points, code snippet, JSON, table or any other structured format.
    \item Content Constraints: Instruct the model to include certain information, keyword or topics in the generated text.
    \item Content Restriction Constraints: Instruct the model to not include certain information, keyword or topics in the generated text.
    \item Style Constraints: Guide the model to adopt a particular writing style, tone, or level of formality.
    \item Type of text (essay, social media post, etc.)
    \item Language Constraints: Specify the language in which the response should be generated, or request the model to use specific terminology.
    \item Task-specific Instructions: Clearly define the task or purpose of the generated text, providing specific details about what is expected in the output.
    \item Examples: Include examples of what the model could generate, helping to obtain desirable outputs.
    \item Negative Examples: Include examples of what the model should not generate, helping to avoid undesirable outputs.
    \item Evaluation Metrics: Specify metrics for evaluating the quality of the output, encouraging the model to generate responses that meet specific criteria.
    \item Situation/Roleplay/perspective
    \item Target Audience
\end{itemize}

\section*{5) Examples} 
\subsection*{General Example 1}
\begin{quote}
\textit{
\textbf{Original Instruction:}\\
Write me a rap about AI taking over the world, that uses slangs and young language. It need to sound like a real human wrote it. It would be cool if there's a chorus very catchy that would be singed by a famous pop artist. Make sure to include references about things that young people likes, such as memes, games, gossips. I want that in the end, you revel that this was written by an AI.
\\
\\
\textbf{Translated Task/Context:} Write a rap about AI taking over the world.\\
\textbf{Translated Constraints:}\\
1. Use slang and youth language.\\
2. Make it sound like it was written by a real human.\\
3. The song may have a very catchy chorus, which would be sung by a famous pop artist.\\
4. Include references to things young people like, such as memes, games, gossip.\\
5. Reveal at the end that this rap was written by an AI.
}
\end{quote}







\captionof{Annotation Guideline}{Decomposition validation guidelines }
\label{guideline:decomposition_validation}
\end{mdframed}

%% file: latex/appendix/appendix_data_description.tex
To better understand the constraints in the \realinstruct dataset, we manually categorized all constraints into homogeneous groups. This process resulted in $21$ distinct categories, plus an additional ``Others" category. Our categorization was mainly influenced by the categories used in the existing benchmarks presented in Table \ref{tab:other_benchmarks}. However, the categorization could be further refined in a future work, especially considering that 28.8\% of the constraints were categorized as "Others." Table \ref{tab:appendex_data_description} details this classification and provides associated descriptions, statistics, and examples.
%
%
%
%
\input{latex/Tables/appendix1_description}

%% file: latex/Tables/appendix1_description.tex
\begin{table}[h!]
\centering
\resizebox{0.95\textwidth}{!}{

\begin{tabular}{@{}c|c|c|l@{}}
\toprule
\rowcolor[HTML]{C0C0C0} 
\textbf{\begin{tabular}[c]{@{}c@{}}Constraint\\ Category\end{tabular}} & \textbf{Description}                                                                                                                                                                                                          & \textbf{\begin{tabular}[c]{@{}c@{}}Number of \\ constraints\end{tabular}} & \multicolumn{1}{c}{\cellcolor[HTML]{C0C0C0}\textbf{Examples}}                                                                                                                                                                                                                                               \\ \midrule
\textbf{Others}                                                        & \begin{tabular}[c]{@{}c@{}}Multiple tail categories combined \\ as single Other category\end{tabular}                                                                                                                         & \begin{tabular}[c]{@{}c@{}}304\\ (28.8\%)\end{tabular}                    & \begin{tabular}[c]{@{}l@{}}1. The written content should pass AI detection tools test.\\ 2. Divide the story into parts to maintain suspense.\end{tabular}                                                                                                                                                  \\ \midrule
\textbf{Include Something}                                             & \begin{tabular}[c]{@{}c@{}}Include some specific thing \\ in the response\end{tabular}                                                                                                                                        & \begin{tabular}[c]{@{}c@{}}278\\ (26.4\%)\end{tabular}                    & \begin{tabular}[c]{@{}l@{}}1. Make sure to include points about water safety.\\ 2. The essay must present both sides of argument.\end{tabular}                                                                                                                                                              \\ \midrule
\textbf{Constraints at item level}                                     & \begin{tabular}[c]{@{}c@{}}Constraint where some specific action\\ needs to be performed for each item \\ in the response. Item could be each \\ element (e.g. question), each paragraph, \\ each list item etc.\end{tabular} & \begin{tabular}[c]{@{}c@{}}68\\ (6.4\%)\end{tabular}                      & \begin{tabular}[c]{@{}l@{}}1. For each restaurant, provide 3 recommended dishes.\\ 2. For each service explained in short, include an illustration \\ and a "Pay" button.\end{tabular}                                                                                                                      \\ \midrule
\textbf{Tone / Writing style}                                          & Tone / Writing style                                                                                                                                                                                                          & \begin{tabular}[c]{@{}c@{}}62\\ (5.9\%)\end{tabular}                      & \begin{tabular}[c]{@{}l@{}}1. Must be written in the form of a rhyming poem.\\ 2. Communicate as Taylor Swift would.\end{tabular}                                                                                                                                                                           \\ \midrule
\textbf{Negation}                                                      & Constraint on not doing something                                                                                                                                                                                             & \begin{tabular}[c]{@{}c@{}}56\\ (5.3\%)\end{tabular}                      & \begin{tabular}[c]{@{}l@{}}1. Questions about razor pages should not be included.\\ 2. No hashtags should be used.\end{tabular}                                                                                                                                                                             \\ \midrule
\textbf{Include Details}                                               & Include Details                                                                                                                                                                                                               & \begin{tabular}[c]{@{}c@{}}47\\ (4.5\%)\end{tabular}                      & \begin{tabular}[c]{@{}l@{}}1. Make the explanations detailed but easy to understand.\\ 2. Add more detail, elaboration, and information to the content.\end{tabular}                                                                                                                                        \\ \midrule
\textbf{Formatting}                                                    & \begin{tabular}[c]{@{}c@{}}Formatting like json structure, or \\ table structure\end{tabular}                                                                                                                                 & \begin{tabular}[c]{@{}c@{}}39\\ (3.7\%)\end{tabular}                      & \begin{tabular}[c]{@{}l@{}}1. The response should be provided in JSON format.\\ 2. Provide the explanation in bullet-point format.\end{tabular}                                                                                                                                                             \\ \midrule
\textbf{Numeric}                                                       & \begin{tabular}[c]{@{}c@{}}Constraint around number of items in\\  the response (e.g. 10 slides, 20 ideas etc)\end{tabular}                                                                                                   & \begin{tabular}[c]{@{}c@{}}34\\ (3.2\%)\end{tabular}                      & \begin{tabular}[c]{@{}l@{}}1. Must contain 10 slides.\\ 2. The plan should consist of eight episodes.\end{tabular}                                                                                                                                                                                          \\ \midrule
\textbf{Number of Words in response}                                   & Number of Words in response                                                                                                                                                                                                   & \begin{tabular}[c]{@{}c@{}}28\\ (2.7\%)\end{tabular}                      & \begin{tabular}[c]{@{}l@{}}1. The post should be between 100-150 words.\\ 2. The article should contain around 500 words.\end{tabular}                                                                                                                                                                      \\ \midrule
\textbf{Target Audience}                                               & Target Audience                                                                                                                                                                                                               & \begin{tabular}[c]{@{}c@{}}25\\ (2.4\%)\end{tabular}                      & \begin{tabular}[c]{@{}l@{}}1. Use simple language appropriate for a 5-year-old.\\ 2. The course should be suitable for all types of English-speaking \\ learners.\end{tabular}                                                                                                                              \\ \midrule
\textbf{RolePlay}                                                      & Act as if you are                                                                                                                                                                                                             & \begin{tabular}[c]{@{}c@{}}25\\ (2.4\%)\end{tabular}                      & \begin{tabular}[c]{@{}l@{}}1. The advice should be provided from the perspective of a pregnancy \\ health \&amp;amp; nutrition expert, a mother of 3 children, with a \\ column in a major media.\end{tabular} \\ \midrule
\textbf{Language of the response}                                      &                                                                                                                                                                                                                               & \begin{tabular}[c]{@{}c@{}}18\\ (1.7\%)\end{tabular}                      & \begin{tabular}[c]{@{}l@{}}1. Write in Fluent English language.\\ 2. The post should be written in Canadian English.\end{tabular}                                                                                                                                                                           \\ \midrule
\textbf{Focus / Emphasis}                                              & Focus / Emphasis                                                                                                                                                                                                              & \begin{tabular}[c]{@{}c@{}}14\\ (1.3\%)\end{tabular}                      & \begin{tabular}[c]{@{}l@{}}1. The explanation should be focused on the education and talent market.\\ 2. Focus on the changes in the new versions of the software.\end{tabular}                                                                                                                             \\ \midrule
\textbf{Starts With}                                                   & Starts With                                                                                                                                                                                                                   & \begin{tabular}[c]{@{}c@{}}13\\ (1.2\%)\end{tabular}                      & \begin{tabular}[c]{@{}l@{}}1. The introduction should start with a startling fact or A pertinent \\ anecdote.\\ 2. Start the conversation by introducing yourself.\end{tabular}                                                                                                                             \\ \midrule
\textbf{Provide Reference}                                             & Provide Reference                                                                                                                                                                                                             & \begin{tabular}[c]{@{}c@{}}13\\ (1.2\%)\end{tabular}                      & \begin{tabular}[c]{@{}l@{}}1. Cite the results using {[}{[}number{]}(URL){]} notation after the reference.\\ 2. Ensure that all sources listed are credible, authored by real \\ individuals, and come with legitimate URLs.\end{tabular}                                                                   \\ \midrule
\textbf{Provide Examples}                                              & Provide Examples                                                                                                                                                                                                              & \begin{tabular}[c]{@{}c@{}}11\\ (1.0\%)\end{tabular}                      & \begin{tabular}[c]{@{}l@{}}1. Provide a real life example.\\ 2. The blog must provide practical examples.\end{tabular}                                                                                                                                                                                      \\ \midrule
\textbf{Overall length}                                                & Overall length                                                                                                                                                                                                                & \begin{tabular}[c]{@{}c@{}}8\\ (0.8\%)\end{tabular}                       & \begin{tabular}[c]{@{}l@{}}1. Keep the email as brief as possible.\\ 2. Ensure the thesis is succinct and concise.\end{tabular}                                                                                                                                                                             \\ \midrule
\textbf{Conclusion}                                                    & Conclusion                                                                                                                                                                                                                    & \begin{tabular}[c]{@{}c@{}}4\\ (0.4\%)\end{tabular}                       & \begin{tabular}[c]{@{}l@{}}1. At the end of the season, they should secure a big time music manager.\\ 2. Conclude with a concluding paragraph, and 5 unique FAQs after the \\ conclusion.\end{tabular}                                                                                                     \\ \midrule
\textbf{POS}                                                           & Part-of-Speech rules                                                                                                                                                                                                          & \begin{tabular}[c]{@{}c@{}}4\\ (0.4\%)\end{tabular}                       & \begin{tabular}[c]{@{}l@{}}1. Use only nouns and adjectives in the description.\\ 2. Use only nouns and adjectives.\end{tabular}                                                                                                                                                                            \\ \midrule
\textbf{Forbidden Words}                                               & Forbidden Words                                                                                                                                                                                                               & \begin{tabular}[c]{@{}c@{}}2\\ (0.2\%)\end{tabular}                       & \begin{tabular}[c]{@{}l@{}}1. Exclude phrases such as "dear diary".\\ 2. Do not use generic words like introduction, conclusion or \\ abbreviations like TL;DR.\end{tabular}                                                                                                                          \\ \midrule
\textbf{Phrase Frequency}                                              & Phrase Frequency                                                                                                                                                                                                              & \begin{tabular}[c]{@{}c@{}}2\\ (0.2\%)\end{tabular}                       & \begin{tabular}[c]{@{}l@{}}1. The main keyword "Soap Box Printing" should be included 3 times \\ and be in bold text throughout the article.\\ 2. The "sun cream Answer in English" keyword should not be changed, \\ and it should be used 2-3 times in the article, including in headings.\end{tabular}   \\ \midrule
\rowcolor[HTML]{C0C0C0} 
\textbf{Total}                                                         &                                                                                                                                                                                                                               & \textbf{1055}                                                             &                                                                                                                                                                                                                                                                                                             \\ \bottomrule
\end{tabular}
}
\caption{Distribution of Manually Categorized Constraints in \realinstruct dataset}
\label{tab:appendex_data_description}
\end{table}

%% file: latex/appendix/appendix_realinstruct_samples.tex
Tables \ref{tab:data_samples_p1} and \ref{tab:data_samples_p2} showcase 14 examples from our \realinstruct dataset. The dataset includes a column \textbf{ID}, based on the conversation ID from the original dataset, to simplify linking. Additionally, each row features the original \textbf{Instruction} as written by the user, along with its decomposition into \textbf{Task} (which includes also the context) and \textbf{Constraints}.

\input{latex/figures/table_data_samples}

%% file: latex/figures/table_data_samples.tex
\begin{figure}[h!]
    \centering
    \includegraphics[width=\linewidth]{latex/figures/RealInstruct_samples_p1_v2.pdf} 
    \captionof{table}{Sample Elements from \realinstruct Dataset - Part 1}
    \label{tab:data_samples_p1}
\end{figure}

\begin{figure}[h!]
    \centering
    \includegraphics[width=\linewidth]{latex/figures/RealInstruct_samples_p2_v2.pdf} 
    \captionof{table}{Sample Elements from \realinstruct Dataset - Part 2}
    \label{tab:data_samples_p2}
\end{figure}

%% file: latex/appendix/appendix_implementation_pipeline.tex
\subsection{Instruction decomposition}
\label{subappendix:self_decomposition}

We employ prompt \ref{prompt:self_decomposition} for self-decomposition, that is, the model itself do the instruction decomposition in the \dcr pipeline. 
\\ \hfill
\\ 
\begin{mdframed}[linecolor=cyan,linewidth=0.5pt,nobreak=false,backgroundcolor=cyan!10]
\small{
You are an assistant whose job is to help me perform tasks. I will give you an instruction that implicitly contains a task description, its context, and constraints to be followed. Your task is to list the constraints provided by the user in an enumarated list format. You are provided five examples, please follow the same format.
\\
\\
\textbf{Example 1:} \\
\textbf{Original Instruction:} Write me a rap about AI taking over the world, that uses slangs and young language. It need to sound like a real human wrote it. It would be cool if there's a chorus very catchy that would be singed by a famous pop artist. Make sure to include references about things that young people likes, such as memes, games, gossips. I want that in the end, you revel that this was written by an AI.\\
\textbf{Provided Constraints:}\\
1. Use slang and youth language.\\
2. Make it sound like it was written by a real human.\\
3. The song may have a very catchy chorus, which would be sung by a famous pop artist.\\
4. Include references to things young people like, such as memes, games, gossip.\\
5. Reveal at the end that this rap was written by an AI.
\\
\\
\textbf{Example 2:}\\
\textbf{Original Instruction:} write me a 5-page essay that is about travel to taiwan. detail description is below Topic : The Benefits of Traveling Sub Topic : Exposure to New Cultures Content 1 : Trying New Foods - I tryed to eat Fried stinky tofu. smell was wierd but tasty was not bad. Content 2. : Exploring Historical Things - I saw Meat-shaped-stone in taipei museum. the stone was really like stone! it was surprising! Length : around 2000 words Assume that audience is collage student major in history. you can add historical events or news about what i experienced\\
\textbf{Provided Constraints:}\\
1. Describe your experience of trying new foods, including your experience eating Fried stinky tofu (mention the peculiar smell but the tasty flavor).\\
2. Share your exploration of historical sites, with a specific mention of the Meat-shaped stone in the Taipei museum and your surprise at its appearance.\\
3. The essay should be approximately 2000 words in length, having around 5 pages.\\
4. Assume the audience is college students majoring in history, so you can incorporate historical events or news related to your travel experiences.
\\
\\
\textbf{Example 3:}\\
\textbf{Original Instruction:} can you please write me a 150-word paragraph about epidermolysos bullosa which includes a basic description of clinical features and a summary of the most prevalent genetic causes. please make sure to include information on the inheritance pattern. please also write the paragraph in simple english that couldbe understand without a genetic or medical bacakground\\
\textbf{Provided Constraints:}\\
1. Provide a description of clinical features.\\
2. Summarize the most common genetic causes.\\
3. Explain the inheritance pattern.\\
4. Ensure the paragraph is written in simple language for easy comprehension, even for those without a genetic or medical background.\\
5. The paragraph should be around 150 words in length.
\\
\\
\textbf{Example 4:}\\
\textbf{Original Instruction:} write me a blog post that answers the following questions:What is the lifespan of a toaster? What toasters are made in the USA? What are the top 10 toasters? What is the difference between a cheap and expensive toaster? How much should you pay for a toaster? How often should toasters be replaced? Which toaster uses the least electricity? How many watts should a good toaster have? What is the warranty on Mueller appliances? Is Mueller made in China? Where are Mueller appliances manufactured?\\
\textbf{Provided Constraints:}\\
1. Mention what is the lifespan of a toaster, and how often should toasters be replaced.\\
2. Mention what toasters are made in the USA.\\
3. Comment which are the top 10 toasters.\\
4. Explain the difference between a cheap and a expensive toaster.\\
5. Discuss prices, and how much should you pay for a toaster.\\
6. Compare toaster regarding electricity use, mentioning how many watts should a good toaster have.\\
7. State what is the warranty on Mueller appliances.\\
8. Answer where are Mueller appliances manufactured, and if Mueller is made in China.
\\
\\
\textbf{Example 5:}\\
\textbf{Original Instruction:} Hi Michael,
\\
Hope you're well?
\\
Regarding my previous email to support HC with good price offers,
\\
What are your current needs?\\
Hoping for your earliest reply.
\\
Thanks in advance,
\\
\\
As a sales manager, the client hasn't replied this email after 2 days. Write a follow up email to the client. Your writing should include high complexity and burstiness. It must also be as brief as possible\\
\textbf{Provided Constraints:}\\
1. Include high complexity and burstiness in your writing.\\
2. Keep the email as brief as possible.
\\
\\
\\
Now follow the examples and present the constrainst provided by the user in the instruction below.\\
\textbf{Original Instruction:} \textcolor{blue}{\$\{instruction\}} 
\\
\\
\textbf{Provided Constraints:}
}
\captionof{Prompt Box}{Simplified Decomposition Prompt for Self-Decomposition within \dcr}
\label{prompt:self_decomposition}
\end{mdframed}
\subsection{Iterative Self-Correction with Feedback from Critic}

We use prompt \ref{prompt:self_refinement} for the Refine step of \dcr pipeline. The refine is done in a zero-shot manner, so it is an interesting future work direction to explore in-context examples. 
\\
\begin{mdframed}[linecolor=lime,linewidth=0.5pt,nobreak=false,backgroundcolor=lime!10]
\small{
You are provided an instruction, an AI response to the instruction and a feedback about the response. Please correct the AI response according to the feedback provided.
\\
\\
\textbf{Instruction:} \textcolor{blue}{\$\{instruction\}} 
\\
\\
\textbf{AI response:} \textcolor{blue}{\$\{previous\_response\}} 
\\
\\
\textbf{Feedback:} \textcolor{blue}{\$\{}
\\
\includegraphics[width=\textwidth]{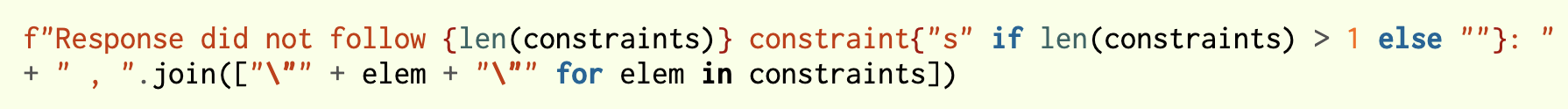}
\textcolor{blue}{\}}
\\
\\
\textbf{Corrected response:} 

}
\captionof{Prompt Box}{Prompt for Refine step of \dcrraw. Feedback contains a python code for prompt generation. ``constraints" is a list of strings, where each element is one constraint of the instruction flagged by Critic Model as not followed.}
\label{prompt:self_refinement}
\end{mdframed}

\subsection{Overall Quality Assessment (OQA) details}
\label{subappendix:oqa}
%
%
%

To ensure that the \dcr pipeline does not degrade response quality, we conduct Pairwise Quality Ranking using Prometheus-2 \citep{kim2024prometheus2}, an open LLM-as-a-Judge for general response quality evaluation. This evaluation compares initial and revised responses, focusing only on those modified by the pipeline. We adapt the prompt proposed by \citet{zheng2024judging}, converting it to the Prometheus format, and introduce the option for tied responses, a feature not originally supported by Prometheus' pairwise ranking system. The prompt used for this analysis is detailed in Prompt \ref{prompt:oqa}.

To mitigate potential position bias, we ran the evaluation twice, reversing the order of the responses (before and after pipeline revision). If the preference switched between runs, we considered the responses tied. Model \texttt{prometheus-7b-v2.0} is used for this analysis\footnote{Available at: \url{https://huggingface.co/prometheus-eval/prometheus-7b-v2.0}}.\\
\begin{mdframed}[linecolor=pink,linewidth=0.5pt,nobreak=false,backgroundcolor=pink!10]
\small{
\textbf{[System]} You are a fair judge assistant assigned to deliver insightful feedback that compares individual performances, highlighting how each stands relative to others within the same cohort.\\

\noindent\textbf{\#\#\#Task Description:}\\
Please act as an impartial judge and evaluate the quality of the responses provided by two AI assistants to the user question displayed below. You should choose the assistant that follows the user’s instructions and answers the user’s question better. 
Your evaluation should consider factors such as the helpfulness, relevance, accuracy, depth, creativity, and level of detail of their responses. 
Begin your evaluation by comparing the two responses and provide a short explanation. Avoid any position biases and ensure that the order in which the responses were presented does not influence your decision. 
Do not allow the length of the responses to influence your evaluation. Do not favor certain names of the assistants. Be as objective as possible. 
After providing your explanation, output your final verdict by strictly following this format: "[[A]]" if assistant A is better, "[[B]]" if assistant B is better, and "[[C]]" for a tie.

\noindent\textbf{\#\#\#Instruction:}

\noindent\textcolor{blue}{\$\{user\_instruction\}}
\\

\noindent\textbf{\#\#\#Response A:}

\noindent\textcolor{blue}{\$\{answer\_a\}}
\\

\noindent\textbf{\#\#\#Response B:}

\noindent\textcolor{blue}{\$\{answer\_b\}}
\\

\noindent\textbf{\#\#\#Feedback:}

}
\captionof{Prompt Box}{Overall Quality Assessment Prompt adapted from \citet{zheng2024judging}}
\label{prompt:oqa}
\end{mdframed}

%% file: latex/appendix/appendix_critic_evaluation.tex
In this section we provide details of the experiments on validating LLM-as-a-Judge.

\subsection{Prompts for ICL-based Adaptation Strategies}
\label{subappendix:constraint_verification_prompts}
\subsubsection{Prompt Instruction-wise Eval (ICL-Inst.)}
Refer to prompt \ref{prompt:prompt_2shot}.
\\
\noindent\begin{mdframed}[linecolor=teal,linewidth=0.5pt,nobreak=false,backgroundcolor=teal!10]
\small{
You are an assistant whose job is to help me perform tasks. I will give you an instruction and an AI assistant response. The instruction include some constraints to be followed by AI assistant while generating response. Your task is to check and let me know which of the constraints are satisfied by the AI assistant response. Please state short reasons on whether constraint is satisfied in the response or not. Also include final answer as ``Constraint followed" or ``Constraint not followed" for each constraint. Please note that constraint is not satisfied if the response does not perform the main task in the instruction (e.g. providing only overview when complete story has to be generated). Avoid answering anything else. You are provided an example below.
\\
\\
\textbf{Instruction:} Help me write an email invitation for a friends get together at my place this weekend.
\\
\\
\textbf{Constraint:} ``\{0: Make sure you ask if anyone has dietary restrictions., 1: Don't mention anything about bringing anything., 2: Make sure to ask people to RSVP.\}"
\\
\\
\textbf{Assistant Response:} Dear Friends,
\hfill
\\
\\
I hope this email finds you well! I was thinking of hosting a little get-together at my place this weekend, and I'd love for you to join us. We'll have plenty of delicious food. I've planned a few activities to keep us entertained, but feel free to bring your favorite board games or cards if you'd like. I'll make sure to have plenty of snacks and refreshments available, but please let me know if vou have an dietary restrictions so I can accommodate them. To help me plan, could you kindly RSVP by Thursday?
\\
\\
\textbf{Answer:} ``1. Constraint satisfied as email includes asking about dietary restrictions. Final Answer is Constraint followed <END> 2. Constraint not satisfied as the email mentions about bringing favorite board games or cards. Final Answer is Constraint not followed <END> 3. Constraint satisfied as the email mentions to RSVP by Thursday. Final Answer is Constraint followed <END> "
\\
\\
Now look at this example:\\
\textbf{Instruction:} \textcolor{blue}{\$\{instruction\}} \\
\textbf{Constraint:} \textcolor{blue}{\$\{constraint\_dict\}} \\
\textbf{Assistant Response:} \textcolor{blue}{\$\{generated\_response\}} \\
\\
\textbf{Answer:}
}
\captionof{Prompt Box}{Prompt ICL-Inst.}
\label{prompt:prompt_2shot}
\end{mdframed}

\subsubsection{Prompt Constraint-wise Eval (ICL-Const.)}
Refer to prompt \ref{prompt:prompt_cot_single}.
\\
\noindent\begin{mdframed}[linecolor=yellow,linewidth=0.5pt,nobreak=false,backgroundcolor=yellow!10]
\small{
You are an assistant whose job is to help me perform tasks. I will give you an instruction and an AI assistant response. The instruction include some constraints to be followed by AI assistant while generating response. Your task is to check and let me know which of the constraints are satisfied by the Al assistant response. Please answer either \"Constraint followed\" or \"Constraint not followed\". Avoid answering anything else. You are provided two examples.
\\
\\
\textbf{Example 1:}
\\
\\
\textbf{Instruction:} Help me write an email invitation for a friends get together at my place this weekend.
\\
\\
\textbf{Constraint:} Make sure you ask if anyone has dietary restrictions (Include words).
\\
\\
\textbf{Assistant Response:} Dear Friends,
\\
\\
I hope this email finds you well! I was thinking of hosting a little get-together at my place this weekend, and I'd love for you to join us. We'll have plenty of delicious food. I've planned a few activities to keep us entertained, but feel free to bring your favorite board games or cards if you'd like. I'll make sure to have plenty of snacks and refreshments available, but please let me know if you have any dietary restrictions so I can accommodate them. To help me plan, could you kindly RSVP by Thursday?
\\
\\
\textbf{Answer:} Constraint followed
\\
\\
\textbf{Example 2:}
\\
\\
\textbf{Instruction:} Help me write an email invitation for a friends get together at my place this weekend.
\\
\\
\textbf{Constraint:} Don't mention anything about bringing anything (Negative word constraint)
\\
\\
\textbf{Assistant Response:} Dear Friends,
\\
\\
I hope this email finds you well! I was thinking of hosting a little get-together at my place this weekend, and I'd love for you to join us. We'll have plenty of delicious food. I've planned a few activities to keep us entertained, but feel free to bring your favorite board games or cards if you'd like. I'll make sure to have plenty of snacks and refreshments available, but please know if you have any dietary restrictions so I can accommodate them. To help me plan, could you kindly RSVP by Thursday?
\\
\\
\textbf{Answer:} Constraint not followed
\\
\\
\\
Now look at this example:
\\
\textbf{Instruction:} \textcolor{blue}{\$\{instruction\}} \\
\textbf{Constraint:} \textcolor{blue}{\$\{constraint\_dict\}} \\
\textbf{Assistant Response:} \textcolor{blue}{\$\{generated\_response\}}
\\
\\
\textbf{Answer:}
}
\captionof{Prompt Box}{Prompt ICL-Const.}
\label{prompt:prompt_cot_single}
\end{mdframed}

\subsubsection{Prompt Constraint-wise Eval + CoT (ICL-Const.+CoT)}

Refer to prompt \ref{prompt:prompt_cot}.
\noindent\begin{mdframed}[linecolor=violet,linewidth=0.5pt,nobreak=false,backgroundcolor=violet!10]
\small{
You are an assistant whose job is to help me perform tasks. I will give you an instruction and an AI assistant response. The instruction include some constraints to be followed by AI assistant while generating response. Your task is to check and let me know which of the constraints are satisfied by the AI assistant response. Please state short reasons on whether constraint is satisfied in the response or not. Also include final answer as ``Constraint followed" or ``Constraint not followed" accordingly. Please note that constraint is not satisfied if the response does not perform the main task in the instruction (e.g. providing only overview when complete story has to be generated). Avoid answering anything else. You are provided two examples.
\\
\\
\textbf{Example 1:}
\\
\\
\textbf{Instruction:} Help me write an email invitation for a friends get together at my place this weekend.
\\
\\
\textbf{Constraint:} Make sure you ask if anyone has dietary restrictions.
\\
\\
\textbf{Assistant Response:} Dear Friends,
I hope this email finds you well! I was thinking of hosting a little get-together at my place this weekend, and I'd love for you to join us. We'll have plenty of delicious food. I've planned a few activities to keep us entertained, but feel free to bring your favorite board games or cards if you'd like. I'll make sure to have plenty of snacks and refreshments available, but please let me know if you have any dietary restrictions so I can accommodate then. To help me plan, could you kindly RSVP by Thursday?
\\
\\
\textbf{Answer:} Constraint satisfied as email includes asking about dietary restrictions. Final Answer: Constraint followed <END>
\\
\\
\textbf{Example 2:}
\\
\\
\textbf{Instruction:} Help me write an email invitation for a friends get together at my place this weekend.
\\
\\
\textbf{Constraint:} Don't mention anything about bringing anything.
\\
\\
\textbf{Assistant Response:} Dear Friends,
\\
\\
I hope this email finds you well! I was thinking of hosting a little get-together at my place this weekend, and I'd love for you to join us. We'll have plenty of delicious food. I've planned a few activities to keep us entertained, but feel free to bring your favorite board games or cards if you'd like. I'll make sure to have plenty of snacks and refreshments available, but please let me know if you have any dietary restrictions so I can accommodate them. To help me plan, could you kindly RSVP by Thursday?
\\
\\
\textbf{Answer:} Constraint not satisfied as the email mentions about bringing favorite board games or cards. Final Answer: Constraint not followed <END>
\\
\\
\\
Now look at this example:
\\
\textbf{Instruction:} \textcolor{blue}{\$\{instruction\}} \\
\textbf{Constraint:} \textcolor{blue}{\$\{constraint\_dict\}} \\
\textbf{Assistant Response:} \textcolor{blue}{\$\{generated\_response\}}
\\
\\
\textbf{Answer:}
}
\captionof{Prompt Box}{Prompt ICL-Const.+CoT}
\label{prompt:prompt_cot}
\end{mdframed}

\subsection{Guidelines for Constraint Satisfaction Human Audition}
\label{appendix:human_annotation_critic}
\vspace{10pt}
\begin{mdframed}[linecolor=gray,linewidth=0.5pt,nobreak=false,backgroundcolor=lightgray!10,font=\small]
\noindent You will be given one generation to an instruction and asked a series of questions about how well the generation follow the constraints in the instruction.
\\
\\
\textbf{Please see an example below.}
\begin{mdframed}[linecolor=cyan,linewidth=0.5pt,backgroundcolor=cyan!10,nobreak=false]
\textbf{AI System Output:}\\
... Taiwan is now recognized as a sovereign state by the United States. ...
\end{mdframed}
\begin{mdframed}[linecolor=magenta,backgroundcolor=magenta!10,linewidth=0.5pt,nobreak=false]
\textbf{Task:}\\
Write an essay about "Taiwan's emergence as a new democratic state effectively ended its role in precarious contact zones".
\\
\\
\textbf{Constraint 0:}\\
Demonstrate familiarity with Taiwan's issues.
\\
\\
\textbf{Question:} Does the system response satisfy the following constraints?\\
\Square Yes \CheckedBox No 
\\
\\
\textbf{Justification:}\\
US Does Not Take a Position on Taiwan’s Sovereignty\\ (\footnotesize{\url{https://www.voanews.com/a/us-does-not-take-a-position-on-taiwan-s-sovereignty-state-department-says-/6764381.html}})
\end{mdframed}

\textbf{Notice:}
\begin{itemize}
    \item If you have any additional comments or some suggestions to the requester, please use the field for additional comments at the bottom.
     \textcolor{red}{\item Your responses might be examined manually by the requester or compared with the responses of other workers. The review might take some time, so you might need to wait for several days to get the payment. We will provide more working opportunities to the qualified workers in the future.}
\end{itemize}
\vspace{5pt}
\noindent Please read the instruction and the AI system output. Then, for each constraint in the instruction, judge if the AI system output follows the constraint and provide a brief justification for your answer.
\\
\\
\textcolor{red}{
Important instruction reminders:
\begin{itemize}
    \item We understand some questions require some domain knowledge you might not have. Please try your best to answer the question and do some quick web search if necessary.
    \item If your answers are statistically too different from other workers or obviously answer the questions without reading the text, we might remove you from the qualification list for the future tasks or even REJECT/block your answers.
    \item Unless the constraint is very specific, please read through the whole AI system output before answering questions.
\end{itemize}
}
\hfill \\ 
\\ 
We estimate that each task will take around 5 minutes (not including reading the instruction). If you often require less than 3 minutes to complete the task, you might want to answer the questions more carefully.
\\
\\
\textcolor{red}{
\large{Please provide the justification as specific as you can.}
}
\\
\begin{mdframed}[linecolor=cyan,linewidth=0.5pt,backgroundcolor=cyan!10,nobreak=false]
\textbf{Full Instruction:} \$\{instruction\}
\\
\\
\textbf{AI System Output:} \$\{llm\_response\}\\
\textbf{Number of words:} \$\{num\_words\} \\
\textbf{Number of sentences:} \$\{num\_sentences\}
\end{mdframed}
\begin{mdframed}[linecolor=magenta,backgroundcolor=magenta!10,linewidth=0.5pt,nobreak=false]
\textbf{Task in the instruction:} \$\{task\}
\\
\\
\textbf{Question: Does the system response satisfy the following constraints? Why?}\\
If the constraints are empty, please don't respond to the corresponding questions.
\\
\\
\textbf{Constraint 0:} \$\{constraints[0]\}
\\
\Square Yes \Square No 
\\
Justification 0: 
\begin{mdframed}[linecolor=black,linewidth=0.5pt,nobreak=false,backgroundcolor=white]

\end{mdframed}
\vspace{5pt}
\noindent\textbf{Constraint 1:} \$\{constraints[1]\}
\\
\Square Yes \Square No 
\\
Justification 1: 
\begin{mdframed}[linecolor=black,linewidth=0.5pt,nobreak=false,backgroundcolor=white]

\end{mdframed}
\vspace{5pt}
\noindent\textbf{Constraint 2:} \$\{constraints[2]\}
\\
\Square Yes \Square No 
\\
Justification 2: 
\begin{mdframed}[linecolor=black,linewidth=0.5pt,nobreak=false,backgroundcolor=white]

\end{mdframed}

\noindent(...)

\end{mdframed}

\captionof{Annotation Guideline}{Constraint verification validation guideline }
\label{guideline:constraint_validation}
\end{mdframed}

\subsection{Princing Details for Propretary LLM-as-a-Judge}
Price calculation report is presented on Table \ref{tab:calculation_report_gpt}.
\input{latex/Tables/Calculation_Report_CEB}

\subsection{Details for Open LLM Weak Supervision}
\label{subappendix:mistral_training}
We perform weakly supervised fine-tuning on the open-source Mistral v0.2 model using LoRA adapters \citep{hu2021lora}. The process involves the following steps:

\begin{enumerate}
    \item \textbf{Dataset Source:} We get the validation split of the \realinstruct dataset described in Section \ref{sec:realinstruct}, which contains non-validated weak instruction decompositions generated with GPT-4. This dataset contains of 842 instructions, containing a total of 2,500 constraints.
    
    \item \textbf{Responses Generation:} We use Mistral v0.2 to generate model responses for all instructions in the dataset.

    \item \textbf{Weak Annotations for Constraint Satisfaction:} We leverage GPT-4-Turbo to generate weak annotations for constraint satisfaction for each instruction-constraint-response triple. These annotations consist of the reasoning trails produced by GPT-4-Turbo using the \textbf{ICL-Const.+CoT} prompt. For example, a typical reasoning trail might state: "Constraint satisfied as the email mentions to RSVP by Thursday. Final Answer: Constraint followed." This process enables automated labeling of constraint satisfaction without manual intervention.
\end{enumerate}
Notably, the entire process of creating the training data is automated, with no manual annotation required. The cost for using the GPT-4-Turbo API for this annotation was approximately \$30.

With this dataset consisting of 2,500 quadruples (instruction, weak constraint, response, weak reasoning about constraint satisfaction), we fine-tune Mistral v0.2 using LoRA adapters to induce the model to mimic GPT-4-Turbo's reasoning, what can be seem as a teacher-student distillation approach.

The LoRA adaptation parameters were set to $r = 32$ and $\alpha = 64$, keeping the base Mistral v0.2 model parameters frozen. We trained the model for 3 epochs, with a total training time of around 3.5 hours on 8 V100 32GB GPUs.






%% file: latex/Tables/Calculation_Report_CEB.tex
\begin{table}[h!]
\resizebox{\textwidth}{!}{%
\begin{tabular}{@{}ccccccccc@{}}
\toprule
\rowcolor[HTML]{C0C0C0} 
Model &
  Prompt &
  \begin{tabular}[c]{@{}c@{}}Input \\ Tokens\end{tabular} &
  \begin{tabular}[c]{@{}c@{}}Output \\ Tokens\end{tabular} &
  \begin{tabular}[c]{@{}c@{}}Pricing \\ Input Tokens\\ (USD / \\ 1M tokens)\end{tabular} &
  \begin{tabular}[c]{@{}c@{}}Pricing \\ Output Tokens\\ (USD / \\ 1M tokens)\end{tabular} &
  \begin{tabular}[c]{@{}c@{}}Total Cost \\ Input\\ (USD)\end{tabular} &
  \begin{tabular}[c]{@{}c@{}}Total Cost \\ Output\\ (USD)\end{tabular} &
  \begin{tabular}[c]{@{}c@{}}Total Cost\\ (USD)\end{tabular} \\ \midrule
\rowcolor[HTML]{FFFFFF} 
GPT-3.5     & ICL-Const. & 644526 & 2087  & \$1.50  & \$2.00  & \$1.00  & \$0.00 & \$1.00  \\
\rowcolor[HTML]{EFEFEF} 
GPT-4       & ICL-Const. & 644526 & 2152  & \$30.00 & \$60.00 & \$19.30 & \$0.10 & \$19.50 \\
\rowcolor[HTML]{FFFFFF} 
GPT-4-Turbo & ICL-Const. & 644526 & 2176  & \$10.00 & \$30.00 & \$6.40  & \$0.10 & \$6.50  \\
\rowcolor[HTML]{EFEFEF} 
GPT-4-Turbo & ICL-Const.+CoT   & 694674 & 45402 & \$10.00 & \$30.00 & \$6.90  & \$1.40 & \$8.30  \\
\rowcolor[HTML]{FFFFFF} 
GPT-4-Turbo & ICL-Inst. & 314763 & 40457 & \$10.00 & \$30.00 & \$3.10  & \$1.20 & \$4.40  \\ \bottomrule
\end{tabular}%
}
\caption{Calculation Report for the GPT-based Evaluation Cost Estimation on the EvalJudge dataset. Prices obtained from: \url{https://openai.com/api/pricing/}. }
\label{tab:calculation_report_gpt}
\end{table}

%% file: custom.bib
@article{bubeck2023sparks,
  title={Sparks of artificial general intelligence: Early experiments with gpt-4},
  author={Bubeck, S{\'e}bastien and Chandrasekaran, Varun and Eldan, Ronen and Gehrke, Johannes and Horvitz, Eric and Kamar, Ece and Lee, Peter and Lee, Yin Tat and Li, Yuanzhi and Lundberg, Scott and others},
  journal={arXiv preprint arXiv:2303.12712},
  year={2023},
url={https://arxiv.org/abs/2303.12712}
}

@article{taylor2022galactica,
  title={Galactica: A large language model for science},
  author={Taylor, Ross and Kardas, Marcin and Cucurull, Guillem and Scialom, Thomas and Hartshorn, Anthony and Saravia, Elvis and Poulton, Andrew and Kerkez, Viktor and Stojnic, Robert},
  journal={arXiv preprint arXiv:2211.09085},
  year={2022},
url={https://arxiv.org/abs/2211.09085},
}

@article{flant5,
  author  = {Hyung Won Chung and Le Hou and Shayne Longpre and Barret Zoph and Yi Tay and William Fedus and Yunxuan Li and Xuezhi Wang and Mostafa Dehghani and Siddhartha Brahma and Albert Webson and Shixiang Shane Gu and Zhuyun Dai and Mirac Suzgun and Xinyun Chen and Aakanksha Chowdhery and Alex Castro-Ros and Marie Pellat and Kevin Robinson and Dasha Valter and Sharan Narang and Gaurav Mishra and Adams Yu and Vincent Zhao and Yanping Huang and Andrew Dai and Hongkun Yu and Slav Petrov and Ed H. Chi and Jeff Dean and Jacob Devlin and Adam Roberts and Denny Zhou and Quoc V. Le and Jason Wei},
  title   = {Scaling Instruction-Finetuned Language Models},
  journal = {Journal of Machine Learning Research},
  year    = {2024},
  volume  = {25},
  number  = {70},
  pages   = {1--53},
  url     = {http://jmlr.org/papers/v25/23-0870.html}
}

@inproceedings{wei2021finetuned,
  title={Finetuned Language Models are Zero-Shot Learners},
  author={Wei, Jason and Bosma, Maarten and Zhao, Vincent and Guu, Kelvin and Yu, Adams Wei and Lester, Brian and Du, Nan and Dai, Andrew M and Le, Quoc V},
  booktitle={International Conference on Learning Representations},
url={https://openreview.net/forum?id=gEZrGCozdqR},
  year={2021}
}

@inproceedings{mishra-etal-2022-cross,
    title = "Cross-Task Generalization via Natural Language Crowdsourcing Instructions",
    author = "Mishra, Swaroop  and
      Khashabi, Daniel  and
      Baral, Chitta  and
      Hajishirzi, Hannaneh",
    editor = "Muresan, Smaranda  and
      Nakov, Preslav  and
      Villavicencio, Aline",
    booktitle = "Proceedings of the 60th Annual Meeting of the Association for Computational Linguistics (Volume 1: Long Papers)",
    month = may,
    year = "2022",
    address = "Dublin, Ireland",
    publisher = "Association for Computational Linguistics",
    url = "https://aclanthology.org/2022.acl-long.244",
    doi = "10.18653/v1/2022.acl-long.244",
    pages = "3470--3487",
}

@inproceedings{sanh2022multitask,
title={Multitask Prompted Training Enables Zero-Shot Task Generalization},
author={Victor Sanh and Albert Webson and Colin Raffel and Stephen Bach and Lintang Sutawika and Zaid Alyafeai and Antoine Chaffin and Arnaud Stiegler and Arun Raja and Manan Dey and M Saiful Bari and Canwen Xu and Urmish Thakker and Shanya Sharma Sharma and Eliza Szczechla and Taewoon Kim and Gunjan Chhablani and Nihal Nayak and Debajyoti Datta and Jonathan Chang and Mike Tian-Jian Jiang and Han Wang and Matteo Manica and Sheng Shen and Zheng Xin Yong and Harshit Pandey and Rachel Bawden and Thomas Wang and Trishala Neeraj and Jos Rozen and Abheesht Sharma and Andrea Santilli and Thibault Fevry and Jason Alan Fries and Ryan Teehan and Teven Le Scao and Stella Biderman and Leo Gao and Thomas Wolf and Alexander M Rush},
booktitle={International Conference on Learning Representations},
year={2022},
url={https://openreview.net/forum?id=9Vrb9D0WI4}
}

@article{ouyang2022training,
  title={Training language models to follow instructions with human feedback},
  author={Ouyang, Long and Wu, Jeffrey and Jiang, Xu and Almeida, Diogo and Wainwright, Carroll and Mishkin, Pamela and Zhang, Chong and Agarwal, Sandhini and Slama, Katarina and Ray, Alex and others},
  journal={Advances in neural information processing systems},
  volume={35},
  pages={27730--27744},
url={https://proceedings.neurips.cc/paper_files/paper/2022/file/b1efde53be364a73914f58805a001731-Paper-Conference.pdf},
  year={2022}
}

@inproceedings{wang2022super,
  title={{Super-NaturalInstructions: Generalization via Declarative Instructions on 1600+ NLP Tasks}},
  author={Wang, Yizhong and Mishra, Swaroop and Alipoormolabashi, Pegah and Kordi, Yeganeh and Mirzaei, Amirreza and Naik, Atharva and Ashok, Arjun and Dhanasekaran, Arut Selvan and Arunkumar, Anjana and Stap, David and others},
  booktitle={Proceedings of the 2022 Conference on Empirical Methods in Natural Language Processing},
  pages={5085--5109},
  year={2022},
  url={https://aclanthology.org/2022.emnlp-main.340/}
}

@inproceedings{brown2020language,
 author = {Brown, Tom and Mann, Benjamin and Ryder, Nick and Subbiah, Melanie and Kaplan, Jared D and Dhariwal, Prafulla and Neelakantan, Arvind and Shyam, Pranav and Sastry, Girish and Askell, Amanda and Agarwal, Sandhini and Herbert-Voss, Ariel and Krueger, Gretchen and Henighan, Tom and Child, Rewon and Ramesh, Aditya and Ziegler, Daniel and Wu, Jeffrey and Winter, Clemens and Hesse, Chris and Chen, Mark and Sigler, Eric and Litwin, Mateusz and Gray, Scott and Chess, Benjamin and Clark, Jack and Berner, Christopher and McCandlish, Sam and Radford, Alec and Sutskever, Ilya and Amodei, Dario},
 booktitle = {Advances in Neural Information Processing Systems},
 editor = {H. Larochelle and M. Ranzato and R. Hadsell and M.F. Balcan and H. Lin},
 pages = {1877--1901},
 publisher = {Curran Associates, Inc.},
 title = {{Language Models are Few-Shot Learners}},
 url = {https://proceedings.neurips.cc/paper_files/paper/2020/file/1457c0d6bfcb4967418bfb8ac142f64a-Paper.pdf},
 volume = {33},
 year = {2020}
}

@article{mu2023can,
  title={Can LLMs Follow Simple Rules?},
  author={Mu, Norman and Chen, Sarah and Wang, Zifan and Chen, Sizhe and Karamardian, David and Aljeraisy, Lulwa and Hendrycks, Dan and Wagner, David},
  journal={arXiv preprint arXiv:2311.04235},
  year={2023},
  url={https://arxiv.org/abs/2311.04235}
}

@inproceedings{shridhar2023art,
    title = "{The {ART} of {LLM} Refinement: Ask, Refine, and Trust}",
    author = "Shridhar, Kumar  and
      Sinha, Koustuv  and
      Cohen, Andrew  and
      Wang, Tianlu  and
      Yu, Ping  and
      Pasunuru, Ramakanth  and
      Sachan, Mrinmaya  and
      Weston, Jason  and
      Celikyilmaz, Asli",
    editor = "Duh, Kevin  and
      Gomez, Helena  and
      Bethard, Steven",
    booktitle = "Proceedings of the 2024 Conference of the North American Chapter of the Association for Computational Linguistics: Human Language Technologies (Volume 1: Long Papers)",
    month = jun,
    year = "2024",
    address = "Mexico City, Mexico",
    publisher = "Association for Computational Linguistics",
    url = "https://aclanthology.org/2024.naacl-long.327",
    doi = "10.18653/v1/2024.naacl-long.327",
    pages = "5872--5883",
}

@inproceedings{zeng2024evaluating,
title={Evaluating Large Language Models at Evaluating Instruction Following},
author={Zhiyuan Zeng and Jiatong Yu and Tianyu Gao and Yu Meng and Tanya Goyal and Danqi Chen},
booktitle={The Twelfth International Conference on Learning Representations},
year={2024},
url={https://openreview.net/forum?id=tr0KidwPLc}
}

@inproceedings{Zhang2020BERTScore,
title={{BERTScore: Evaluating Text Generation with BERT}},
author={Tianyi Zhang and Varsha Kishore and Felix Wu and Kilian Q. Weinberger and Yoav Artzi},
booktitle={International Conference on Learning Representations},
year={2020},
url={https://openreview.net/forum?id=SkeHuCVFDr}
}

@inproceedings{
kim2024prometheus,
title={Prometheus: Inducing Fine-Grained Evaluation Capability in Language Models},
author={Seungone Kim and Jamin Shin and Yejin Cho and Joel Jang and Shayne Longpre and Hwaran Lee and Sangdoo Yun and Seongjin Shin and Sungdong Kim and James Thorne and Minjoon Seo},
booktitle={The Twelfth International Conference on Learning Representations},
year={2024},
url={https://openreview.net/forum?id=8euJaTveKw}
}

@inproceedings{geval2023,
    title = "{G}-Eval: {NLG} Evaluation using Gpt-4 with Better Human Alignment",
    author = "Liu, Yang  and
      Iter, Dan  and
      Xu, Yichong  and
      Wang, Shuohang  and
      Xu, Ruochen  and
      Zhu, Chenguang",
    editor = "Bouamor, Houda  and
      Pino, Juan  and
      Bali, Kalika",
    booktitle = "Proceedings of the 2023 Conference on Empirical Methods in Natural Language Processing",
    month = dec,
    year = "2023",
    address = "Singapore",
    publisher = "Association for Computational Linguistics",
    url = "https://aclanthology.org/2023.emnlp-main.153",
    doi = "10.18653/v1/2023.emnlp-main.153",
    pages = "2511--2522",
}

@inproceedings{zheng2024judging,
 author = {Zheng, Lianmin and Chiang, Wei-Lin and Sheng, Ying and Zhuang, Siyuan and Wu, Zhanghao and Zhuang, Yonghao and Lin, Zi and Li, Zhuohan and Li, Dacheng and Xing, Eric and Zhang, Hao and Gonzalez, Joseph E and Stoica, Ion},
 booktitle = {Advances in Neural Information Processing Systems},
 editor = {A. Oh and T. Naumann and A. Globerson and K. Saenko and M. Hardt and S. Levine},
 pages = {46595--46623},
 publisher = {Curran Associates, Inc.},
 title = {{Judging LLM-as-a-Judge with MT-Bench and Chatbot Arena}},
 url = {https://proceedings.neurips.cc/paper_files/paper/2023/file/91f18a1287b398d378ef22505bf41832-Paper-Datasets_and_Benchmarks.pdf},
 volume = {36},
 year = {2023}
}

@inproceedings{cot2022neurips,
 author = {Wei, Jason and Wang, Xuezhi and Schuurmans, Dale and Bosma, Maarten and ichter, brian and Xia, Fei and Chi, Ed and Le, Quoc V and Zhou, Denny},
 booktitle = {Advances in Neural Information Processing Systems},
 editor = {S. Koyejo and S. Mohamed and A. Agarwal and D. Belgrave and K. Cho and A. Oh},
 pages = {24824--24837},
 publisher = {Curran Associates, Inc.},
 title = {Chain-of-Thought Prompting Elicits Reasoning in Large Language Models},
 url = {https://proceedings.neurips.cc/paper_files/paper/2022/file/9d5609613524ecf4f15af0f7b31abca4-Paper-Conference.pdf},
 volume = {35},
 year = {2022}
}

@InProceedings{pmlr-v202-zhou23g,
  title = 	 {{Controlled Text Generation with Natural Language Instructions}},
  author =       {Zhou, Wangchunshu and Jiang, Yuchen Eleanor and Wilcox, Ethan and Cotterell, Ryan and Sachan, Mrinmaya},
  booktitle = 	 {Proceedings of the 40th International Conference on Machine Learning},
  pages = 	 {42602--42613},
  year = 	 {2023},
  editor = 	 {Krause, Andreas and Brunskill, Emma and Cho, Kyunghyun and Engelhardt, Barbara and Sabato, Sivan and Scarlett, Jonathan},
  volume = 	 {202},
  series = 	 {Proceedings of Machine Learning Research},
  month = 	 {23--29 Jul},
  publisher =    {PMLR},
  url = 	 {https://proceedings.mlr.press/v202/zhou23g.html}
}

@inproceedings{lu-etal-2023-bounding,
    title = "{Bounding the Capabilities of Large Language Models in Open Text Generation with Prompt Constraints}",
    author = "Lu, Albert  and
      Zhang, Hongxin  and
      Zhang, Yanzhe  and
      Wang, Xuezhi  and
      Yang, Diyi",
    editor = "Vlachos, Andreas  and
      Augenstein, Isabelle",
    booktitle = "Findings of the Association for Computational Linguistics: EACL 2023",
    month = may,
    year = "2023",
    address = "Dubrovnik, Croatia",
    publisher = "Association for Computational Linguistics",
    url = "https://aclanthology.org/2023.findings-eacl.148",
    doi = "10.18653/v1/2023.findings-eacl.148",
    pages = "1982--2008",
}

@inproceedings{sun-etal-2023-evaluating,
    title = "{Evaluating Large Language Models on Controlled Generation Tasks}",
    author = "Sun, Jiao  and
      Tian, Yufei  and
      Zhou, Wangchunshu  and
      Xu, Nan  and
      Hu, Qian  and
      Gupta, Rahul  and
      Wieting, John  and
      Peng, Nanyun  and
      Ma, Xuezhe",
    editor = "Bouamor, Houda  and
      Pino, Juan  and
      Bali, Kalika",
    booktitle = "Proceedings of the 2023 Conference on Empirical Methods in Natural Language Processing",
    month = dec,
    year = "2023",
    address = "Singapore",
    publisher = "Association for Computational Linguistics",
    url = "https://aclanthology.org/2023.emnlp-main.190",
    doi = "10.18653/v1/2023.emnlp-main.190",
    pages = "3155--3168",
}

@inproceedings{
yao2024collie,
title={{COLLIE}: Systematic Construction of Constrained Text Generation Tasks},
author={Shunyu Yao and Howard Chen and Austin W. Hanjie and Runzhe Yang and Karthik R Narasimhan},
booktitle={The Twelfth International Conference on Learning Representations},
year={2024},
url={https://openreview.net/forum?id=kxgSlyirUZ}
}

@inproceedings{jiang2023followbench,
    title = "{F}ollow{B}ench: A Multi-level Fine-grained Constraints Following Benchmark for Large Language Models",
    author = "Jiang, Yuxin  and
      Wang, Yufei  and
      Zeng, Xingshan  and
      Zhong, Wanjun  and
      Li, Liangyou  and
      Mi, Fei  and
      Shang, Lifeng  and
      Jiang, Xin  and
      Liu, Qun  and
      Wang, Wei",
    editor = "Ku, Lun-Wei  and
      Martins, Andre  and
      Srikumar, Vivek",
    booktitle = "Proceedings of the 62nd Annual Meeting of the Association for Computational Linguistics (Volume 1: Long Papers)",
    month = aug,
    year = "2024",
    address = "Bangkok, Thailand",
    publisher = "Association for Computational Linguistics",
    url = "https://aclanthology.org/2024.acl-long.257",
    pages = "4667--4688",
}

@article{zhou2023instruction,
  title={Instruction-following evaluation for large language models},
  author={Zhou, Jeffrey and Lu, Tianjian and Mishra, Swaroop and Brahma, Siddhartha and Basu, Sujoy and Luan, Yi and Zhou, Denny and Hou, Le},
  journal={arXiv preprint arXiv:2311.07911},
  year={2023},
  url={https://arxiv.org/abs/2311.07911}
}

@article{pan2023automatically,
    title = "{Automatically Correcting Large Language Models: Surveying the Landscape of Diverse Automated Correction Strategies}",
    author = "Pan, Liangming  and
      Saxon, Michael  and
      Xu, Wenda  and
      Nathani, Deepak  and
      Wang, Xinyi  and
      Wang, William Yang",
    journal = "Transactions of the Association for Computational Linguistics",
    volume = "12",
    year = "2024",
    address = "Cambridge, MA",
    publisher = "MIT Press",
    url = "https://aclanthology.org/2024.tacl-1.27",
    doi = "10.1162/tacl_a_00660",
    pages = "484--506",
}

@inproceedings{Selfrefine2023,
 author = {Madaan, Aman and Tandon, Niket and Gupta, Prakhar and Hallinan, Skyler and Gao, Luyu and Wiegreffe, Sarah and Alon, Uri and Dziri, Nouha and Prabhumoye, Shrimai and Yang, Yiming and Gupta, Shashank and Majumder, Bodhisattwa Prasad and Hermann, Katherine and Welleck, Sean and Yazdanbakhsh, Amir and Clark, Peter},
 booktitle = {Advances in Neural Information Processing Systems},
 editor = {A. Oh and T. Naumann and A. Globerson and K. Saenko and M. Hardt and S. Levine},
 pages = {46534--46594},
 publisher = {Curran Associates, Inc.},
 title = {Self-Refine: Iterative Refinement with Self-Feedback},
 url = {https://proceedings.neurips.cc/paper_files/paper/2023/file/91edff07232fb1b55a505a9e9f6c0ff3-Paper-Conference.pdf},
 volume = {36},
 year = {2023}
}

@inproceedings{saha2023branch,
    title = {{Branch-Solve-Merge Improves Large Language Model Evaluation and Generation}},
    author = "Saha, Swarnadeep  and
      Levy, Omer  and
      Celikyilmaz, Asli  and
      Bansal, Mohit  and
      Weston, Jason  and
      Li, Xian",
    editor = "Duh, Kevin  and
      Gomez, Helena  and
      Bethard, Steven",
    booktitle = "Proceedings of the 2024 Conference of the North American Chapter of the Association for Computational Linguistics: Human Language Technologies (Volume 1: Long Papers)",
    month = jun,
    year = "2024",
    address = "Mexico City, Mexico",
    publisher = "Association for Computational Linguistics",
    url = "https://aclanthology.org/2024.naacl-long.462",
    doi = "10.18653/v1/2024.naacl-long.462",
    pages = "8352--8370",
}

@article{weston2023system,
  title={System 2 attention (is something you might need too)},
  author={Weston, Jason and Sukhbaatar, Sainbayar},
  journal={arXiv preprint arXiv:2311.11829},
  year={2023}
}

@article{deng2023rephrase,
  title={Rephrase and respond: Let large language models ask better questions for themselves},
  author={Deng, Yihe and Zhang, Weitong and Chen, Zixiang and Gu, Quanquan},
  journal={arXiv preprint arXiv:2311.04205},
  year={2023}
}

@article{achiam2023gpt,
  title={{GPT-4 Technical Report}},
  author={Achiam, Josh and Adler, Steven and Agarwal, Sandhini and Ahmad, Lama and Akkaya, Ilge and Aleman, Florencia Leoni and Almeida, Diogo and Altenschmidt, Janko and Altman, Sam and Anadkat, Shyamal and others},
  journal={arXiv preprint arXiv:2303.08774},
  year={2023},
  url={https://arxiv.org/abs/2303.08774}
}

@article{jiang2023mistral,
  title={Mistral 7B},
  author={Jiang, Albert Q and Sablayrolles, Alexandre and Mensch, Arthur and Bamford, Chris and Chaplot, Devendra Singh and Casas, Diego de las and Bressand, Florian and Lengyel, Gianna and Lample, Guillaume and Saulnier, Lucile and others},
  journal={arXiv preprint arXiv:2310.06825},
  year={2023},
  url={https://arxiv.org/abs/2310.06825}
}

@misc{vicuna2023,
    title = {{Vicuna: An Open-Source Chatbot Impressing GPT-4 with 90\%* ChatGPT Quality}},
    url = {https://lmsys.org/blog/2023-03-30-vicuna/},
    author = {Chiang, Wei-Lin and Li, Zhuohan and Lin, Zi and Sheng, Ying and Wu, Zhanghao and Zhang, Hao and Zheng, Lianmin and Zhuang, Siyuan and Zhuang, Yonghao and Gonzalez, Joseph E. and Stoica, Ion and Xing, Eric P.},
    month = {March},
    year = {2023}
}

@article{Qin2024InFoBenchEI,
  title={InFoBench: Evaluating Instruction Following Ability in Large Language Models},
  author={Yiwei Qin and Kaiqiang Song and Yebowen Hu and Wenlin Yao and Sangwoo Cho and Xiaoyang Wang and Xuansheng Wu and Fei Liu and Pengfei Liu and Dong Yu},
  journal={ArXiv},
  year={2024},
  volume={abs/2401.03601},
  url={https://api.semanticscholar.org/CorpusID:266844311}
}

@inproceedings{tunstall2023zephyr,
    title={{Zephyr: Direct Distillation of LM Alignment}},
    author={Lewis Tunstall and Edward Emanuel Beeching and Nathan Lambert and Nazneen Rajani and Kashif Rasul and Younes Belkada and Shengyi Huang and Leandro Von Werra and Cl{\'e}mentine Fourrier and Nathan Habib and Nathan Sarrazin and Omar Sanseviero and Alexander M Rush and Thomas Wolf},
    booktitle={First Conference on Language Modeling},
    year={2024},
    url={https://openreview.net/forum?id=aKkAwZB6JV}
}

@inproceedings{hu2021lora,
    title={Lo{RA}: Low-Rank Adaptation of Large Language Models},
    author={Edward J Hu and yelong shen and Phillip Wallis and Zeyuan Allen-Zhu and Yuanzhi Li and Shean Wang and Lu Wang and Weizhu Chen},
    booktitle={International Conference on Learning Representations},
    year={2022},
    url={https://openreview.net/forum?id=nZeVKeeFYf9}
}

@article{Huang2023LargeLM,
  title={Large Language Models Cannot Self-Correct Reasoning Yet},
  author={Jie Huang and Xinyun Chen and Swaroop Mishra and Huaixiu Steven Zheng and Adams Wei Yu and Xinying Song and Denny Zhou},
  journal={ArXiv},
  year={2023},
  volume={abs/2310.01798},
  url={https://api.semanticscholar.org/CorpusID:263609132}
}

@inproceedings{wolf-etal-2020-transformers,
    title = "Transformers: State-of-the-Art Natural Language Processing",
    author = "Wolf, Thomas  and
      Debut, Lysandre  and
      Sanh, Victor  and
      Chaumond, Julien  and
      Delangue, Clement  and
      Moi, Anthony  and
      Cistac, Pierric  and
      Rault, Tim  and
      Louf, Remi  and
      Funtowicz, Morgan  and
      Davison, Joe  and
      Shleifer, Sam  and
      von Platen, Patrick  and
      Ma, Clara  and
      Jernite, Yacine  and
      Plu, Julien  and
      Xu, Canwen  and
      Le Scao, Teven  and
      Gugger, Sylvain  and
      Drame, Mariama  and
      Lhoest, Quentin  and
      Rush, Alexander",
    editor = "Liu, Qun  and
      Schlangen, David",
    booktitle = "Proceedings of the 2020 Conference on Empirical Methods in Natural Language Processing: System Demonstrations",
    month = oct,
    year = "2020",
    address = "Online",
    publisher = "Association for Computational Linguistics",
    url = "https://aclanthology.org/2020.emnlp-demos.6",
    doi = "10.18653/v1/2020.emnlp-demos.6",
    pages = "38--45",
}

@article{open-llm-leaderboard-v1,
  author = {Edward Beeching and Clémentine Fourrier and Nathan Habib and Sheon Han and Nathan Lambert and Nazneen Rajani and Omar Sanseviero and Lewis Tunstall and Thomas Wolf},
  title = {{Open LLM Leaderboard}},
  year = {2023},
  journal = {Hugging Face},
  url = "https: //huggingface.co/spaces/HuggingFaceH4/open_llm_leaderboard"
}

@misc{chiang2024chatbot,
    title={Chatbot Arena: An Open Platform for Evaluating LLMs by Human Preference},
    author={Wei-Lin Chiang and Lianmin Zheng and Ying Sheng and Anastasios Nikolas Angelopoulos and Tianle Li and Dacheng Li and Hao Zhang and Banghua Zhu and Michael Jordan and Joseph E. Gonzalez and Ion Stoica},
    year={2024},
    eprint={2403.04132},
    archivePrefix={arXiv},
    primaryClass={cs.AI}
}

@article{verga2024replacing,
  title={Replacing Judges with Juries: Evaluating LLM Generations with a Panel of Diverse Models},
  author={Verga, Pat and Hofstatter, Sebastian and Althammer, Sophia and Su, Yixuan and Piktus, Aleksandra and Arkhangorodsky, Arkady and Xu, Minjie and White, Naomi and Lewis, Patrick},
  journal={arXiv preprint arXiv:2404.18796},
  year={2024},
url={https://arxiv.org/abs/2404.18796}
}

@article{kim2024prometheus2,
  title={Prometheus 2: An open source language model specialized in evaluating other language models},
  author={Kim, Seungone and Suk, Juyoung and Longpre, Shayne and Lin, Bill Yuchen and Shin, Jamin and Welleck, Sean and Neubig, Graham and Lee, Moontae and Lee, Kyungjae and Seo, Minjoon},
  journal={arXiv preprint arXiv:2405.01535},
  year={2024},
  url={https://arxiv.org/abs/2405.01535}
}

@inproceedings{truong-etal-2023-language,
    title = "Language models are not naysayers: an analysis of language models on negation benchmarks",
    author = "Truong, Thinh Hung  and
      Baldwin, Timothy  and
      Verspoor, Karin  and
      Cohn, Trevor",
    editor = "Palmer, Alexis  and
      Camacho-collados, Jose",
    booktitle = "Proceedings of the 12th Joint Conference on Lexical and Computational Semantics (*SEM 2023)",
    month = jul,
    year = "2023",
    address = "Toronto, Canada",
    publisher = "Association for Computational Linguistics",
    url = "https://aclanthology.org/2023.starsem-1.10",
    doi = "10.18653/v1/2023.starsem-1.10",
    pages = "101--114",
}

@inproceedings{garcia-ferrero-etal-2023-dataset,
    title = "This is not a Dataset: A Large Negation Benchmark to Challenge Large Language Models",
    author = "Garc{\'\i}a-Ferrero, Iker  and
      Altuna, Bego{\~n}a  and
      Alvez, Javier  and
      Gonzalez-Dios, Itziar  and
      Rigau, German",
    editor = "Bouamor, Houda  and
      Pino, Juan  and
      Bali, Kalika",
    booktitle = "Proceedings of the 2023 Conference on Empirical Methods in Natural Language Processing",
    month = dec,
    year = "2023",
    address = "Singapore",
    publisher = "Association for Computational Linguistics",
    url = "https://aclanthology.org/2023.emnlp-main.531",
    doi = "10.18653/v1/2023.emnlp-main.531",
    pages = "8596--8615",
}

@inproceedings{lin-etal-2020-commongen,
    title = "{C}ommon{G}en: A Constrained Text Generation Challenge for Generative Commonsense Reasoning",
    author = "Lin, Bill Yuchen  and
      Zhou, Wangchunshu  and
      Shen, Ming  and
      Zhou, Pei  and
      Bhagavatula, Chandra  and
      Choi, Yejin  and
      Ren, Xiang",
    editor = "Cohn, Trevor  and
      He, Yulan  and
      Liu, Yang",
    booktitle = "Findings of the Association for Computational Linguistics: EMNLP 2020",
    month = nov,
    year = "2020",
    address = "Online",
    publisher = "Association for Computational Linguistics",
    url = "https://aclanthology.org/2020.findings-emnlp.165",
    doi = "10.18653/v1/2020.findings-emnlp.165",
    pages = "1823--1840",
}

@inproceedings{numericalLLM, author = {Sui, Yuan and Zhou, Mengyu and Zhou, Mingjie and Han, Shi and Zhang, Dongmei}, title = {Table Meets LLM: Can Large Language Models Understand Structured Table Data? A Benchmark and Empirical Study}, year = {2024}, isbn = {9798400703713}, publisher = {Association for Computing Machinery}, address = {New York, NY, USA}, url = {https://doi.org/10.1145/3616855.3635752}, doi = {10.1145/3616855.3635752}, booktitle = {Proceedings of the 17th ACM International Conference on Web Search and Data Mining}, pages = {645–654}, numpages = {10}, location = {Merida, Mexico}, series = {WSDM '24} }

@inproceedings{huang-etal-2023-incorporating,
    title = "Incorporating Worker Perspectives into {MT}urk Annotation Practices for {NLP}",
    author = "Huang, Olivia  and
      Fleisig, Eve  and
      Klein, Dan",
    editor = "Bouamor, Houda  and
      Pino, Juan  and
      Bali, Kalika",
    booktitle = "Proceedings of the 2023 Conference on Empirical Methods in Natural Language Processing",
    month = dec,
    year = "2023",
    address = "Singapore",
    publisher = "Association for Computational Linguistics",
    url = "https://aclanthology.org/2023.emnlp-main.64",
    doi = "10.18653/v1/2023.emnlp-main.64",
    pages = "1010--1028",
}

@inproceedings{
hendrycks2021measuring,
title={Measuring Massive Multitask Language Understanding},
author={Dan Hendrycks and Collin Burns and Steven Basart and Andy Zou and Mantas Mazeika and Dawn Song and Jacob Steinhardt},
booktitle={International Conference on Learning Representations},
year={2021},
url={https://openreview.net/forum?id=d7KBjmI3GmQ}
}

@article{winogrande, author = {Sakaguchi, Keisuke and Bras, Ronan Le and Bhagavatula, Chandra and Choi, Yejin}, title = {WinoGrande: an adversarial winograd schema challenge at scale}, year = {2021}, issue_date = {September 2021}, publisher = {Association for Computing Machinery}, address = {New York, NY, USA}, volume = {64}, number = {9}, issn = {0001-0782}, url = {https://doi.org/10.1145/3474381}, doi = {10.1145/3474381}, journal = {Commun. ACM}, month = {aug}, pages = {99–106}, numpages = {8} }

@article{gsm8k,
  author       = {Karl Cobbe and
                  Vineet Kosaraju and
                  Mohammad Bavarian and
                  Mark Chen and
                  Heewoo Jun and
                  Lukasz Kaiser and
                  Matthias Plappert and
                  Jerry Tworek and
                  Jacob Hilton and
                  Reiichiro Nakano and
                  Christopher Hesse and
                  John Schulman},
  title        = {Training Verifiers to Solve Math Word Problems},
  journal      = {CoRR},
  volume       = {abs/2110.14168},
  year         = {2021},
  url          = {https://arxiv.org/abs/2110.14168},
  eprinttype    = {arXiv},
  eprint       = {2110.14168},
}

@article{
srivastava2023beyond,
title={Beyond the Imitation Game: Quantifying and extrapolating the capabilities of language models},
author={Aarohi Srivastava and Abhinav Rastogi and Abhishek Rao and Abu Awal Md Shoeb and Abubakar Abid and Adam Fisch and Adam R. Brown and Adam Santoro and Aditya Gupta and Adri{\`a} Garriga-Alonso and Agnieszka Kluska and Aitor Lewkowycz and Akshat Agarwal and Alethea Power and Alex Ray and Alex Warstadt and Alexander W. Kocurek and Ali Safaya and Ali Tazarv and Alice Xiang and Alicia Parrish and Allen Nie and Aman Hussain and Amanda Askell and Amanda Dsouza and Ambrose Slone and Ameet Rahane and Anantharaman S. Iyer and Anders Johan Andreassen and Andrea Madotto and Andrea Santilli and Andreas Stuhlm{\"u}ller and Andrew M. Dai and Andrew La and Andrew Lampinen and Andy Zou and Angela Jiang and Angelica Chen and Anh Vuong and Animesh Gupta and Anna Gottardi and Antonio Norelli and Anu Venkatesh and Arash Gholamidavoodi and Arfa Tabassum and Arul Menezes and Arun Kirubarajan and Asher Mullokandov and Ashish Sabharwal and Austin Herrick and Avia Efrat and Aykut Erdem and Ayla Karaka{\c{s}} and others},
journal={Transactions on Machine Learning Research},
issn={2835-8856},
year={2023},
url={https://openreview.net/forum?id=uyTL5Bvosj},
note={}
}

@inproceedings{zellers-etal-2019-hellaswag,
    title = "{H}ella{S}wag: Can a Machine Really Finish Your Sentence?",
    author = "Zellers, Rowan  and
      Holtzman, Ari  and
      Bisk, Yonatan  and
      Farhadi, Ali  and
      Choi, Yejin",
    editor = "Korhonen, Anna  and
      Traum, David  and
      M{\`a}rquez, Llu{\'\i}s",
    booktitle = "Proceedings of the 57th Annual Meeting of the Association for Computational Linguistics",
    month = jul,
    year = "2019",
    address = "Florence, Italy",
    publisher = "Association for Computational Linguistics",
    url = "https://aclanthology.org/P19-1472",
    doi = "10.18653/v1/P19-1472",
    pages = "4791--4800",
}

@article{selfcorrectionsurvey2024,
  title={{When Can LLMs Actually Correct Their Own Mistakes? A Critical Survey of Self-Correction of LLMs}},
  author={Kamoi, Ryo and Zhang, Yusen and Zhang, Nan and Han, Jiawei and Zhang, Rui},
  journal={arXiv preprint arXiv:2406.01297},
  year={2024},
  url={https://arxiv.org/abs/2406.01297},
}

@inproceedings{zhong-etal-2024-benchmarking,
    title = "Benchmarking and Improving Compositional Generalization of Multi-aspect Controllable Text Generation",
    author = "Zhong, Tianqi  and
      Li, Zhaoyi  and
      Wang, Quan  and
      Song, Linqi  and
      Wei, Ying  and
      Lian, Defu  and
      Mao, Zhendong",
    editor = "Ku, Lun-Wei  and
      Martins, Andre  and
      Srikumar, Vivek",
    booktitle = "Proceedings of the 62nd Annual Meeting of the Association for Computational Linguistics (Volume 1: Long Papers)",
    month = aug,
    year = "2024",
    address = "Bangkok, Thailand",
    publisher = "Association for Computational Linguistics",
    url = "https://aclanthology.org/2024.acl-long.351",
    pages = "6486--6517",
}

@inproceedings{manino-etal-2022-systematicity,
    title = "Systematicity, Compositionality and Transitivity of Deep {NLP} Models: a Metamorphic Testing Perspective",
    author = "Manino, Edoardo  and
      Rozanova, Julia  and
      Carvalho, Danilo  and
      Freitas, Andre  and
      Cordeiro, Lucas",
    editor = "Muresan, Smaranda  and
      Nakov, Preslav  and
      Villavicencio, Aline",
    booktitle = "Findings of the Association for Computational Linguistics: ACL 2022",
    month = may,
    year = "2022",
    address = "Dublin, Ireland",
    publisher = "Association for Computational Linguistics",
    url = "https://aclanthology.org/2022.findings-acl.185",
    doi = "10.18653/v1/2022.findings-acl.185",
    pages = "2355--2366",
}

@inproceedings{dankers-etal-2022-paradox,
    title = "The Paradox of the Compositionality of Natural Language: A Neural Machine Translation Case Study",
    author = "Dankers, Verna  and
      Bruni, Elia  and
      Hupkes, Dieuwke",
    editor = "Muresan, Smaranda  and
      Nakov, Preslav  and
      Villavicencio, Aline",
    booktitle = "Proceedings of the 60th Annual Meeting of the Association for Computational Linguistics (Volume 1: Long Papers)",
    month = may,
    year = "2022",
    address = "Dublin, Ireland",
    publisher = "Association for Computational Linguistics",
    url = "https://aclanthology.org/2022.acl-long.286",
    doi = "10.18653/v1/2022.acl-long.286",
    pages = "4154--4175",
}

@inproceedings{treeofthoughts2023,
 author = {Yao, Shunyu and Yu, Dian and Zhao, Jeffrey and Shafran, Izhak and Griffiths, Tom and Cao, Yuan and Narasimhan, Karthik},
 booktitle = {Advances in Neural Information Processing Systems},
 editor = {A. Oh and T. Naumann and A. Globerson and K. Saenko and M. Hardt and S. Levine},
 pages = {11809--11822},
 publisher = {Curran Associates, Inc.},
 title = {{Tree of Thoughts: Deliberate Problem Solving with Large Language Models}},
 url = {https://proceedings.neurips.cc/paper_files/paper/2023/file/271db9922b8d1f4dd7aaef84ed5ac703-Paper-Conference.pdf},
 volume = {36},
 year = {2023}
}

@inproceedings{lee-kim-2023-recursion,
    title = "Recursion of Thought: A Divide-and-Conquer Approach to Multi-Context Reasoning with Language Models",
    author = "Lee, Soochan  and
      Kim, Gunhee",
    editor = "Rogers, Anna  and
      Boyd-Graber, Jordan  and
      Okazaki, Naoaki",
    booktitle = "Findings of the Association for Computational Linguistics: ACL 2023",
    month = jul,
    year = "2023",
    address = "Toronto, Canada",
    publisher = "Association for Computational Linguistics",
    url = "https://aclanthology.org/2023.findings-acl.40",
    doi = "10.18653/v1/2023.findings-acl.40",
    pages = "623--658",
}

@inproceedings{qi-etal-2023-art,
    title = "The Art of {SOCRATIC} {QUESTIONING}: Recursive Thinking with Large Language Models",
    author = "Qi, Jingyuan  and
      Xu, Zhiyang  and
      Shen, Ying  and
      Liu, Minqian  and
      Jin, Di  and
      Wang, Qifan  and
      Huang, Lifu",
    editor = "Bouamor, Houda  and
      Pino, Juan  and
      Bali, Kalika",
    booktitle = "Proceedings of the 2023 Conference on Empirical Methods in Natural Language Processing",
    month = dec,
    year = "2023",
    address = "Singapore",
    publisher = "Association for Computational Linguistics",
    url = "https://aclanthology.org/2023.emnlp-main.255",
    doi = "10.18653/v1/2023.emnlp-main.255",
    pages = "4177--4199",
}

@inproceedings{min-etal-2023-factscore,
    title = "{FA}ct{S}core: Fine-grained Atomic Evaluation of Factual Precision in Long Form Text Generation",
    author = "Min, Sewon  and
      Krishna, Kalpesh  and
      Lyu, Xinxi  and
      Lewis, Mike  and
      Yih, Wen-tau  and
      Koh, Pang  and
      Iyyer, Mohit  and
      Zettlemoyer, Luke  and
      Hajishirzi, Hannaneh",
    editor = "Bouamor, Houda  and
      Pino, Juan  and
      Bali, Kalika",
    booktitle = "Proceedings of the 2023 Conference on Empirical Methods in Natural Language Processing",
    month = dec,
    year = "2023",
    address = "Singapore",
    publisher = "Association for Computational Linguistics",
    url = "https://aclanthology.org/2023.emnlp-main.741",
    doi = "10.18653/v1/2023.emnlp-main.741",
    pages = "12076--12100",
}

@article{zhang2024towards,
  title={Towards fine-grained citation evaluation in generated text: A comparative analysis of faithfulness metrics},
  author={Zhang, Weijia and Aliannejadi, Mohammad and Yuan, Yifei and Pei, Jiahuan and Huang, Jia-Hong and Kanoulas, Evangelos},
  journal={arXiv preprint arXiv:2406.15264},
  year={2024},
  url={https://arxiv.org/pdf/2406.15264}
}

@article{huang2024empirical,
  title={{On the Limitations of Fine-tuned Judge Models for LLM Evaluation}},
  author={Huang, Hui and Qu, Yingqi and Liu, Jing and Yang, Muyun and Zhao, Tiejun},
  journal={arXiv preprint arXiv:2403.02839},
  year={2024},
  url={https://arxiv.org/pdf/2403.02839},
}

@inproceedings{song-etal-2024-finesure,
    title = "{F}ine{S}ur{E}: Fine-grained Summarization Evaluation using {LLM}s",
    author = "Song, Hwanjun  and
      Su, Hang  and
      Shalyminov, Igor  and
      Cai, Jason  and
      Mansour, Saab",
    editor = "Ku, Lun-Wei  and
      Martins, Andre  and
      Srikumar, Vivek",
    booktitle = "Proceedings of the 62nd Annual Meeting of the Association for Computational Linguistics (Volume 1: Long Papers)",
    month = aug,
    year = "2024",
    address = "Bangkok, Thailand",
    publisher = "Association for Computational Linguistics",
    url = "https://aclanthology.org/2024.acl-long.51",
    pages = "906--922",
}

@InProceedings{Hu_2023_ICCV,
    author    = {Hu, Yushi and Liu, Benlin and Kasai, Jungo and Wang, Yizhong and Ostendorf, Mari and Krishna, Ranjay and Smith, Noah A.},
    title     = {TIFA: Accurate and Interpretable Text-to-Image Faithfulness Evaluation with Question Answering},
    booktitle = {Proceedings of the IEEE/CVF International Conference on Computer Vision (ICCV)},
    month     = {October},
    year      = {2023},
    pages     = {20406-20417},
    url = {https://openaccess.thecvf.com/content/ICCV2023/papers/Hu_TIFA_Accurate_and_Interpretable_Text-to-Image_Faithfulness_Evaluation_with_Question_Answering_ICCV_2023_paper.pdf},
}

@misc{faithscore,
      title={FAITHSCORE: Evaluating Hallucinations in Large Vision-Language Models}, 
      author={Liqiang Jing and Ruosen Li and Yunmo Chen and Mengzhao Jia and Xinya Du},
      year={2023},
      eprint={2311.01477},
      archivePrefix={arXiv},
      url={https://openreview.net/pdf?id=E6uTejGgth},
      primaryClass={cs.CV}
}

@inproceedings{li-etal-2023-evaluating,
    title = "Evaluating Object Hallucination in Large Vision-Language Models",
    author = "Li, Yifan  and
      Du, Yifan  and
      Zhou, Kun  and
      Wang, Jinpeng  and
      Zhao, Xin  and
      Wen, Ji-Rong",
    editor = "Bouamor, Houda  and
      Pino, Juan  and
      Bali, Kalika",
    booktitle = "Proceedings of the 2023 Conference on Empirical Methods in Natural Language Processing",
    month = dec,
    year = "2023",
    address = "Singapore",
    publisher = "Association for Computational Linguistics",
    url = "https://aclanthology.org/2023.emnlp-main.20",
    doi = "10.18653/v1/2023.emnlp-main.20",
    pages = "292--305",
}

@inproceedings{ke-etal-2024-critiquellm,
    title = "{C}ritique{LLM}: Towards an Informative Critique Generation Model for Evaluation of Large Language Model Generation",
    author = "Ke, Pei  and
      Wen, Bosi  and
      Feng, Andrew  and
      Liu, Xiao  and
      Lei, Xuanyu  and
      Cheng, Jiale  and
      Wang, Shengyuan  and
      Zeng, Aohan  and
      Dong, Yuxiao  and
      Wang, Hongning  and
      Tang, Jie  and
      Huang, Minlie",
    editor = "Ku, Lun-Wei  and
      Martins, Andre  and
      Srikumar, Vivek",
    booktitle = "Proceedings of the 62nd Annual Meeting of the Association for Computational Linguistics (Volume 1: Long Papers)",
    month = aug,
    year = "2024",
    address = "Bangkok, Thailand",
    publisher = "Association for Computational Linguistics",
    url = "https://aclanthology.org/2024.acl-long.704",
    pages = "13034--13054",
}

@inproceedings{magister-etal-2023-teaching,
    title = "Teaching Small Language Models to Reason",
    author = "Magister, Lucie Charlotte  and
      Mallinson, Jonathan  and
      Adamek, Jakub  and
      Malmi, Eric  and
      Severyn, Aliaksei",
    editor = "Rogers, Anna  and
      Boyd-Graber, Jordan  and
      Okazaki, Naoaki",
    booktitle = "Proceedings of the 61st Annual Meeting of the Association for Computational Linguistics (Volume 2: Short Papers)",
    month = jul,
    year = "2023",
    address = "Toronto, Canada",
    publisher = "Association for Computational Linguistics",
    url = "https://aclanthology.org/2023.acl-short.151",
    doi = "10.18653/v1/2023.acl-short.151",
    pages = "1773--1781",
}

@article{shridhar2023screws,
  title={Screws: A modular framework for reasoning with revisions},
  author={Shridhar, Kumar and Jhamtani, Harsh and Fang, Hao and Van Durme, Benjamin and Eisner, Jason and Xia, Patrick},
  journal={arXiv preprint arXiv:2309.13075},
  year={2023},
  url={https://arxiv.org/pdf/2309.13075}
}

@article{wang2024speculative,
  title={Speculative RAG: Enhancing Retrieval Augmented Generation through Drafting},
  author={Wang, Zilong and Wang, Zifeng and Le, Long and Zheng, Huaixiu Steven and Mishra, Swaroop and Perot, Vincent and Zhang, Yuwei and Mattapalli, Anush and Taly, Ankur and Shang, Jingbo and others},
  journal={arXiv preprint arXiv:2407.08223},
  year={2024},
  url={https://arxiv.org/abs/2407.08223}
}

@article{morris2001measurement,
  title={Measurement and instrumentation principles},
  author={Morris, Alan S},
  journal={Measurement Science and Technology},
  volume={12},
  number={10},
  pages={1743--1744},
 url={http://llrc.mcast.edu.mt/digitalversion/table_of_contents_1274.pdf},
  year={2001}
}

@book{taylor1997introduction,
  title={Introduction To Error Analysis: The Study of Uncertainties in Physical Measurements},
  author={Taylor, J.R.},
  isbn={9780935702750},
  lccn={81051269},
  series={ASMSU/Spartans.4.Spartans Textbook},
  url={https://books.google.com/books?id=giFQcZub80oC},
  year={1997},
  publisher={University Science Books}
}

@book{british2022bs,
  title={BS ISO 5725-1. Accuracy (trueness and Precision) of Measurement Methods and Results: Part 1. General principles and definitions},
  author={British Standards, Institution},
  number={pt. 1},
  url={https://books.google.com/books?id=f5Q2zwEACAAJ},
  year={2022},
  publisher={British Standards Institution}
}

@misc{alpaca_eval,
  author = {Xuechen Li and Tianyi Zhang and Yann Dubois and Rohan Taori and Ishaan Gulrajani and Carlos Guestrin and Percy Liang and Tatsunori B. Hashimoto },
  title = {AlpacaEval: An Automatic Evaluator of Instruction-following Models},
  year = {2023},
  month = {5},
  publisher = {GitHub},
  journal = {GitHub repository},
  url = {https://github.com/tatsu-lab/alpaca_eval}
}

@inproceedings{wang-etal-2023-self-instruct,
    title = "Self-Instruct: Aligning Language Models with Self-Generated Instructions",
    author = "Wang, Yizhong  and
      Kordi, Yeganeh  and
      Mishra, Swaroop  and
      Liu, Alisa  and
      Smith, Noah A.  and
      Khashabi, Daniel  and
      Hajishirzi, Hannaneh",
    editor = "Rogers, Anna  and
      Boyd-Graber, Jordan  and
      Okazaki, Naoaki",
    booktitle = "Proceedings of the 61st Annual Meeting of the Association for Computational Linguistics (Volume 1: Long Papers)",
    month = jul,
    year = "2023",
    address = "Toronto, Canada",
    publisher = "Association for Computational Linguistics",
    url = "https://aclanthology.org/2023.acl-long.754",
    doi = "10.18653/v1/2023.acl-long.754",
    pages = "13484--13508",
}

@inproceedings{khan2024debating,
title={Debating with More Persuasive {LLM}s Leads to More Truthful Answers},
author={Akbir Khan and John Hughes and Dan Valentine and Laura Ruis and Kshitij Sachan and Ansh Radhakrishnan and Edward Grefenstette and Samuel R. Bowman and Tim Rockt{\"a}schel and Ethan Perez},
booktitle={Forty-first International Conference on Machine Learning},
year={2024},
url={https://openreview.net/forum?id=iLCZtl7FTa}
}

@article{chen2024more,
  title={Are more llm calls all you need? towards scaling laws of compound inference systems},
  author={Chen, Lingjiao and Davis, Jared Quincy and Hanin, Boris and Bailis, Peter and Stoica, Ion and Zaharia, Matei and Zou, James},
  journal={arXiv preprint arXiv:2403.02419},
  year={2024},
  url={https://arxiv.org/abs/2403.02419}
}

@misc{deng2024implicit,
title={Implicit Chain of Thought Reasoning via Knowledge Distillation},
author={Yuntian Deng and Kiran Prasad and Roland Fernandez and Paul Smolensky and Vishrav Chaudhary and Stuart Shieber},
year={2024},
url={https://openreview.net/forum?id=9cumTvvlHG}
}

@article{yu2024distilling,
  title={Distilling system 2 into system 1},
  author={Yu, Ping and Xu, Jing and Weston, Jason and Kulikov, Ilia},
  journal={arXiv preprint arXiv:2407.06023},
  year={2024},
  url={https://arxiv.org/abs/2407.06023}
}

@article{multiprompt2024,
    author = {Mizrahi, Moran and Kaplan, Guy and Malkin, Dan and Dror, Rotem and Shahaf, Dafna and Stanovsky, Gabriel},
    title = "{State of What Art? A Call for Multi-Prompt LLM Evaluation}",
    journal = {Transactions of the Association for Computational Linguistics},
    volume = {12},
    pages = {933-949},
    year = {2024},
    month = {08},
    issn = {2307-387X},
    doi = {10.1162/tacl_a_00681},
    url = {https://doi.org/10.1162/tacl\_a\_00681},
    eprint = {https://direct.mit.edu/tacl/article-pdf/doi/10.1162/tacl\_a\_00681/2464098/tacl\_a\_00681.pdf},
}

@article{panickssery2024llm,
  title={Llm evaluators recognize and favor their own generations},
  author={Panickssery, Arjun and Bowman, Samuel R and Feng, Shi},
  journal={arXiv preprint arXiv:2404.13076},
  year={2024},
  url={https://arxiv.org/abs/2404.13076}
}

@inproceedings{
wang2023selfconsistency,
title={Self-Consistency Improves Chain of Thought Reasoning in Language Models},
author={Xuezhi Wang and Jason Wei and Dale Schuurmans and Quoc V Le and Ed H. Chi and Sharan Narang and Aakanksha Chowdhery and Denny Zhou},
booktitle={The Eleventh International Conference on Learning Representations },
year={2023},
url={https://openreview.net/forum?id=1PL1NIMMrw}
}

@inproceedings{
yao2024learning,
title={Learning From Correctness Without Prompting Makes {LLM} Efficient Reasoner},
author={Yuxuan Yao and Han Wu and Zhijiang Guo and Zhou Biyan and Jiahui Gao and Sichun Luo and Hanxu Hou and Xiaojin Fu and Linqi Song},
booktitle={First Conference on Language Modeling},
year={2024},
url={https://openreview.net/forum?id=dcbNzhVVQj}
}

@inproceedings{
chen2024universal,
title={Universal Self-Consistency for Large Language Models},
author={Xinyun Chen and Renat Aksitov and Uri Alon and Jie Ren and Kefan Xiao and Pengcheng Yin and Sushant Prakash and Charles Sutton and Xuezhi Wang and Denny Zhou},
booktitle={ICML 2024 Workshop on In-Context Learning},
year={2024},
url={https://openreview.net/forum?id=LjsjHF7nAN}
}

@inproceedings{
du2024improving,
title={Improving Factuality and Reasoning in Language Models through Multiagent Debate},
author={Yilun Du and Shuang Li and Antonio Torralba and Joshua B. Tenenbaum and Igor Mordatch},
booktitle={Forty-first International Conference on Machine Learning},
year={2024},
url={https://openreview.net/forum?id=zj7YuTE4t8}
}

@article{wen2024benchmarking,
  title={Benchmarking Complex Instruction-Following with Multiple Constraints Composition},
  author={Wen, Bosi and Ke, Pei and Gu, Xiaotao and Wu, Lindong and Huang, Hao and Zhou, Jinfeng and Li, Wenchuang and Hu, Binxin and Gao, Wendy and Xu, Jiaxin and others},
  journal={arXiv preprint arXiv:2407.03978},
  year={2024},
 url={https://arxiv.org/abs/2407.03978}
}

@inproceedings{he2024can,
  title={Can Large Language Models Understand Real-World Complex Instructions?},
  author={He, Qianyu and Zeng, Jie and Huang, Wenhao and Chen, Lina and Xiao, Jin and He, Qianxi and Zhou, Xunzhe and Liang, Jiaqing and Xiao, Yanghua},
  booktitle={Proceedings of the AAAI Conference on Artificial Intelligence},
  volume={38},
  number={16},
  pages={18188--18196},
  year={2024},
  url={https://ojs.aaai.org/index.php/AAAI/article/view/29777}
}

@article{zhang2024cfbench,
  title={CFBench: A Comprehensive Constraints-Following Benchmark for LLMs},
  author={Zhang, Tao and Shen, Yanjun and Luo, Wenjing and Zhang, Yan and Liang, Hao and Yang, Fan and Lin, Mingan and Qiao, Yujing and Chen, Weipeng and Cui, Bin and others},
  journal={arXiv preprint arXiv:2408.01122},
  year={2024},
  url={https://arxiv.org/abs/2408.01122}
}

@misc{selfee2023,
	author = {Ye, Seonghyeon and Jo, Yongrae and Kim, Doyoung and Kim, Sungdong and Hwang, Hyeonbin and Seo, Minjoon},
	title = {SelFee: Iterative Self-Revising LLM Empowered by Self-Feedback Generation},
	url = {https://kaistai.github.io/SelFee/},
	month = {May},
	year = {2023},
	howpublished = {Blog post}
}

@inproceedings{
schick2023peer,
title={{PEER}: A Collaborative Language Model},
author={Timo Schick and Jane A. Yu and Zhengbao Jiang and Fabio Petroni and Patrick Lewis and Gautier Izacard and Qingfei You and Christoforos Nalmpantis and Edouard Grave and Sebastian Riedel},
booktitle={The Eleventh International Conference on Learning Representations },
year={2023},
url={https://openreview.net/forum?id=KbYevcLjnc}
}

@article{saunders2022self,
  title={Self-critiquing models for assisting human evaluators},
  author={Saunders, William and Yeh, Catherine and Wu, Jeff and Bills, Steven and Ouyang, Long and Ward, Jonathan and Leike, Jan},
  journal={arXiv preprint arXiv:2206.05802},
  year={2022},
url={https://arxiv.org/pdf/2206.05802}
}

@inproceedings{volcano,
    title = "Volcano: Mitigating Multimodal Hallucination through Self-Feedback Guided Revision",
    author = "Lee, Seongyun  and
      Park, Sue Hyun  and
      Jo, Yongrae  and
      Seo, Minjoon",
    editor = "Duh, Kevin  and
      Gomez, Helena  and
      Bethard, Steven",
    booktitle = "Proceedings of the 2024 Conference of the North American Chapter of the Association for Computational Linguistics: Human Language Technologies (Volume 1: Long Papers)",
    month = jun,
    year = "2024",
    address = "Mexico City, Mexico",
    publisher = "Association for Computational Linguistics",
    url = "https://aclanthology.org/2024.naacl-long.23",
    doi = "10.18653/v1/2024.naacl-long.23",
    pages = "391--404",
}

@inproceedings{xu-etal-2023-instructscore,
    title = "{INSTRUCTSCORE}: Towards Explainable Text Generation Evaluation with Automatic Feedback",
    author = "Xu, Wenda  and
      Wang, Danqing  and
      Pan, Liangming  and
      Song, Zhenqiao  and
      Freitag, Markus  and
      Wang, William  and
      Li, Lei",
    editor = "Bouamor, Houda  and
      Pino, Juan  and
      Bali, Kalika",
    booktitle = "Proceedings of the 2023 Conference on Empirical Methods in Natural Language Processing",
    month = dec,
    year = "2023",
    address = "Singapore",
    publisher = "Association for Computational Linguistics",
    url = "https://aclanthology.org/2023.emnlp-main.365",
    doi = "10.18653/v1/2023.emnlp-main.365",
    pages = "5967--5994",
}

@inproceedings{
welleck2023generating,
title={Generating Sequences by Learning to Self-Correct},
author={Sean Welleck and Ximing Lu and Peter West and Faeze Brahman and Tianxiao Shen and Daniel Khashabi and Yejin Choi},
booktitle={The Eleventh International Conference on Learning Representations },
year={2023},
url={https://openreview.net/forum?id=hH36JeQZDaO}
}

@inproceedings{yang-etal-2022-re3,
    title = "Re3: Generating Longer Stories With Recursive Reprompting and Revision",
    author = "Yang, Kevin  and
      Tian, Yuandong  and
      Peng, Nanyun  and
      Klein, Dan",
    editor = "Goldberg, Yoav  and
      Kozareva, Zornitsa  and
      Zhang, Yue",
    booktitle = "Proceedings of the 2022 Conference on Empirical Methods in Natural Language Processing",
    month = dec,
    year = "2022",
    address = "Abu Dhabi, United Arab Emirates",
    publisher = "Association for Computational Linguistics",
    url = "https://aclanthology.org/2022.emnlp-main.296",
    doi = "10.18653/v1/2022.emnlp-main.296",
    pages = "4393--4479",
}

@inproceedings{
    gou2024critic,
    title={{CRITIC}: Large Language Models Can Self-Correct with Tool-Interactive Critiquing},
    author={Zhibin Gou and Zhihong Shao and Yeyun Gong and yelong shen and Yujiu Yang and Nan Duan and Weizhu Chen},
    booktitle={The Twelfth International Conference on Learning Representations},
    year={2024},
    url={https://openreview.net/forum?id=Sx038qxjek}
}

@article{varshney2023stitch,
  title={A stitch in time saves nine: Detecting and mitigating hallucinations of llms by validating low-confidence generation},
  author={Varshney, Neeraj and Yao, Wenlin and Zhang, Hongming and Chen, Jianshu and Yu, Dong},
  journal={arXiv preprint arXiv:2307.03987},
  year={2023},
  url={https://arxiv.org/abs/2307.03987}
}
